\documentclass[11pt]{article}

\usepackage[final]{style/acl}

\usepackage{times}
\usepackage{latexsym}

\usepackage[T1]{fontenc}
\usepackage{titletoc}
\usepackage[utf8]{inputenc}

\usepackage{microtype}

\usepackage{inconsolata}

\usepackage{graphicx}

\usepackage{hyperref}
\usepackage{url}
\usepackage{graphicx}
\usepackage{subcaption}

\usepackage{booktabs}
\usepackage{colortbl}

\usepackage{pifont}

\usepackage[english]{babel}
\usepackage{enumitem}

\usepackage{graphicx}
\usepackage[most]{tcolorbox}
\usepackage[export]{adjustbox}
\usepackage[ruled,vlined]{algorithm2e}

\usepackage{caption}

\usepackage{tabularx}
\usepackage{longtable}

\usepackage{tikz}

\usepackage{placeins}

\usepackage{makecell}
\usepackage{multirow}

\usepackage{tcolorbox}
\tcbset{
  enhanced,
  breakable,        
  rounded corners,
}
\tcbuselibrary{breakable}

\usepackage{array}
\usepackage{ragged2e}
\usepackage{ltablex}

\keepXColumns
\newcolumntype{Y}{>{\RaggedRight\arraybackslash}X}

\usepackage{booktabs}    
\usepackage{array}       
\usepackage{makecell}    
\usepackage{ragged2e}    
\usepackage{ltablex}     
\keepXColumns            
\usepackage{tabularray}
\UseTblrLibrary{booktabs}

\newcolumntype{Y}{>{\RaggedRight\arraybackslash}X}

\renewcommand{\arraystretch}{1.08}

\tcbset{width=\linewidth}     

\tcbuselibrary{breakable,skins,listings,theorems}
\title{Democratic ICAI: Debating Our Way to Steering Principles from Preferences}

\author{
 \textbf{Kevin Kingslin},
 \textbf{Anish Natekar},
 \textbf{Ashutosh Ranjan},
 \textbf{Vivek Srivastava},
\\
 \textbf{Savita Bhat},
 \textbf{Shirish Karande}
\\
\\
 TCS Research, India
\\
 {\small
 \begin{tabular}{c}
    \texttt{\{kevin.kingslin, anish.natekar, ashutosh.ranjan2, srivastava.vivek2,} \\
    \texttt{savita.bhat, shirish.karande\}@tcs.com}
 \end{tabular}
 }
}

\begin{document}
\maketitle
\begin{abstract}
Preference-based alignment often struggles to capture the reasoning that underlies human judgments. Many evaluations rely on multiple interacting criteria, yet pairwise labels reveal only the final choice rather than the considerations that shape preferences. Inverse Constitutional AI (ICAI) improves interpretability in decision making by summarizing preferences into natural-language principles, but its single-pass explanations miss much of the nuance involved in complex decisions. We introduce \textit{Democratic ICAI} (DICAI), a novel approach that gathers multiple competing rationales through structured persona debate, offering a broader and more expressive account of the factors influencing each comparison. From these richer signals, we derive clearer and more comprehensive steering principles and use them to guide preference decision modeling through both LLM-based and decision-tree judges, as well as downstream model training via constitution-induced preference labels. Experiments on creative preference benchmarks, MuCE-Pref and LiTBench, across multiple creative task categories show that Democratic ICAI yields a more faithful preference structure. It improves average preference prediction across tasks relative to deliberative prompting and principle-based baselines, while producing constitutions that LLM annotators prefer.
\end{abstract}

\section{Introduction}
\label{sec: intro}
\begin{figure*}[t]
\centering
\includegraphics[width=0.7\linewidth]{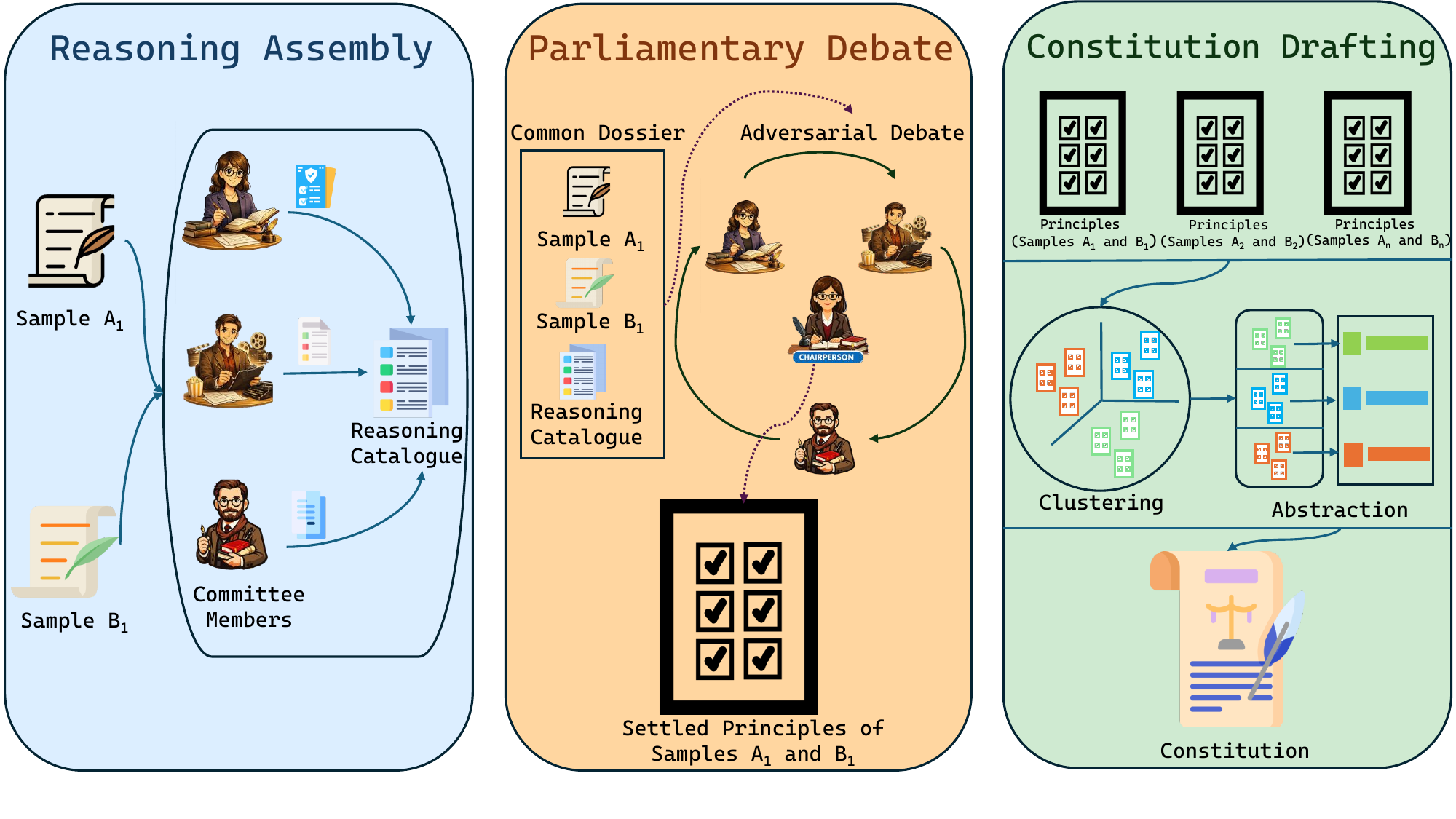}
\caption{\textbf{Architecture of \textit{Democratic ICAI}.} A committee of domain‑expert personas first generates detailed rationales for each preference pair. These rationales are then subjected to an adversarial debate procedure, through which the evaluative principles relevant to each comparison are surfaced. Finally, the full collection of principles is clustered and abstracted to draft a concise, human‑readable constitution.}
\label{fig: architecture}
\end{figure*}

Human preference data is central to aligning modern language models, yet many real-world judgment tasks demand far richer reasoning than what pairwise preference labels reveal. Evaluating creative writing, assessing alternative designs, comparing research explanations, or ranking artistic outputs often depends on qualitative and context-sensitive criteria that cannot be directly inferred from a single comparison. When annotators choose between two options, they integrate multiple considerations, but the recorded preference captures only the final decision and not the reasoning behind it. This gap poses fundamental challenges for preference-based alignment \citep{stiennon2020learning}.
These challenges emerge clearly in existing post-training pipelines. Reward models trained on preference pairs often latch onto superficial artifacts, which makes them vulnerable to reward hacking \citep{fu2025reward, liu2024rrm}. Human preference datasets themselves contain structural biases, such as favoring assertive or belief-matching responses over truthful ones \citep{findeis2025icai}, contributing to well-known issues like sycophancy \citep{sharma2023towards}. Evaluation systems that rely on LLM-as-a-judge introduce additional instability: single-shot judgments fluctuate under changes in prompt phrasing, sampling strategy, or formatting \citep{gu2024survey, schroeder2024can}. These problems intensify in open-ended or creative settings where judgments emerge from interactions among coherence, tone, originality, and stylistic intent.

Inverse Constitutional AI (ICAI) provides a promising direction by transforming preference datasets into natural-language principles \citep{findeis2025icai}. These constitutions can expose annotator biases and offer interpretable summaries of evaluative tendencies. However, ICAI relies on a single explanation per preference example, which limits the diversity of rationales it can uncover. Complex human judgments rarely hinge on one consideration, and single-pass explanations may miss important reasoning.

To address this limitation, we introduce \emph{Democratic ICAI} (refer Figure~\ref{fig: architecture}). Instead of relying on a single justification, we elicit multiple, competing rationales through a structured debate among expert personas. Debate provides a mechanism for surfacing information that individual annotators or single explanations may fail to reveal \citep{irving2018ai}. We then distill the resulting set of rationales into compact, human-readable \emph{steering principles}. These principles provide explicit guidance for generation, transparent evaluation, and downstream alignment, where better preference labels can provide cleaner supervision for model training. Our approach complements the inductive process in ICAI and aligns with evidence from Constitutional AI showing that clearly articulated principles can be used as training-time guidance to steer model behavior \citep{bai2022constitutional}.

Our key contributions are as follows:

\begin{enumerate}[noitemsep,nolistsep,leftmargin=*]
    \item We introduce \textbf{\textit{Democratic ICAI}}, a novel framework for deriving steering principles from human preference data through structured multi-persona deliberation.
    
    \item We discuss \textbf{constitution-guided decision modeling} and instantiate it with two complementary evaluators: an LLM-as-judge model and a decision-tree--based judge that operationalizes the learned principles.
    
    \item We analyze the \textbf{limitations of the ICAI algorithm} presented in \citep{findeis2025icai} when applied to creative task preference data, and we demonstrate how our approach more effectively captures nuanced reasoning in \textbf{complex human evaluative tasks}.
\end{enumerate}

\section{Related Work}

\textbf{Preference Learning and Post-Training:}
Preference-based alignment methods such as RLHF and DPO improve model behavior by training on human comparisons, yet they still compress rich judgments into a single opaque scalar. This scalar easily reflects artifacts like response length or formatting, which enables reward hacking and distorts preference structure \citep{stiennon2020learning, ouyang2022instructgpt, rafailov2023dpo, liu2024rrm}. Preference datasets also encode systematic biases, for example sycophancy, that aligned models can reproduce \citep{sharma2023towards}. These limitations motivate methods that expose preference criteria rather than relying on scalar rewards.

\textbf{LLM-as-a-Judge for Scalable Evaluation:}
LLM judges offer scalable evaluation for open-ended tasks. MT-Bench and Chatbot Arena show strong agreement with human judgments but also reveal position effects, verbosity bias, and sensitivity to prompt phrasing that weaken single-shot reliability \citep{zheng2023judging}. Rubric-based protocols such as G-Eval improve consistency by breaking decisions into explicit sub-criteria, although they still depend on one-pass reasoning and can inherit hidden heuristics from the base model \citep{liu2023g}. A recent survey underscores the need for transparent, decomposed procedures rather than solitary overall scores \citep{gu2024survey}.

\textbf{Debate and Deliberative Supervision:}
Debate has been proposed to surface information that direct judging may miss, with theory suggesting that structured debate can reveal evidence otherwise inaccessible to single evaluators \citep{irving2018ai}. However, empirical work on multi-agent debate reports collapse into echo-chambers or majority convergence when protocols lack safeguards, especially among similar models \citep{estornell2024multi, liang2024encouraging}. Effective debate therefore requires role design, turn-taking, and aggregation mechanisms that maintain diversity of reasoning.

\textbf{Principle-Based Alignment:}
Constitutional AI demonstrates that explicit, human-readable principles can guide behavior through AI-mediated feedback \citep{bai2022constitutional}. Inverse Constitutional AI (ICAI) derives such principles from preference data by generating candidate explanations and selecting those that best reconstruct the dataset, revealing latent evaluative dimensions and biases \citep{findeis2025icai}. Single-pass explanations and clustering may miss less frequent criteria and blur distinct evaluative dimensions. Grounded Constitutional AI (GCAI) \citep{bell2026beyond} incorporates human-written reasons and value statements to better ground principles in stakeholder perspectives, while AutoRubric \citep{xie2025auto} extracts explicit rubrics from preference data to improve transparency. However, these approaches typically rely on single-agent reasoning or filtering principles based on training-set reconstruction, leaving limited mechanisms for surfacing and reconciling multiple competing rationales.

\textbf{Positioning Democratic ICAI:}
Across these areas, existing methods often compress judgments too aggressively or rely on single-shot reasoning that fails to capture the broad space of criteria. Our approach, \emph{Democratic ICAI}, uses structured debate to elicit multiple competing justifications, then distills them into a compact set of steering principles that are easy to inspect and apply. Instead of one scalar score or one explanation, we evaluate with an adaptive sequence of clear binary checks. This increases transparency, improves reliability, and makes it easier to identify where and why disagreements or failures occur.

\begin{figure*}[t]
    \centering

    \begin{subfigure}[t]{0.24\textwidth}
        \centering
        \includegraphics[width=\linewidth]{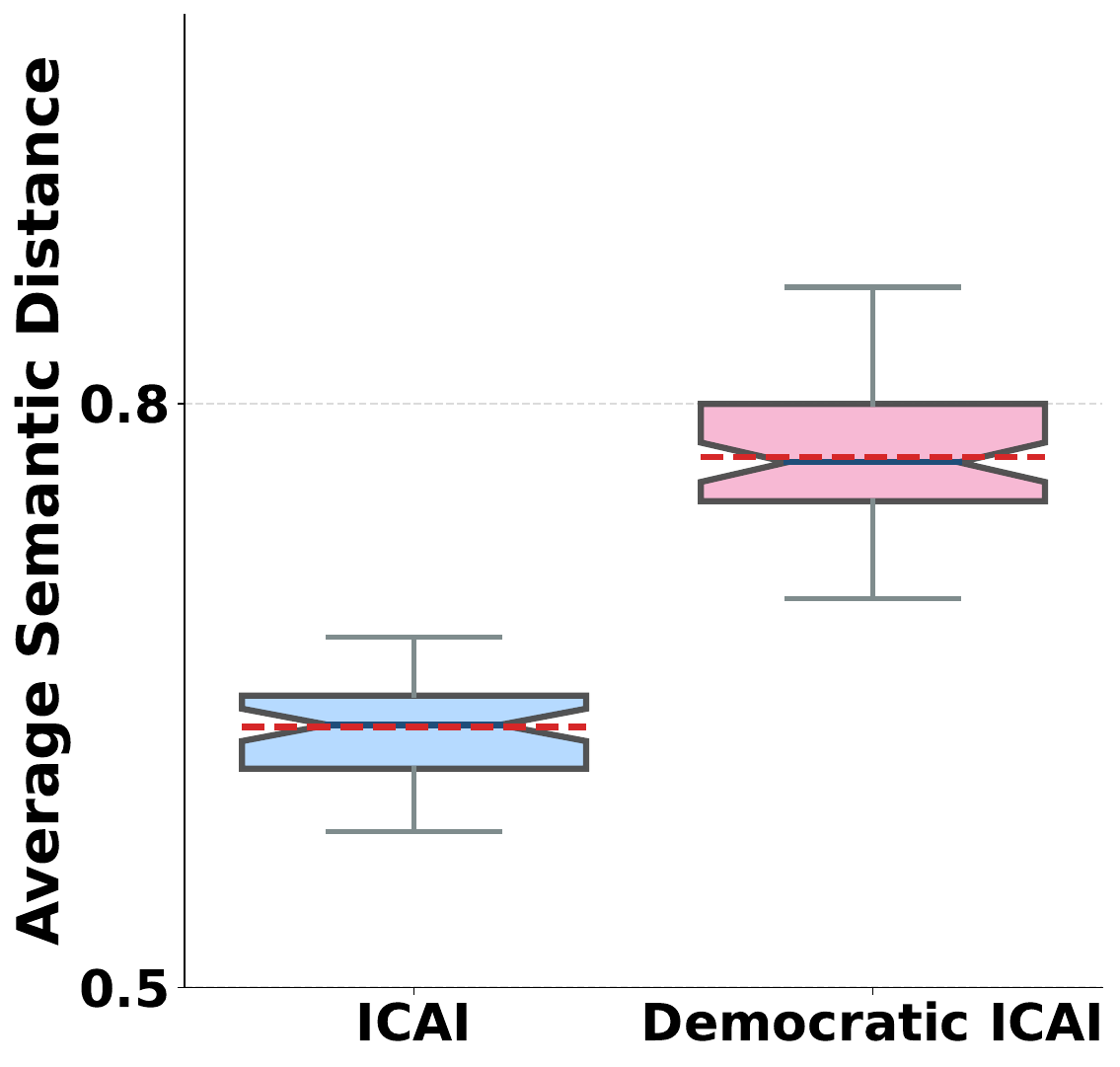}
        \caption{Long Stories (LiTBench)}
    \end{subfigure}
    \hfill
    \begin{subfigure}[t]{0.24\textwidth}
        \centering
        \includegraphics[width=\linewidth]{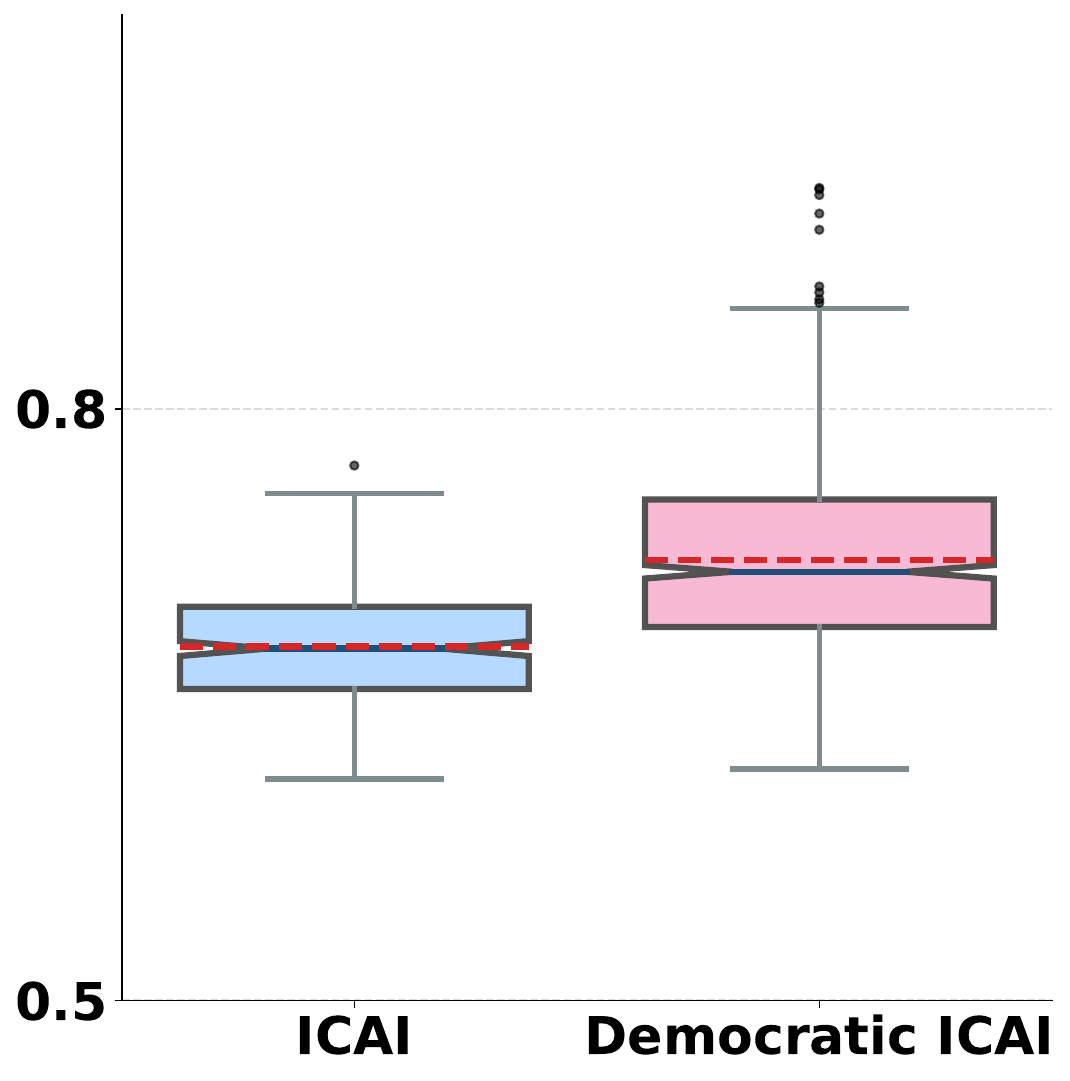}
        \caption{Hypothesis Generation}
    \end{subfigure}
    \hfill
    \begin{subfigure}[t]{0.24\textwidth}
        \centering
        \includegraphics[width=\linewidth]{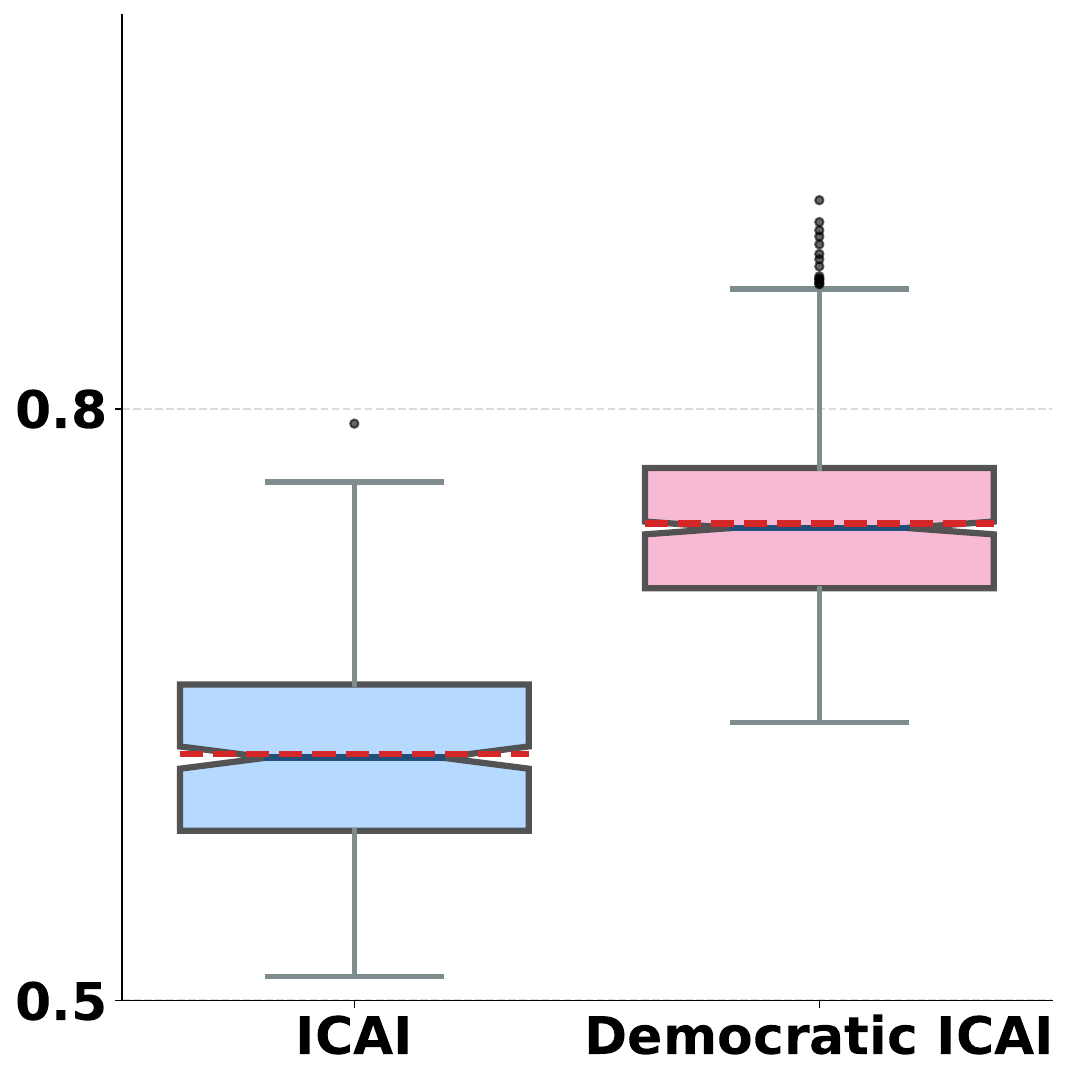}
        \caption{Metaphors}
    \end{subfigure}
    \hfill
    \begin{subfigure}[t]{0.24\textwidth}
        \centering
        \includegraphics[width=\linewidth]{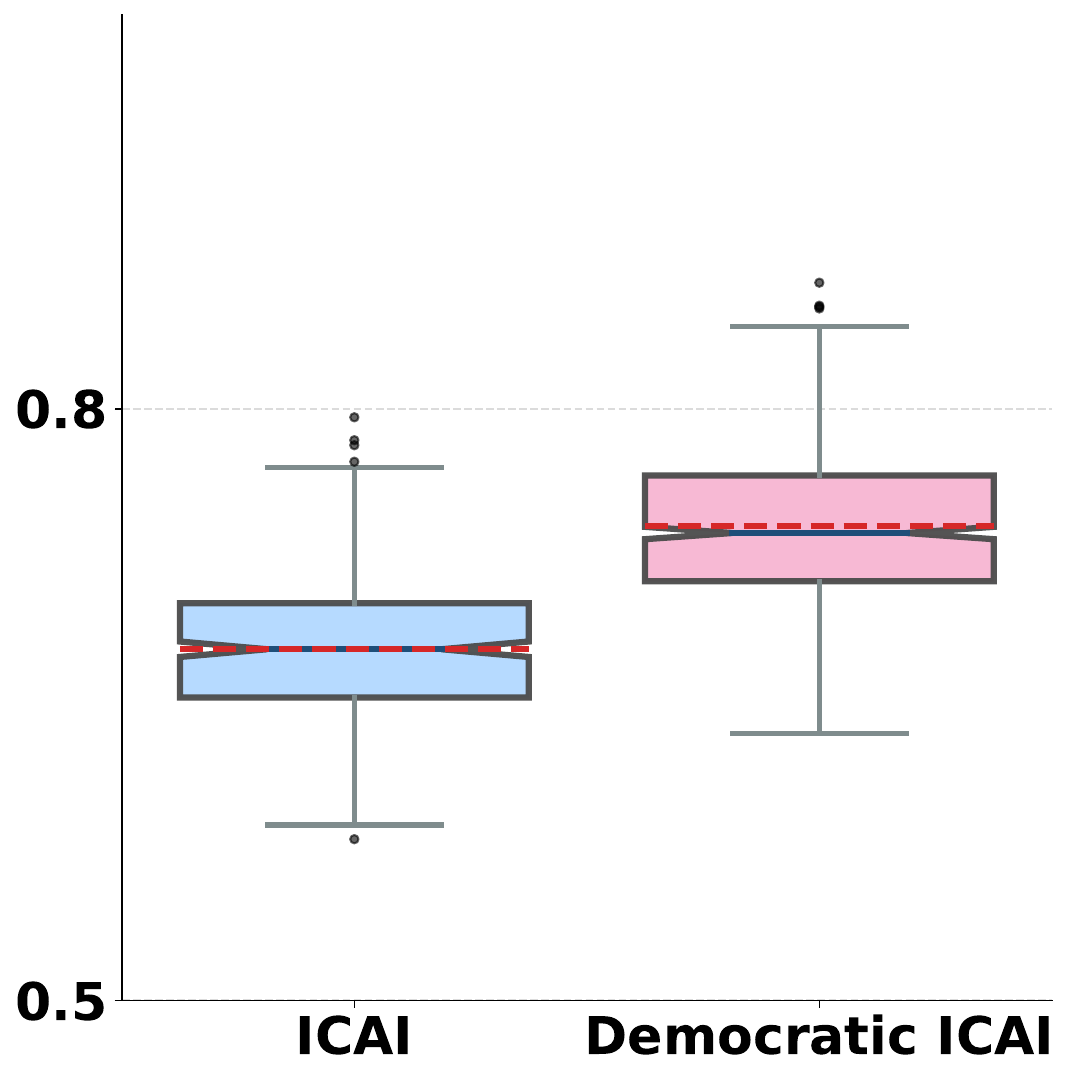}
        \caption{Creative Problem Solving}
    \end{subfigure}

    \caption{
    \textbf{Distribution of average semantic distance between principles within a constitution.}
    For each method, the distance is computed as the average cosine distance of a principle from all other principles in the constitution. Lower values indicate reduced diversity (narrower constitutional scope), while higher values reflect greater conceptual separation and normative breadth.
    }
    \label{fig:constitutional_distance}
\end{figure*}

\section{Our Approach}

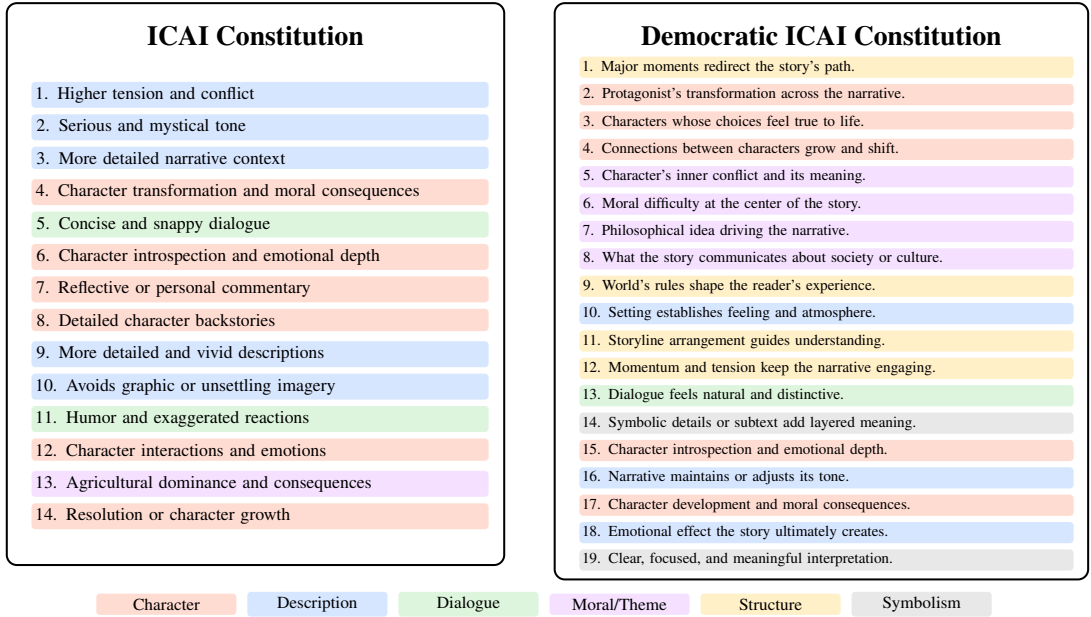
\begin{figure*}[t]
\centering
\resizebox{0.9\textwidth}{!}{
\begin{tikzpicture}[
   font=\scriptsize,
   box/.style={draw, rounded corners=4pt, thick, inner sep=5pt},
   item/.style={rounded corners=1.5pt, inner sep=2pt, text width=6.2cm, align=left},
   title/.style={font=\bfseries\normalsize}
]
\definecolor{charcol}{RGB}{255,220,210}
\definecolor{desccol}{RGB}{215,230,255}
\definecolor{dialogcol}{RGB}{220,245,220}
\definecolor{moralcol}{RGB}{245,225,255}
\definecolor{structcol}{RGB}{255,240,200}
\definecolor{symcol}{RGB}{232,232,232}
\node[box, minimum width=6.9cm, minimum height=7.8cm, anchor=north west] (icai) at (0,0) {};
\node[title, anchor=north] at ([yshift=-0.20cm]icai.north) {ICAI Constitution};
\node[item, fill=desccol, anchor=north west] at (0.35,-1.10) {1. Higher tension and conflict};
\node[item, fill=desccol, anchor=north west] at (0.35,-1.55) {2. Serious and mystical tone};
\node[item, fill=desccol, anchor=north west] at (0.35,-2.00) {3. More detailed narrative context};
\node[item, fill=charcol, anchor=north west] at (0.35,-2.45) {4. Character transformation and moral consequences};
\node[item, fill=dialogcol, anchor=north west] at (0.35,-2.90) {5. Concise and snappy dialogue};
\node[item, fill=charcol, anchor=north west] at (0.35,-3.35) {6. Character introspection and emotional depth};
\node[item, fill=charcol, anchor=north west] at (0.35,-3.80) {7. Reflective or personal commentary};
\node[item, fill=charcol, anchor=north west] at (0.35,-4.25) {8. Detailed character backstories};
\node[item, fill=desccol, anchor=north west] at (0.35,-4.70) {9. More detailed and vivid descriptions};
\node[item, fill=desccol, anchor=north west] at (0.35,-5.15) {10. Avoids graphic or unsettling imagery};
\node[item, fill=dialogcol, anchor=north west] at (0.35,-5.60) {11. Humor and exaggerated reactions};
\node[item, fill=charcol, anchor=north west] at (0.35,-6.05) {12. Character interactions and emotions};
\node[item, fill=moralcol, anchor=north west] at (0.35,-6.50) {13. Agricultural dominance and consequences};
\node[item, fill=charcol, anchor=north west] at (0.35,-6.95) {14. Resolution or character growth};
\node[box, minimum width=7.4cm, minimum height=8.0cm, anchor=north west] (dicai) at (7.6,0) {};
\node[title, anchor=north] at ([yshift=-0.20cm]dicai.north) {Democratic ICAI Constitution};
\tikzset{ditem/.style={rounded corners=1.5pt, inner sep=1.5pt, text width=6.75cm, align=left, font=\tiny}}
\node[ditem, fill=structcol, anchor=north west] at (7.95,-0.75) {1. Major moments redirect the story's path.};
\node[ditem, fill=charcol, anchor=north west] at (7.95,-1.13) {2. Protagonist's transformation across the narrative.};
\node[ditem, fill=charcol, anchor=north west] at (7.95,-1.51) {3. Characters whose choices feel true to life.};
\node[ditem, fill=charcol, anchor=north west] at (7.95,-1.89) {4. Connections between characters grow and shift.};
\node[ditem, fill=moralcol, anchor=north west] at (7.95,-2.27) {5. Character's inner conflict and its meaning.};
\node[ditem, fill=moralcol, anchor=north west] at (7.95,-2.65) {6. Moral difficulty at the center of the story.};
\node[ditem, fill=moralcol, anchor=north west] at (7.95,-3.03) {7. Philosophical idea driving the narrative.};
\node[ditem, fill=moralcol, anchor=north west] at (7.95,-3.41) {8. What the story communicates about society or culture.};
\node[ditem, fill=structcol, anchor=north west] at (7.95,-3.79) {9. World's rules shape the reader's experience.};
\node[ditem, fill=desccol, anchor=north west] at (7.95,-4.17) {10. Setting establishes feeling and atmosphere.};
\node[ditem, fill=structcol, anchor=north west] at (7.95,-4.55) {11. Storyline arrangement guides understanding.};
\node[ditem, fill=structcol, anchor=north west] at (7.95,-4.93) {12. Momentum and tension keep the narrative engaging.};
\node[ditem, fill=dialogcol, anchor=north west] at (7.95,-5.31) {13. Dialogue feels natural and distinctive.};
\node[ditem, fill=symcol, anchor=north west] at (7.95,-5.69) {14. Symbolic details or subtext add layered meaning.};
\node[ditem, fill=charcol, anchor=north west] at (7.95,-6.07) {15. Character introspection and emotional depth.};
\node[ditem, fill=desccol, anchor=north west] at (7.95,-6.45) {16. Narrative maintains or adjusts its tone.};
\node[ditem, fill=charcol, anchor=north west] at (7.95,-6.83) {17. Character development and moral consequences.};
\node[ditem, fill=desccol, anchor=north west] at (7.95,-7.21) {18. Emotional effect the story ultimately creates.};
\node[ditem, fill=symcol, anchor=north west] at (7.95,-7.59) {19. Clear, focused, and meaningful interpretation.};
\end{tikzpicture}}

\resizebox{0.75\textwidth}{!}{
\begin{tikzpicture}[font=\scriptsize]
\definecolor{charcol}{RGB}{255,220,210}
\definecolor{desccol}{RGB}{215,230,255}
\definecolor{dialogcol}{RGB}{220,245,220}
\definecolor{moralcol}{RGB}{245,225,255}
\definecolor{structcol}{RGB}{255,240,200}
\definecolor{symcol}{RGB}{232,232,232}
\tikzset{leg/.style={rounded corners=1.5pt, inner sep=2pt, text width=1.9cm, align=center}}
\node[leg, fill=charcol] at (0,0) {Character};
\node[leg, fill=desccol] at (2.2,0) {Description};
\node[leg, fill=dialogcol] at (4.4,0) {Dialogue};
\node[leg, fill=moralcol] at (6.6,0) {Moral/Theme};
\node[leg, fill=structcol] at (8.8,0) {Structure};
\node[leg, fill=symcol] at (11.0,0) {Symbolism};
\end{tikzpicture}
}
\caption{
\textbf{Qualitative comparison of constitution for Long Stories (LiTBench, GPT-4o)}. ICAI repeatedly emphasizes overlapping Character, Description, and Dialogue principles, while Democratic ICAI distributes principles across Character, Description, Dialogue, Moral/Theme, Structure, and Symbolism, indicating broader coverage.
}
\label{fig:constitution_redundancy}
\end{figure*}

\subsection{Reasoning Assembly}
\label{sec: reasoning}
Given a dataset of paired samples $(A_i, B_i)$ with human preference labels, we generate rationales explaining why the selected option is preferred. These rationales are produced by a committee of $P$ expert persona agents designed to represent expertise relevant to the task. To further diversify the produced rationales, each expert persona agent is paired with a distinct reasoning strategy: chain-of-thought \citep{wei2022chainofthought}, self-consistency \citep{wang2022selfconsistency}, or self-refinement/reflective justification \citep{madaan2023selfrefine}. The corresponding prompts are provided in Figures~\ref{cot_strategy}, \ref{self_consistency_strategy}, and~\ref{self_refine_strategy} in Appendix~\ref{app:dicai_prompts}. These strategies encourage complementary reasoning modes, improving coverage and reducing reliance on any single heuristic.  As a result, each expert persona agent differs in \textit{what} each member treats as a signal of quality, while strategies shape \textit{how} each member arrives at a justification.

\noindent For each pair $(A_i, B_i)$, every persona independently generates a rationale conditioned on the preference label using its assigned reasoning strategy. The resulting set of rationales is collected as
\begin{equation}
   R_i = \{r_i^{(1)}, r_i^{(2)}, \ldots, r_i^{(P)}\},
\end{equation}
forming a Reasoning Catalogue, which is then passed to the parliamentary debate stage.

\subsection{Parliamentary Debate}
\label{sec:debate}
The assembled rationales capture diverse perspectives but may also contain redundancy, superficial reasoning or conflicting interpretations. We refine them through a structured debate that extracts and consolidates evaluative principles, with a chairperson agent enforcing de-duplication and consistency \cite{liang2024encouraging}. 

\noindent To operationalize this, we introduce a structured adversarial debate mechanism \citep{foster2004defense,huang2025dare}, illustrated in Figure~\ref{fig: architecture}. In this mechanism, persona debater agents examine the same preference pair $(A_i, B_i)$, argue for the principles implied by their rationales, and critique each other's claims. This turns the initial rationale set into a set of evaluative principles rather than isolated explanations. In each round, persona debater agents can propose principles, defend them with evidence, challenge competing principles, revise their own principles, or withdraw unsupported claims, whereas the chairperson agent can request grounding, merge duplicates, discard weak principles, and produce a settled set of principles. After $T$ rounds, the chairperson agent reviews the debate transcript, groups similar principles, discards unsupported principles, and rewrites the remaining principles into concise final principles. These are collected as a set of settled principles. The prompts for chairperson and persona debater agents are provided in Appendix~\ref{app:dicai_prompts} (Figures~\ref{chairperson_prompt} and~\ref{debater_prompt}).
\begin{equation}
Q_i = \left\{ Q_i^{(1)}, Q_i^{(2)}, \ldots \right\}.
\end{equation}

\subsection{Constitution Drafting}
The settled principles obtained across all sample pairs are aggregated into a unified pool
\begin{equation}
Q^{all} = \bigcup_{i=1}^{N} Q_i,
\end{equation}
where $N$ denotes the number of sample pairs and $Q_i$ is the set of refined principles for pair $(A_i, B_i)$.

\noindent To transform this large collection into a generalised constitution, we adopt a two-stage process: clustering groups semantically similar principles, and abstraction rewrites each group into a broader reusable principle.

\noindent First, we group semantically similar principles by performing embedding-based clustering over $Q^{all}$, yielding a set of clusters:
\begin{equation}
\mathcal{C} = \{ C_1, C_2, \ldots, C_M \},
\end{equation}
where each cluster $C_j \subseteq Q^{\text{all}}$ contains principles with similar semantic intent. Next, for each cluster $C_j$, we generate an abstract representative principle $\tilde{Q}_j$ that captures its shared rationale while removing sample-specific details:
\begin{equation}
\tilde{Q}_j = \mathcal{A}(C_j),
\end{equation}
where $\mathcal{A}(\cdot)$ denotes an abstraction operator. The prompt for the abstraction operator $\mathcal{A}(\cdot)$ is presented in Figure~\ref{abstraction_prompt} in Appendix \ref{app:dicai_prompts}.

\noindent The final constitution is defined as:
\begin{equation}
\mathcal{K} = \{ \tilde{Q}_1, \tilde{Q}_2, \ldots, \tilde{Q}_M \},
\end{equation}
where each $\tilde{Q}_j$ is a human-readable principle representing a core dimension of human preference.
\subsection{Constitution-Guided Inference}
\label{sec: judge}
Once the constitution $\mathcal{K}$ is induced, we use it as an explicit decision rule for preference reconstruction. Each judge applies the principles to a pair of responses and predicts which better satisfies the constitution. We implement this in two ways:
\begin{itemize}[noitemsep,nolistsep,leftmargin=*]
    \item \textbf{LLM-based judge}: The LLM judge is prompted with the constitution and the pair, and asked to adjudicate between the two samples according to the listed principles, following the paradigm of \cite{findeis2025icai}. Comparing its judgments against the human preference labels measures how well the constitution captures the underlying preference structure under language-based application of the principles. We present the prompt for the LLM-judge in Figure~\ref{ICAI_prompt} in Appendix~\ref{app:dicai_prompts}.
    \item \textbf{Decision Tree Judge}: The LLM judge applies principles holistically and offers no audit trail at the level of individual principles. To complement it, we operationalize a decision-tree judge. For each pair $(A_i, B_i)$, we construct a feature table by scoring both samples against every principle $\tilde{Q}_j \in \mathcal{K}$ on a 1--5 rubric. We train a decision tree on the feature table to predict the human preference $y_i$, and use the trained tree as final judge. We acknowledge the relatively higher cost of the decision-tree judge, which is intended as an interpretability-focused alternative rather than the default inference mechanism. Additional details on training and inference using decision trees are provided in Appendix~\ref{sec: dt_details}.
\end{itemize}
\begin{table*}[t]
\centering
\small
\setlength{\tabcolsep}{4pt}
\renewcommand{\arraystretch}{1.05}
\resizebox{\textwidth}{!}{
\begin{tabular}{lccccccc}
\toprule
Task & CoT & CoT-SC & ToT & Self-Refine & AutoRubric & ICAI & DICAI \\
\midrule
Alternate Uses of Objects & \textbf{78.61} & 70.30 & 76.46 & 73.78 & 58.70 & 66.40 & 74.23 \\
Consequences & 48.29 & 47.24 & 47.40 & 51.10 & 52.44 & 61.20 & \textbf{73.21} \\
Design Solutions & 49.25 & 42.29 & 40.20 & 38.61 & 45.05 & 75.20 & \textbf{75.26} \\
Experiment Design & 65.66 & 63.31 & 62.41 & 64.53 & 55.46 & 73.80 & \textbf{79.93} \\
Hypothesis Generation & 56.10 & 54.96 & 55.42 & 52.56 & 54.92 & 64.60 & \textbf{77.22} \\
Metaphors & 40.00 & 60.00 & 60.00 & 60.00 & 60.00 & 71.40 & \textbf{74.01} \\
Real-Life Creative Problem Solving & 38.07 & 35.99 & 35.60 & 32.01 & 42.39 & 56.00 & \textbf{70.03} \\
Research Questions & 56.80 & 50.78 & 53.55 & 53.79 & 53.55 & 70.20 & \textbf{80.21} \\
Long Stories (LiTBench) & 70.69 & 66.15 & 69.28 & \textbf{71.90} & 59.22 & 62.89 & 68.70 \\
Short Stories (MuCE) & 71.79 & 67.09 & 68.75 & 67.50 & 69.51 & 71.40 & \textbf{78.22} \\
\midrule
Average & 57.53 $\pm$ 13.15 & 55.81 $\pm$ 10.92 & 56.91 $\pm$ 12.46 & 56.58 $\pm$ 13.06 & 55.12 $\pm$ 7.31 & 67.31 $\pm$ 5.82 & \textbf{75.10 $\pm$ 3.69} \\
\bottomrule
\end{tabular}
}

\caption{\textbf{Preference accuracy (\%) across tasks for Democratic ICAI and baseline methods using LLM as a judge (GPT-4o).} Bold values indicate the best performance per task.}
\label{tab:4o_results}
\end{table*}

\begin{table*}[t]
\centering
\small
\setlength{\tabcolsep}{4pt}
\renewcommand{\arraystretch}{1.05}

\resizebox{\textwidth}{!}{
\begin{tabular}{lccccccc}
\toprule
Task & CoT & CoT-SC & ToT & Self-Refine & AutoRubric & ICAI & DICAI \\
\midrule
Alternate Uses of Objects & 74.94 & 70.77 & \textbf{77.72} & 75.35 & 59.22 & 69.40 & 73.22 \\
Consequences & 54.56 & 50.33 & 52.27 & 52.51 & 56.77 & 67.80 & \textbf{75.23} \\
Design Solutions & 46.27 & 43.50 & 43.56 & 45.05 & 45.88 & 79.80 & \textbf{80.21} \\
Experiment Design & 66.22 & 64.96 & 64.87 & 62.52 & 57.12 & 73.40 & \textbf{81.77} \\
Hypothesis Generation & 56.10 & 55.16 & 54.13 & 48.43 & 56.79 & 66.40 & \textbf{72.80} \\
Metaphors & 60.00 & 60.00 & 60.00 & 60.00 & 62.33 & 74.40 & \textbf{75.22} \\
Real-Life Creative Problem Solving & 38.37 & 34.33 & 35.18 & 32.11 & 43.65 & 64.20 & \textbf{77.05} \\
Research Questions & 58.12 & 57.28 & 56.96 & 54.87 & 58.76 & 71.80 & \textbf{76.20} \\
Long Stories (LiTBench) & 70.69 & 68.43 & \textbf{71.29} & 68.02 & 63.21 & 66.80 & 71.20 \\
Short Stories (MuCE) & 75.64 & 73.42 & 73.41 & 67.09 & 72.22 & \textbf{77.60} & 77.00 \\
\midrule
Average & 60.09 $\pm$ 11.51 & 57.82 $\pm$ 11.83 & 58.94 $\pm$ 12.74 & 56.59 $\pm$ 12.08 & 57.59 $\pm$ 7.78 & 71.16 $\pm$ 4.86 & \textbf{75.99 $\pm$ 3.09} \\
\bottomrule
\end{tabular}
}

\caption{\textbf{Preference accuracy (\%) across tasks for Democratic ICAI and baseline methods using LLM as a judge (GPT-5).} Bold values indicate the best performance per task.}
\label{tab:5_results}
\end{table*}




\section{Experimental Setup} \label{sec:expsetup}

\subsection{Dataset and Models}
We conduct our experiments primarily using the MuCE-Pref dataset \cite{ismayilzada2025creative}, which captures preference judgments across a diverse range of creative tasks. We list the MuCE tasks in Table~\ref{tab:muce_dataset} in Appendix~\ref{sec: additional_details}. For each task, we sample 500 training pairs to derive task-specific constitutions, ensuring that each pair contains unique accepted and rejected responses. Evaluation is performed on the official test split for each category containing in total 4500 samples, allowing us to assess whether the learned principles generalize to unseen creative examples.
Additionally, we include LiTBench \cite{fein2025litbench} to complement our evaluation with long-form creative writing, particularly story generation. For this setting, we use 1,000 pairs for constitution induction and 2,000 held-out pairs for evaluation. 

For our experiments, we use two models: GPT-4o \citep{hurst2024gpt} and GPT-5 \citep{singh2025openai} as an LLM judge.

\subsection{Baselines}
We compare against two classes of baselines to separate the effect of stronger LLM reasoning from the effect of explicit principle induction. First, \textbf{deliberative prompting methods} evaluate whether an LLM-as-a-judge can reconstruct pairwise preferences directly without inducing a constitution \citep{zheng2023judging,gu2024survey}. These methods are also widely used to improve reasoning performance, making them natural baselines for testing whether preference reconstruction gains can be obtained through prompting alone. This group includes \textbf{Chain-of-Thought (CoT)} \citep{wei2022chainofthought}, which prompts the LLM judge to produce step-by-step reasoning before issuing a preference label; \textbf{Self-Consistency (CoT-SC)} \citep{wang2022selfconsistency}, which samples multiple reasoning paths and aggregates their outputs via majority vote; \textbf{Tree of Thought (ToT)} \citep{yao2023treeofthoughts}, which formulates preference prediction as a branching search over intermediate judgments; and \textbf{Self-Refine} \citep{madaan2023selfrefine}, which iteratively improves an initial prediction via self-critique. The corresponding prompts are provided in Appendix~\ref{app:deliberative_prompts}.

\noindent Second, \textbf{principle-based methods} evaluate whether constructing explicit evaluative principles provides useful structure beyond prompting-based reasoning alone. This group includes \textbf{ICAI} \citep{findeis2025icai}, which induces natural-language principles from preference pairs and uses them for prediction in a single pass, and \textbf{AutoRubric} \citep{xie2025auto}, which extracts rubrics from preference data and applies them through an LLM judge to produce preference labels.
\subsection{Training Details}
\label{training_details_and_eval}
\paragraph{Reasoning trace generation.}
We first generate candidate reasoning traces and principles using a reasoning committee composed of $P=3$ expert persona agents tailored to the target domain. We use three personas to cover multiple domain-relevant viewpoints while keeping the debate tractable. Table~\ref{tab:persona_table} in Appendix~\ref{app:training_details} provides the task-wise expert persona committee details, and Appendix~\ref{app:reasoning_personas} provides example expert persona agent prompts. Each expert persona agent is randomly assigned one reasoning strategy: Chain-of-Thought, Self-Consistency, or Self-Refine, to guide the generation process. The corresponding prompts for these strategies are provided in Appendix~\ref{app:dicai_prompts} (Figures~\ref{cot_strategy}, \ref{self_consistency_strategy}, and~\ref{self_refine_strategy}).

\paragraph{Principle refinement and clustering.}
The resulting principles are refined through a structured multi‑agent debate implemented with AutoGen \citep{wu2024autogen} (prompts shown in Figures~\ref{chairperson_prompt} and~\ref{debater_prompt} in Appendix~\ref{app:dicai_prompts}). The debate runs for three rounds, after which the chairperson agent prepares the settled principles. To further ensure diversity and remove redundancy among principles, we embed all candidates and apply K‑Means clustering. The embedding model used is OpenAI's Text-Embedding-3 small. We determine the number of clusters using the silhouette score, which quantifies the trade‑off between intra‑cluster cohesion and inter‑cluster separation. We set $M = M^*$, where $M^*$ denotes the value that maximizes the silhouette score (refer Table~\ref{tab:silhouette_optimal_clusters} in Appendix~\ref{app:ablation_studies}). Ablations on abstraction strategy, clustering granularity, and constitution size are provided in Tables~\ref{tab:ablation_abstraction_strategy}, \ref{tab:ablation_clustering_granularity}, and~\ref{tab:ablation_induced_principles} in Appendix~\ref{app:ablation_studies}.

\paragraph{Constitution-aligned model training.}
\label{sec:constitution_aligned_model_eval}
We further test whether induced constitutions provide useful supervision for downstream model training. For each preference-label source (CoT, CoT-SC, ToT, Self-Refine, AutoRubric, ICAI, and DICAI), we construct MuCE preference labels using either its induced constitution or its corresponding prompting procedure. We then train one CrPO-style DPO model \citep{rafailov2023dpo, ismayilzada2025creative} per label source for each base model, using Llama-3.1-8B-Instruct and Mistral-7B-Instruct. All runs use LoRA-based DPO \citep{hu2022lora} with fixed hyperparameters. Runs differ only in preference-label source. We also include CrPO-cre, the creativity-optimized CrPO variant from \citet{ismayilzada2025creative}, as a reference baseline. Results for Llama-3.1-8B-Instruct are reported in Table~\ref{tab:dpo_llama_results}. The full label-generation procedure, CrPO details, DPO configuration, Mistral-7B-Instruct results, and Llama qualitative examples are provided in Appendix~\ref{app:constitution_dpo_results} (Section~\ref{sec:crpo_details}, Tables~\ref{tab:constitution_dpo_config}, \ref{tab:dpo_mistral_results}, and~\ref{tab:choice_poem_examples}).

\subsection{Evaluation Details}

\noindent For evaluation, we consider four dimensions:
\begin{itemize}[noitemsep,nolistsep,leftmargin=*]
    \item \textbf{Constitution Quality Evaluation.} First, we directly evaluate the quality of the constitutions themselves. We draw from foundational work in political philosophy, constitutional theory, and recent AI alignment literature. Following Lon Fuller’s \textit{The Morality of Law} \citep{fuller1969morality}, and subsequent work by Nwokora \citep{nwokora2022feasibility} and Hedling \citep{hedling2023faithfulness}, we evaluate constitutions along five dimensions: \textbf{generality}, \textbf{clarity}, \textbf{coherence}, \textbf{feasibility}, and \textbf{faithfulness}.
    An LLM judge is prompted with ICAI and Democratic ICAI constitutions and asked to compare them pairwise along each dimension. For each dimension, we report the preference share, that is, the percentage of comparisons in which the evaluator selects one constitution as better than the other. To mitigate circularity and model-specific bias, we perform this evaluation using three different models Qwen2.5-32B \citep{qwen2025qwen25technicalreport}, GPT-4o, and GPT-5. The prompt for the LLM annotator is presented in Figure~\ref{consitution_analysis} in Appendix~\ref{app:dicai_prompts}. Detailed definitions of these dimensions are provided in Section~\ref{sec: const_qual_appendix} in the Appendix.
    
    \item \textbf{Preference Reconstruction Evaluation.} Second, as described in Section~\ref{sec: judge}, we test how well each constitution guides preference decisions by comparing predictions from a constitution-guided LLM judge and a Decision Tree judge against human preference labels. This evaluation measures how accurately the extracted principles reconstruct the underlying preference structure, with decision-tree construction and inference details provided in Appendix~\ref{sec: dt_details}.
    
    \item \textbf{Bias, spuriousness, and harmfulness audit.} While the constitution-quality evaluation measures whether a constitution is well formed, a well-formed constitution may still encode unsafe or unreliable decision criteria. We therefore audit each induced principle along three axes. First, following concerns raised in ICAI that induced constitutions can partially reflect annotator biases, we measure whether principles encode or propagate demographic or social biases. Second, we audit for spurious criteria, such as response length, verbosity, formatting, or stylistic artifacts, which may correlate with dataset preferences without representing genuine quality and can reduce cross-domain generalization. Prior work \citep{xie2025auto, liu2024rrm} has shown that rule-generation methods can capture criteria that are either highly specific or superficial. Third, motivated by the observation that reconstructed principles can reveal potentially harmful interpretations encoded in preference datasets, we evaluate harmfulness by measuring whether principles could steer a judge toward unsafe, biased, or otherwise undesirable decisions. Specifically, we measure the proportion of generated principles that are harmful, i.e., principles that encourage unsafe, biased, or otherwise undesirable judgments.

    To reduce circularity, we use Qwen2.5-32B \citep{qwen2025qwen25technicalreport} as the primary external auditor rather than the GPT models used to generate constitutions. Using a different model family reduces self-preference effects \citep{panickssery2024llm}, where an evaluator may favor outputs aligned with its own generation style, and provides a more independent audit of GPT-generated principles. To further validate the robustness of our findings across evaluator families, we additionally conduct evaluations using Llama-3.1-70B \citep{grattafiori2024llama}. The audit prompt is provided in Figure~\ref{fig:audit-prompt} in Appendix~\ref{app:dicai_prompts}.

    \item \textbf{Constitution-Aligned Model Evaluation.} Finally, we test whether the induced preference labels can supervise downstream creative generation. Following CrPO \citep{ismayilzada2025creative}, we evaluate each trained model on held-out MuCE prompts using novelty, diversity, surprise, and quality, where higher values indicate stronger performance (details in Appendix~\ref{sec:crpo_details}). The prompt set, decoding setup, and scoring pipeline are fixed across methods, isolating the effect of the preference-label source on downstream generation quality.


\end{itemize}






\begin{figure*}[t]
\centering

\begin{subfigure}{0.32\textwidth}
    \centering
    \includegraphics[width=\linewidth]{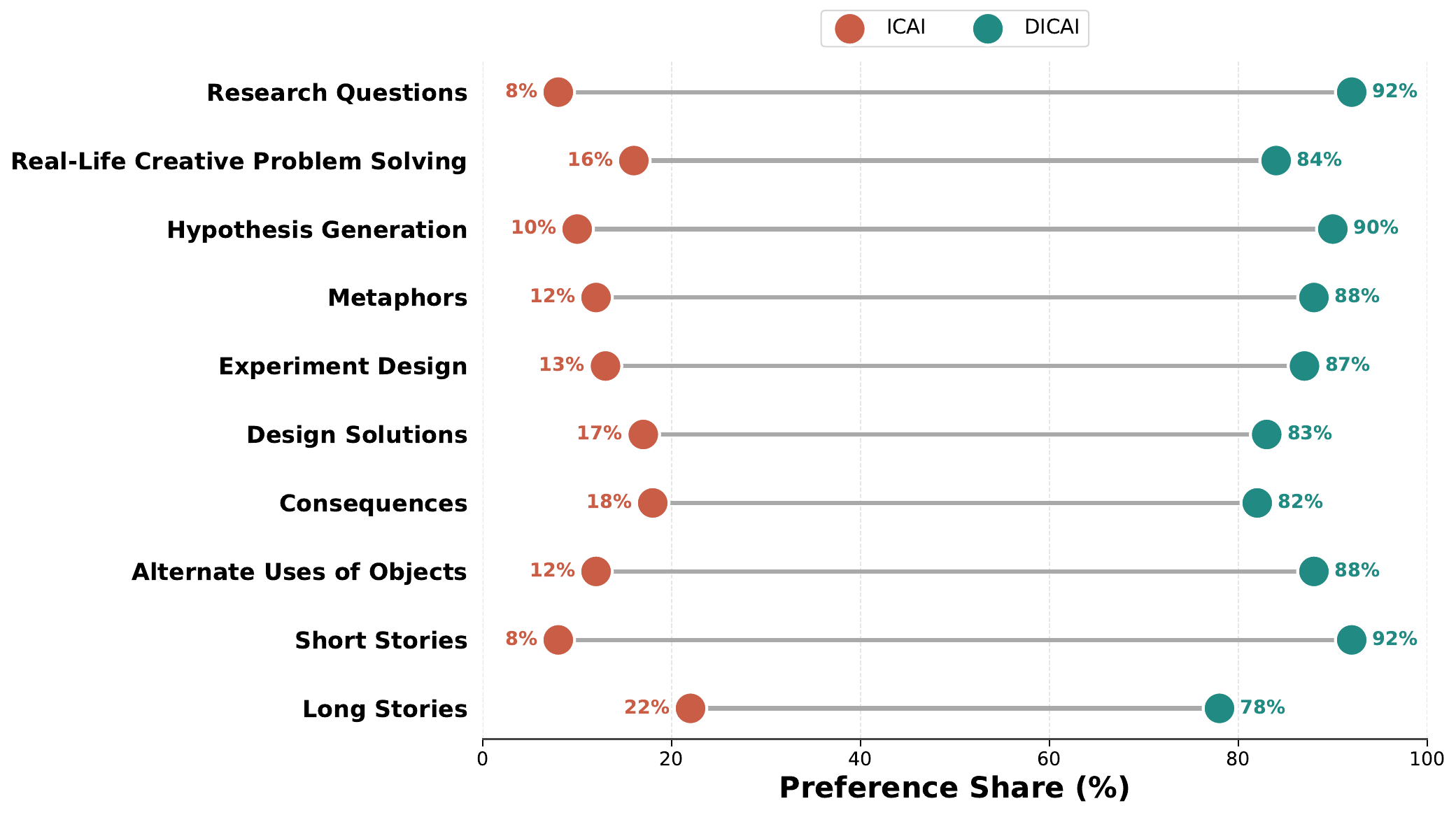}
    \caption{Generality}
\end{subfigure}
\hfill
\begin{subfigure}{0.32\textwidth}
    \centering
    \includegraphics[width=\linewidth]{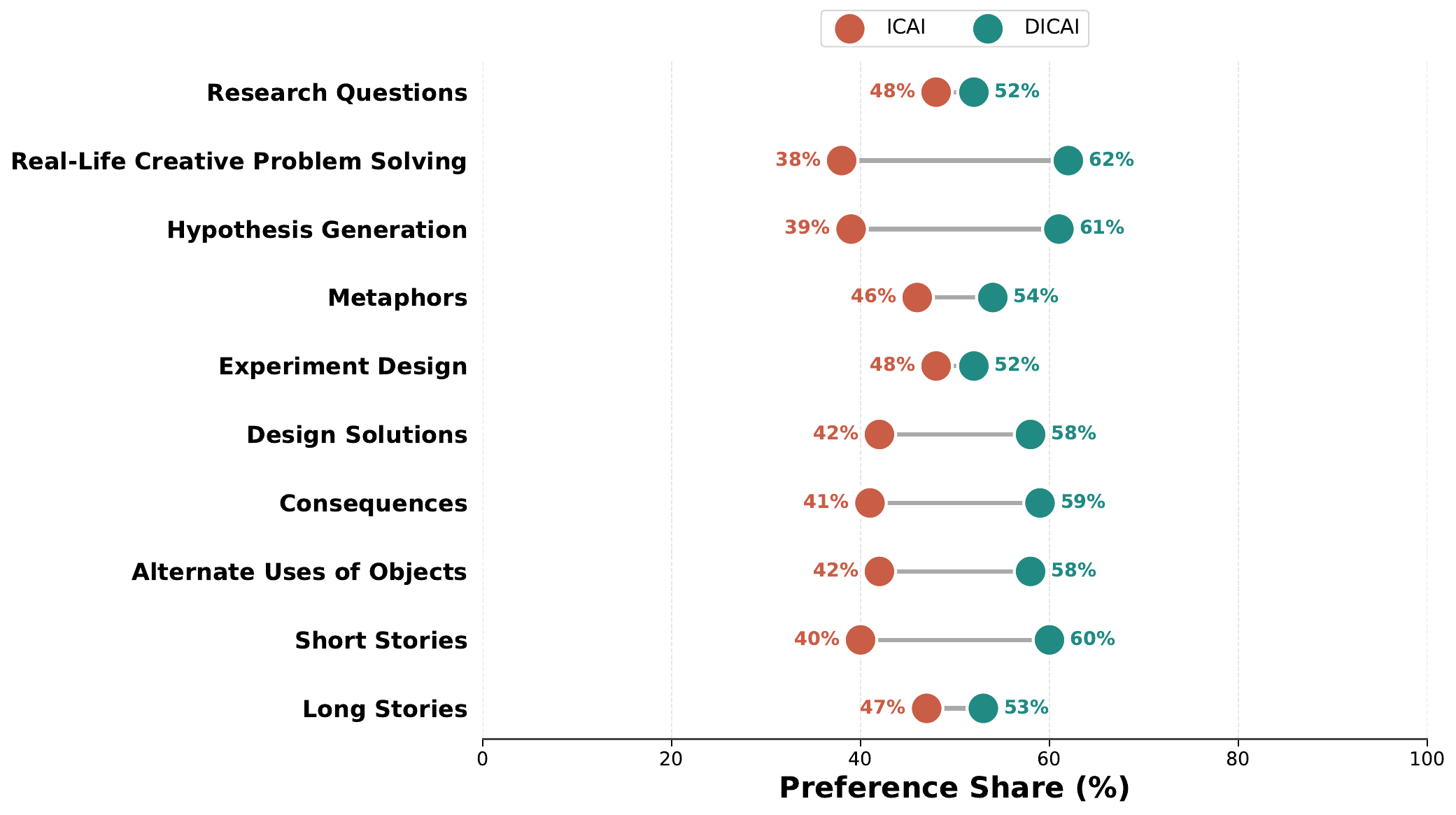}
    \caption{Clarity}
\end{subfigure}
\hfill
\begin{subfigure}{0.32\textwidth}
    \centering
    \includegraphics[width=\linewidth]{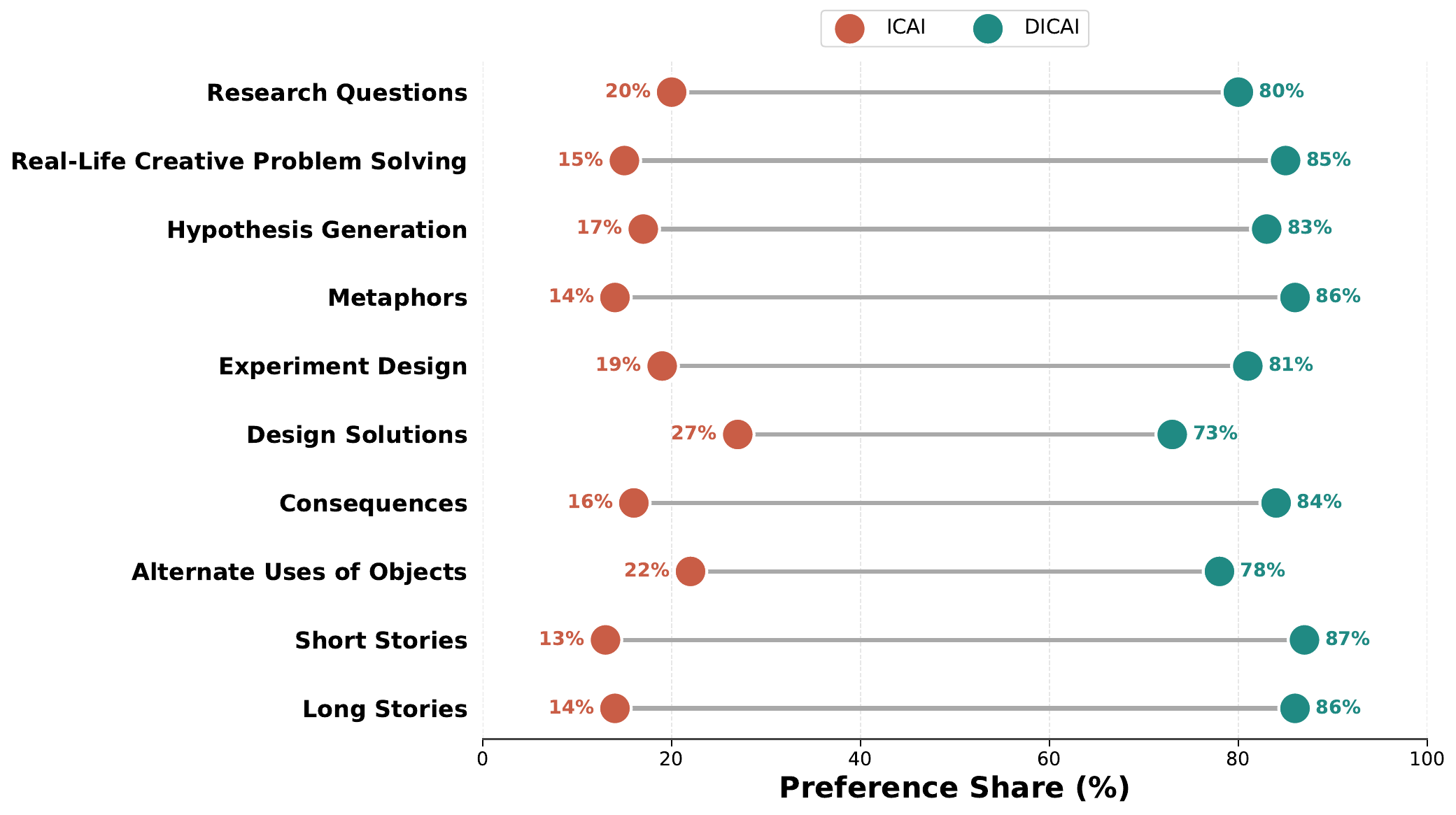}
    \caption{Coherence}
\end{subfigure}

\vspace{0.6em}

\begin{subfigure}{0.32\textwidth}
    \centering
    \includegraphics[width=\linewidth]{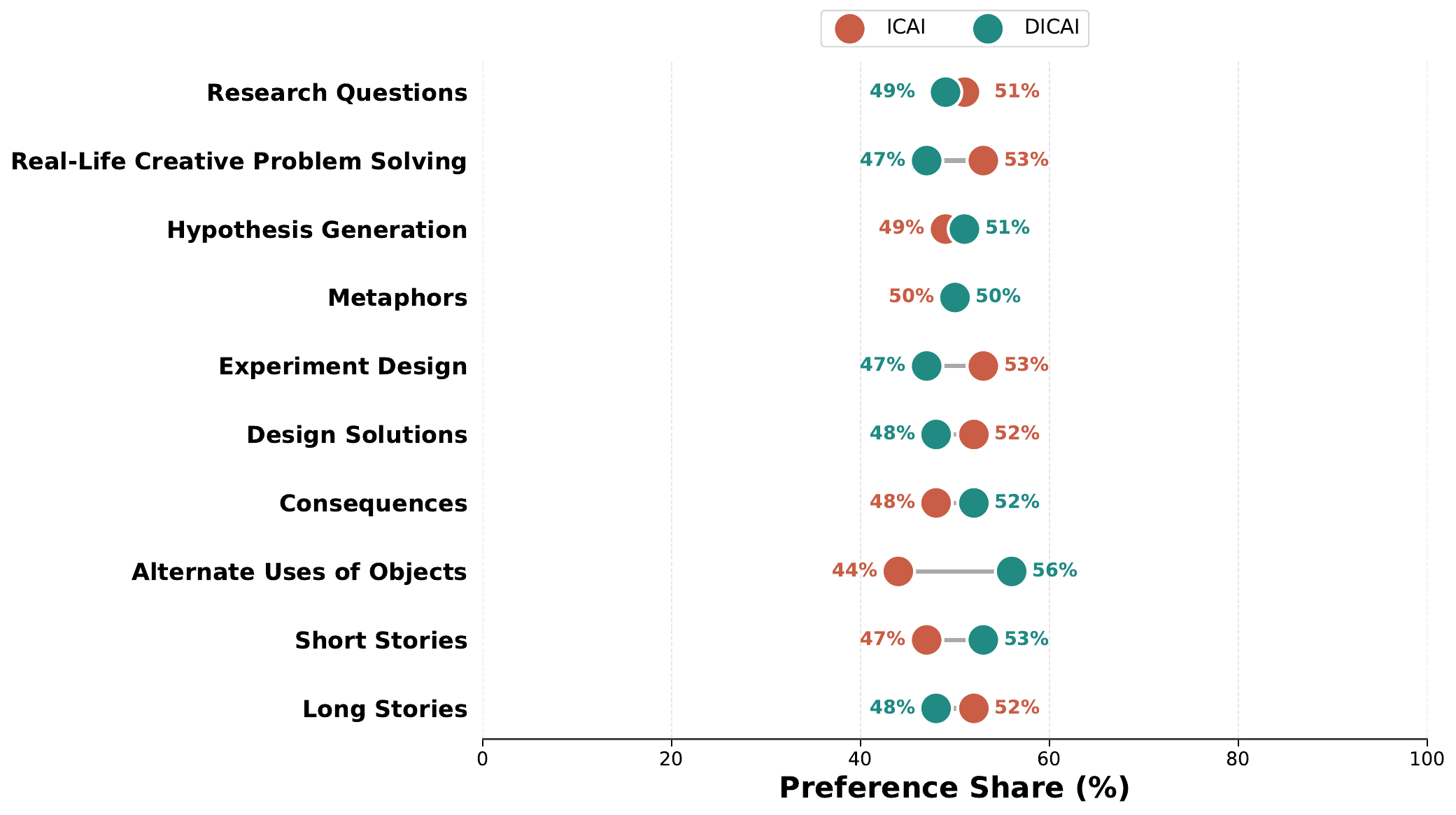}
    \caption{Feasibility}
\end{subfigure}
\hspace{0.06\textwidth}
\begin{subfigure}{0.32\textwidth}
    \centering
    \includegraphics[width=\linewidth]{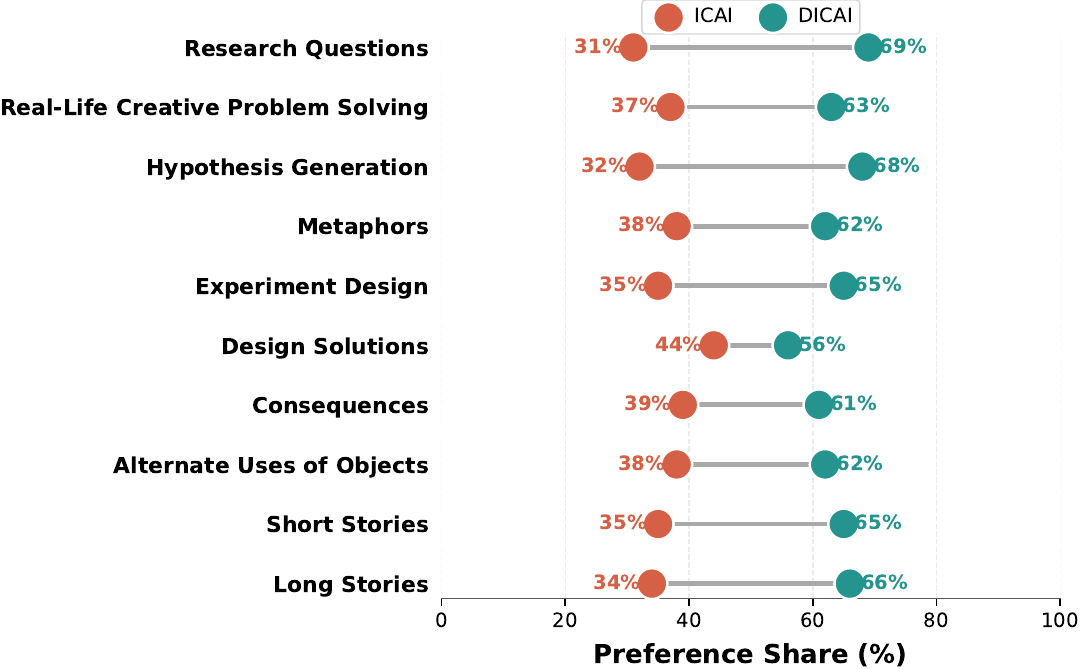}
    \caption{Faithfulness}
\end{subfigure}

\caption{
\textbf{Comparison of Democratic ICAI and ICAI across five dimensions using Qwen2.5-32B.}
Each subplot reports preference shares across ten tasks.
}
\label{fig: constitution_metrics}
\end{figure*}

\section{Results and Analyses}

\begin{enumerate}[noitemsep,nolistsep,leftmargin=*]
\item \textbf{Democratic ICAI produces relatively more informative, interpretable, and robust constitutions:}
Across all tasks with GPT-4o and GPT-5, Democratic ICAI outperforms ICAI and AutoRubric in preference accuracy (refer Tables~\ref{tab:4o_results} and~\ref{tab:5_results}), while remaining competitive with deliberative prompting baselines such as CoT, CoT-SC, ToT, and Self-Refine. Although it does not outperform all baselines on every task, Democratic ICAI achieves the highest average preference accuracy across tasks and exhibits the lowest standard deviation under both models. The gains are most evident on tasks where single-pass explanations often fail to capture or balance competing rationales, such as Research Questions and Real-Life Creative Problem Solving.

These improvements are further supported by an independent Decision Tree judge, where Democratic ICAI achieves higher preference accuracy (see Figure~\ref{fig:dt} in Appendix~\ref{sec: dt_details}), suggesting that the gains are not tied to a specific judge. Finally, LLM judges consistently prefer Democratic ICAI-generated constitutions across qualitative dimensions, including generality, clarity, coherence, and faithfulness, highlighting the interpretability and robustness of the induced principles (see Figure~\ref{fig: constitution_metrics}).

\item \textbf{Cognitively aligned principles enable the construction of diverse constitutions:} Democratic ICAI constructs principles through adversarial debate, where agents justify competing preferences. This mirrors the multi-dimensional reasoning humans employ when evaluating creative outputs, enabling more cognitively aligned principle formation \citep{huang2025dare}. As a result, it produces more diverse constitutions: embeddings of the induced principles, computed using the same text-embedding model used for clustering, show broader dispersion, larger interquartile ranges, and more pronounced outliers compared to ICAI (refer Figure~\ref{fig:constitutional_distance} and Appendix Figure~\ref{fig: constitutional_distance_appendix}).

\begin{table}[t]
\centering
\small
\begin{tabular}{lrrrr}
\toprule
Model & Novelty & Diversity & Surprise & Quality \\
\midrule
CoT         & 0.1059 & 0.3988 & 301.3932 & 14.9636 \\
CoT-SC      & 0.1081 & 0.3997 & 229.2335 & 11.4423 \\
ToT         & 0.1001 & 0.4002 & 266.5760 & 12.9342 \\
Self-Refine & 0.1029 & 0.4067 & \textbf{317.4136} & 14.1948 \\
AutoRubric  & 0.1265 & 0.4332 & 196.9862 & 7.2427 \\
ICAI        & \underline{0.1337} & \textbf{0.4863} &
              \underline{302.4753} & 16.5250 \\
DICAI       & \textbf{0.1351} & \underline{0.4611} &
              221.0350 & \underline{23.8819} \\
CrPO-cre    & 0.1334 & 0.4587 & 187.7784 & \textbf{30.7781} \\
\bottomrule
\end{tabular}
\caption{Creative-generation results for constitution-aligned model evaluation using Llama-3.1-8B-Instruct as a base model. The best and second-best results for each metric are shown in bold and underlined, respectively.}
\label{tab:dpo_llama_results}
\end{table}

This diversity reflects richer conceptual coverage and the emergence of distinctive principles beyond the dominant semantic cluster, improving generalization (refer Figure~\ref{fig: constitution_metrics} and Appendix Figure~\ref{fig:constitution_metrics_combined}).

\item \textbf{Democratic ICAI achieves relatively less lossy compression:}
 Democratic ICAI exhibits reduced lossy compression by inducing a more diverse and semantically broad set of principles. The optimal cluster count, $M^*$, is the value at which the silhouette score is maximized. Since $M^*$ is higher for Democratic ICAI than for ICAI in 8 of 10 tasks, its principles require more distinct conceptual groups to represent their semantic structure. Its higher peak silhouette values further indicate that these groups are well formed rather than artefacts of over-clustering (see Table~\ref{tab:silhouette_optimal_clusters} in Appendix~\ref{app:ablation_studies}).
This distinction is evident in Figure~\ref{fig:constitution_redundancy}, where Democratic ICAI’s principles span a wider range of mechanisms, while ICAI’s are more overlapping and narrowly focused. Additional examples are provided in Appendix~\ref{app:example_constitutions}, and case-level comparisons are provided in Table~\ref{tab:case_level_comparison} in Section~\ref{sec:case_level_constitution_analysis} of the Appendix. Consistent with this, LLM annotators rate Democratic ICAI higher in clarity, indicating that increased semantic diversity translates into more interpretable and better-aligned principles.
 \item \textbf{Complex decision-making processes are harder to compress:} Democratic ICAI induces constitutions that are more diverse (refer Figure~\ref{fig:constitutional_distance}), containing both broad evaluative principles (e.g., holistic narrative quality and emotional impact) and fine‑grained mechanisms (e.g., character‑driven causality, realism of reactions, and layered subtext). This enables the constitution to represent the complexity of human decision-making processes, where judgments emerge from both high-level goals and detailed narrative cues. While the increased granularity can reduce feasibility compared to simpler heuristics as can be seen in Figure~\ref{fig: constitution_metrics}, it improves the ability to capture overall quality, as reflected in LLM-based evaluations. In contrast, ICAI’s compact principles emphasize simplicity at the cost of expressive fidelity.

\item \textbf{Induced principles show limited high-severity bias and spuriousness under external audits.:} Audits of the induced constitutions, using Qwen2.5-32B and Llama-3.1-70B as external auditors, report no high-severity flags on either the demographic-bias or spurious-criteria axis (refer Tables~\ref{tab:bias_spurious_flags} and \ref{tab:bias_llama_audit} in Appendix~\ref{sec:audit_prompt_examples}). The rare medium-severity flags correspond to protective guardrails (e.g., avoiding culturally marked framing) rather than encoded stereotypes, while low-severity spurious flags reflect creative-writing trade-offs such as tonal or stylistic preferences that genuinely correlate with quality (refer Table~\ref{tab:audit_examples} in Appendix~\ref{sec:audit_prompt_examples}). Because the principles are human-readable, they can be inspected and edited before deployment, unlike black-box reward models.

\item \textbf{Both adversarial debate and persona diversity are necessary for the observed gains:} Ablations isolate the contribution of each component, including persona count and persona design (refer Tables~\ref{tab:ablation_debate_rounds}, \ref{tab:ablation_mad}, and~\ref{tab:ablation_persona_design} in Appendix~\ref{app:ablation_studies}). Removing the debate stage and extracting principles directly from rationales reduces performance to the level of ICAI, even when using multiple personas, while a single persona without debate performs even worse. In contrast, the full Democratic ICAI pipeline outperforms these variants, indicating that gains arise from the combination of adversarial debate and persona diversity rather than either component alone.
Within the debate stage, most improvements emerge early: even a single round substantially improves over ICAI, with diminishing returns in later rounds. This suggests that Democratic ICAI achieves strong performance without requiring extended multi-round debate, maintaining computational efficiency.

\item \textbf{Democratic ICAI-derived constitutions provide stronger supervision for creative generation:} Across Llama and Mistral (Table~\ref{tab:dpo_llama_results}, Table~\ref{tab:dpo_mistral_results} in Appendix~\ref{app:training_details}), models trained on Democratic ICAI-induced preferences show strong creative-generation performance. On Llama-3.1-8B-Instruct, Democratic ICAI achieves the best novelty score and the second-best diversity and quality scores. On Mistral-7B-Instruct, Democratic ICAI obtains the second-best novelty, diversity, and quality scores, while substantially improving quality over ICAI. Although ICAI achieves higher surprise and CrPO-cre the highest quality, Democratic ICAI provides the strongest overall constitution-derived supervision, suggesting that debate-derived principles favor coherent and meaningful creativity over raw unpredictability. Table~\ref{tab:choice_poem_examples} in Appendix~\ref{app:training_details} provides qualitative examples from the trained Llama model variants. Overall, Democratic ICAI's constitutions transfer beyond preference reconstruction and provide useful supervision for downstream model alignment.

\end{enumerate}

\section{Conclusion}
Democratic ICAI models the reasoning underlying human preferences through multi-persona deliberation, adversarial refinement, and abstraction. It produces diverse, expressive constitutions aligned with human judgment, achieving higher performance and lower variance than ICAI and AutoRubric. Gains are evident in multi-criteria tasks, where single-pass methods underperform. It also improves model alignment by generating better preference labels that capture multi-criteria judgments. Overall, Democratic ICAI better captures the evaluative structure of complex human decisions without relying on a dominant heuristic.

\section{Limitations}
Explicit ground-truth principles underlying human judgments are often unavailable, making it difficult to determine whether learned principles are fully accurate or complete. As a result, evaluation relies on indirect proxy measures rather than direct validation against human intent. Moreover, preference data may encode systematic biases, which can propagate into induced principles and influence downstream behavior. Democratic ICAI mitigates some of these risks by producing human-readable principles that support inspection and iterative refinement. However, this transparency does not eliminate bias, and expert oversight remains necessary to identify subtle or systemic issues and ensure alignment with intended values.


\section{Ethics and Societal Impact}
Our work aims to improve the transparency and usability of preference data, which plays a central role in training and evaluating large language models. By inducing human-readable constitutions, Democratic ICAI offers a more interpretable alternative to black-box reward models, enabling practitioners to understand, inspect, and refine the principles guiding model behavior. This increased transparency can support better accountability and more informed decision-making in downstream applications.

At the same time, the use of preference data raises important ethical considerations. Biases present in underlying datasets can be reflected in induced principles and subsequently influence model behavior. Democratic ICAI partially addresses this challenge by producing explicit, editable principles, allowing practitioners to identify and revise undesirable patterns before deployment. This visibility distinguishes it from approaches where such biases remain hidden within opaque models.

We recommend that generated constitutions be treated as editable artifacts rather than fixed outputs. Careful review is essential to ensure alignment with intended values and to mitigate the risk of reinforcing harmful or unintended biases. When used responsibly, we believe Democratic ICAI can contribute to more interpretable, controllable, and accountable AI systems.

\bibliography{bib/custom}
\clearpage



\appendix

\label{sec:appendix}

\section*{Appendix}
\startcontents[appendix]
\printcontents[appendix]{}{1}{\setcounter{tocdepth}{2}}

\onecolumn

\section{Dataset and Baselines}
\label{sec: additional_details}
ICAI is tested on four datasets \citep{findeis2025icai}. These include synthetic data with known ground-truth principles, AlpacaEval \citep{dubois2024length}, Chatbot Arena conversations \citep{chiang2024chatbot}, and PRISM demographic data \citep{kirk2024prism}. All of these are rooted in instruction-following scenarios like factual questions, coding, and summarization. Preferences in these settings mostly reward correctness, completeness, and fluency.
Our work shifts the focus to creative domains where, judgments are far less predictable. We test our work mainly on two datasets:
\paragraph{LiTBench.}
LiTBench~\citep{fein2025litbench} is a preference benchmark for creative writing evaluation, designed to test whether language models can reliably judge the quality of long-form narrative text. Each instance consists of a short story prompt paired with two candidate completions and a human preference label indicating which completion is preferred. Preferences in LiTBench are derived from number of upvotes the story obtained on the subreddit. We use 1{,}000 pairs for constitution induction and 2{,}000 held-out pairs for evaluation. 
Representative LiTBench story examples are shown in Figures~\ref{litbench_story_a} and~\ref{litbench_story_b}.

\paragraph{MuCE.}
MuCE~\citep{ismayilzada2025creative} is a large-scale human-preference dataset spanning a diverse range of creative tasks, curated from psychological creativity assessments. The full dataset contains over 200{,}000 responses with ratings from more than 30 creativity tasks, of which we evaluate on nine categories that span markedly different notions of creative success: Alternate Uses of Objects, Consequences, Design Solutions, Experiment Design, Hypothesis Generation, Metaphors, Real-Life Creative Problem Solving, Research Questions, and Short Stories. This category-level diversity makes MuCE complementary to LiTBench: while LiTBench focuses on long-form narrative judgment, MuCE enables evaluation of within-task generalization across a diverse set of short-form creative tasks, where different categories privilege different evaluative criteria. For each task we sample 500 unique accepted-rejected response pairs to derive a task-specific constitution and evaluate on the official test split.

 \vspace{\baselineskip}
We choose baselines from two categories to isolate the effect of explicit principle induction. Deliberative prompting baselines such as CoT, CoT-SC, ToT, and Self-Refine \citep{wei2022chainofthought,wang2022selfconsistency,yao2023treeofthoughts,madaan2023selfrefine} test whether gains come from stronger test-time reasoning without inducing a constitution. Principle-based methods such as ICAI and AutoRubric \citep{findeis2025icai,xie2025auto} test against approaches that induce reusable principles or rubrics from preference data. The prompt templates for the deliberative prompting baselines are provided in Appendix~\ref{app:deliberative_prompts}. We do not include GCAI as a primary baseline because it relies on human-written reasons and does not provide a directly reproducible public implementation. GCAI \citep{bell2026beyond} also differs in its inputs and methodology by incorporating human-written reasons and using clustering and summarization to distill principles. Given these differences, we focus on ICAI and AutoRubric as the most directly comparable and reproducible principle-based baselines.

{\tiny
\begin{tabularx}{\textwidth}{YYYY}

\toprule
\thead{Task} & \thead{Prompt} & \thead{Response A} & \thead{Response B} \\
\midrule
\endfirsthead

\toprule
\thead{Task} & \thead{Prompt} & \thead{Response A} & \thead{Response B} \\
\midrule
\endhead

\midrule
\multicolumn{4}{r}{\footnotesize Continued on next page}
\endfoot

\bottomrule
\noalign{\vskip 14pt}
\caption{Examples of tasks from the MuCE dataset \cite{ismayilzada2025creative}. We present 9 unique tasks from the dataset.}
\label{tab:muce_dataset}
\\
\endlastfoot
\tiny
Alternate Uses of Objects &
Come up with an original and creative use for the following object: toothbrush &
brushing dog hair &
As a beam to hold up a little fort. \\

Consequences &
Come up with an original and creative consequence for the following scenario: What would be the result if everyone suddenly lost the sense of balance and were unable to stay in the upright position for more than a moment? &
We would eventually die &
motorcycles would no longer be a usable form of transportation \\

Design Solutions &
Come up with an original and creative solution to make remote learning more engaging and effective. &
using twitch for lectures &
Board game with coding challenges \\

Experiment Design &
Come up with an original and creative experiment to test the following hypothesis or research question: You think students at your school are friendlier than students at most schools. How could you test that hypothesis? &
Have a random group with a variety of demographics take a survey of how their experience is with other students at each school. Two different groups of students enrolled at each school &
distribute a mandatory, randomized survey \\

Hypothesis Generation &
You just started going to a new school. You notice that more students sing in the hallways at this school than at your old school. What hypotheses do you have about why that is? &
this school's tradition &
the kids are in the chorus \\

Metaphors &
Finish the sentence with an original and creative metaphor: The sweet candy is.... &
life is like a box of sweet candy &
horrible \\

Real-Life Creative Problem Solving &
Come up with an original and creative solution for the following real-world problem: Clara, a junior pre-med student, is working part-time and taking a 15 hour credit load at school. Clara enjoys her pre-med courses very much, but they are very difficult and time consuming. Her current job as a 'gopher' at an architectural design firm requires her to work 25 hours a week which really cuts into her available study time. In fact, she is barely getting 'C's' in two of the classes she needs for her major. The pay in her present job is good, and she enjoys the work, although she is not getting a lot of practical experience. Clara does not want to drop any of her classes as she needs them to graduate so that she can be admitted to medical school in the coming year. She also knows that she needs good grades to be admitted to medical school. Up until now, Clara has been able to work at her job and still get good grades, but the difficult courses she is taking now require much more of her time. Clara is not sure how to solve her problem.. &
she should weigh out her priorities. it seems as if money is the only real reason for her staying at her curret job so she should just apply for fafsa or loans and go through school with that money. &
Clara needs to cut down on her hours at work because her grades are more important, in order for her to get into med school. If she needs to work those hours for financial reasons then she needs to not have a full load of classes and only go to school part time. \\

Research Questions &
You created the first vehicle that can reach an unexplored part of the ocean floor. What scientific questions could you ask about this part of the ocean floor? &
is there life down there? &
does this part of the ocean floor receive any light from above? if so, how much? \\

Short Stories (MuCE) &
Come up with an original and creative story which includes the following 3 words, make it around 5 sentences long: year, week, embark. &
Janet was very excited about her upcoming vacation. She would embark on her adventure in one week but had already started to pack. Janet had waited a whole year for travel to Hawaii and couldn't wait to leave. She had even bought a small phrase book so she could talk to the locals in Hawaiian. She almost bought a one-way ticket but remembered her family lived nearby, so she bought a round-trip ticket instead. &
This year, I was very serious about finally going on a diet. I have tried and failed so many times, but after the devastating news I received last week, I knew I had nothing left to lose. I know very well this journey I was about to embark was incredibly challenging for me, but it was something that had to be done. My health and my family were suffering because of me, and I couldn't stand to see them in pain any longer. How did I let this get so out of control? \\
\end{tabularx}
}

\begin{tcolorbox}[
  enhanced jigsaw,
  breakable,
  pad at break=2mm,
  colback=blue!5!white,
  colframe=blue!75!black,
  title=LiTBench Story Example 1,
  rounded corners
]
Mine was not a glorious death, nor one that you'd expect to read about in the history books, and neither was my life. I never really accomplished anything spectacular in my life. I worked my family's land when I was young, and answered my king's call to arms when it came, yet I didn't participate in any particularly memorable battles, or leave a wife and child behind to carry on my legacy. I had a few friends that I met whilst undertaking my military service, but I sincerely doubt that any of them lived much longer than I did, as they were mostly an adventurous lot and were always finding themselves in strange and new places, whilst I played it safe by staying back and tending to the horses. I managed to survive my entire military career without actually drawing my sword in combat, which made me the butt of many jokes amongst my friends. They'd joke and say that my sword was magical and that it must need the strength of a thousand men to pull it from my hilt, because I never actually drew it in battle. Some of the men even jokingly claimed that I must have slain dragons in a previous lifetime and that fighting humans in battle was beneath me; they nicknamed me ``Chief Dragon''.\vspace{\baselineskip}

My best friend wasn't one for the history books, either. My best friend didn't have a real vocation, having been raised by people subsisting on berries and roots out in the forest lands, so his social skills (or lack thereof) left people with the belief that he was quite mad. He was a good friend, and equally uninterested in joining combat, but he came along with me on sorties, just to keep me company (and, I suspect, as an excuse to forage for new herbs to smoke). Because of the cold winters, and the fact that he was completely bereft of either title or military rank, he was not afforded a suit of armour or even a horse, so he wore a body-length cloak that he had fashioned out of some decrepid curtains we found in an abandoned village, and used a long stick as a walking aide. I've missed him, I wonder what happened to him after I died.\vspace{\baselineskip}

I heard stories of purgatory and the afterlife from my grandmother, and oftentimes heard parables read out to us by the priest at our church, but I never really took much notice of any of it. I don't really know where I am, but I know that in this place, those whose bodies have died, come here to share their stories. I often meet with people I've heard stories about from the others, with weird and fascinating names, from strange and unbelievable lands, but for the most part, nobody really hangs around long enough for me to even remember them anymore. There are countless millions of people here, but of those who came before me, I could fit their entirety into a simple country chapel.\vspace{\baselineskip}

I had completely given up with bothering to get to know any of the newcomers, until a man dressed in the most peculiar outfit I'd ever seen, came charging up to me with a spirit of determination that I'd not seen since I was in my physical body.\vspace{\baselineskip}

``Is it really you?'' he asked, with a confused expression on his face.\vspace{\baselineskip}

``I don't know who you think I am, I was just a farmer and a simple squire when I was alive.'' I responded, utterly confused about why he was interested in me.\vspace{\baselineskip}

``I've heard stories about you, and saw the murals they painted in the chapels, I've even seen the tomb where you were buried! You're my hero, your final quest is legend! Can you tell me, did you actually find it? Where did you bury it?'' the stranger excitingly blabbled.\vspace{\baselineskip}

\end{tcolorbox}
\begin{center}
\captionof{figure}{Example of a story from LiTBench.}
\label{litbench_story_a}
\end{center}

\begin{tcolorbox}[
  enhanced jigsaw,
  breakable,
  colback=blue!5!white,
  colframe=blue!75!black,
  title=LiTBench Story Example 2,
  rounded corners
]
Eternity is boring. You know how when you have to sit through something, or you don't want to be somewhere, time seems to stretch out and elongate? Eternity is like that -- one stretching, elongated moment that never ends.\vspace{\baselineskip}

The company was enough at first. It was something else to mingle with all these entities who accomplished so much in their physical lives. To hear about new stories and new events that led to these individual's triumphant or infamous transcendence.\vspace{\baselineskip}

But after a few thousand years hearing about the actions that led to infamy, to posterity, to being remembered, all of that gets old and dull. It takes on a bland tastelessness. With time everything turns to dust -- everything regresses into this one saturated moment.\vspace{\baselineskip}

There has always been a constant nagging, a differentiation, a uniqueness that grew inside me precisely because it was an unknown, and not knowing breeds uncertainty, and uncertainty punctuates the monotony.\vspace{\baselineskip}

That's over now. It turns out that I am the face of the nameless, the individual incarnation of the myriad masses that lived and died before, during and after my short life on Earth. My little hovel is preserved as some sort of monument to the ancestors on which present-day civilization stands. My one bowl sits on my small table, both encased in a glass box.\vspace{\baselineskip}

I am the personalized incarnation of all who come before, and it is only by standing on my shoulders that humanity will one day reach the stars.\vspace{\baselineskip}

But me? I now know why I'm here. The unknown has become the known and all the wondering it wrought is already being subsumed by
\end{tcolorbox}
\begin{center}
\captionof{figure}{Example of a story from LiTBench.}
\label{litbench_story_b}
\end{center}

\section{Training Details}
\label{app:training_details}
\subsection{Task-wise Expert Persona Committee}
The task-wise persona committees used during reasoning assembly (Section~\ref{sec: reasoning}) are listed below. Example persona agent prompts are provided in Appendix~\ref{app:reasoning_personas}.
\begingroup
\begin{center}
\centering
\small
\begin{tabularx}{\linewidth}{>{\raggedright\arraybackslash}p{5cm} X}
\toprule
\textbf{Task} & \textbf{Committee Experts} \\
\midrule
Real-Life Creative Problem Solving & Innovation Consultant / Design Thinking Expert; Organizational Psychologist; Business Strategy Expert \\

Metaphors & Cognitive Linguist; Poet / Creative Writer; Professor of Literature \\

Alternate Uses of Objects & Creativity Researcher (Divergent Thinking Specialist); Product / Industrial Designer; Cognitive Psychologist \\

Design Solutions & UX / Product Designer; Design Engineer; Innovation Consultant / Design Thinking Expert \\

Short Stories (MuCE) \& Long Stories (LiTBench) & Screenwriter; Professor of Literature; Literary Critic \\

Consequences & Futurist / Scenario Planner; Cognitive Psychologist (Counterfactual Thinking); Science Fiction Writer \\

Experiment Design & Experimental Scientist; Research Methods / Statistics Professor; Principal Investigator (PI) \\

Hypothesis Generation & Domain Scientist; Theory / Conceptual Framework Researcher; Science Writer / Scientific Communicator \\

Research Questions & Principal Investigator (PI); Journal Editor or Academic Reviewer; Research Methods Professor \\
\bottomrule
\end{tabularx}
\addtocounter{table}{-1}
\renewcommand{\theHtable}{persona-table}
\refstepcounter{table}
\label{tab:persona_table}
\par\smallskip
Table~\thetable: Task-wise assignment of committee experts used for reasoning.
\addcontentsline{lot}{table}{\protect\numberline{\thetable}{Task-wise assignment of committee experts used for reasoning.}}
\end{center}
\endgroup

\subsection{Constitution-Aligned Model Training}
\label{app:constitution_dpo_results}

\subsubsection{Training Setup}

\paragraph{Preference data.}
Let $\mathcal{M}=\{\mathrm{CoT}, \mathrm{CoT\text{-}SC}, \mathrm{ToT}, \mathrm{Self\text{-}Refine}, \mathrm{AutoRubric}, \mathrm{ICAI}, \mathrm{DICAI}\}$ denote the set of preference-label sources used for DPO training. For methods that produce an explicit constitution, $m \in \{\mathrm{AutoRubric}, \mathrm{ICAI}, \mathrm{DICAI}\}$, let $\mathcal{K}^m$ denote the induced constitution. Starting from MuCE preference pairs, we randomly swap the accepted and rejected responses to reduce position bias. An LLM judge is then given the task prompt, the two responses, and the constitution $\mathcal{K}^m$, and is asked to select the response that better follows the listed principles. The constitution-guided judge prompt is provided in Figure~\ref{ICAI_prompt} in Appendix~\ref{app:dicai_prompts}. For prompt-only baselines, we instead use their corresponding prompting procedures \citep{wei2022chainofthought,wang2022selfconsistency,yao2023treeofthoughts,madaan2023selfrefine} to generate preference labels, with prompt templates provided in Appendix~\ref{app:deliberative_prompts}. For our experiments, we use GPT-4o as our LLM judge. We treat the selected response as preferred and the other response as rejected, producing a method-induced preference dataset:
\[
D_{\mathrm{pref}}^m =
\{(x_i, y_i^{m,+}, y_i^{m,-})\}_{i=1}^{N},
\]
where $m \in \mathcal{M}$ denotes the label source used to construct the training pairs.

\paragraph{DPO objective and configuration.}
For each base model (Llama-3.1-8B-Instruct and Mistral-7B-Instruct), we train one LoRA-based DPO model per preference-label source. All runs use the same rank, target modules, learning rate, batch size, gradient accumulation, and DPO $\beta$. The only difference is the source of the preference labels (Table~\ref{tab:constitution_dpo_config}). We optimize with a CrPO-augmented DPO loss, which extends standard DPO \citep{rafailov2023dpo} with creativity-oriented objectives \citep{ismayilzada2025creative}.

\paragraph{Creative Preference Optimization (CrPO).}
\label{sec:crpo_details}
Creative Preference Optimization (CrPO) \citep{ismayilzada2025creative} adapts preference optimization to creative-generation tasks by combining preference learning with creativity-oriented objectives. Instead of optimizing only for which response is preferred, CrPO additionally evaluates generated outputs along four creative dimensions: \textbf{novelty}, \textbf{diversity}, \textbf{surprise}, and \textbf{quality}. Novelty measures how original a response is relative to typical outputs, diversity measures variation across generated responses, surprise captures unexpectedness, and quality measures whether the output remains coherent and useful. In our setting, CrPO is used both as the DPO-style training objective and as the evaluation protocol for trained models. We keep the CrPO setup fixed across all label sources so that differences in model behavior can be attributed to the preference labels induced by each method.

We set all four CrPO creativity weights to 1.0 during training. We also report \textbf{CrPO-cre}, the creativity-optimized CrPO variant from \citet{ismayilzada2025creative}, as an additional reference baseline in the model-training results. For evaluation, each trained model is run on the same held-out MuCE prompts with the same decoding setup and scoring pipeline. Results for Llama-3.1-8B-Instruct and Mistral-7B-Instruct are reported in Tables~\ref{tab:dpo_llama_results} and~\ref{tab:dpo_mistral_results}, respectively.

\begin{center}
\small
\begin{tabular}{ll}
\toprule
\textbf{Parameter} & \textbf{Value} \\
\midrule
Base models & Llama-3.1-8B-Instruct; Mistral-7B-Instruct \\
Learning rate & 5e-6 \\
rank & 128 \\
alpha & 256 \\
Gradient accumulation steps & 8 \\
Batch size & 4 per device \\
Number of epochs & 2 \\
DPO $\beta$ & 0.1 \\
Lambda diversity & 1.0 \\
Lambda novelty & 1.0 \\
Lambda surprise & 1.0 \\
Lambda quality & 1.0 \\ 
\bottomrule
\end{tabular}
\captionof{table}{Direct Preference Optimization (DPO) configuration}
\label{tab:constitution_dpo_config}
\end{center}

\subsubsection{Additional DPO Results and Qualitative Generation Examples}

\begin{center}
\centering
\small
\begin{tabular}{lrrrr}
\toprule
Model & Novelty & Diversity & Surprise & Quality \\
\midrule
CoT         & 0.1075 & 0.4381 & 354.8670 & 6.0387 \\
CoT-SC      & 0.1047 & 0.4351 & 390.0637 & -5.8568 \\
ToT         & 0.1034 & 0.4380 & 384.5814 & 1.6715 \\
Self-Refine & 0.1036 & 0.4442 & 321.5030 & 5.9554 \\
AutoRubric  & 0.0997 & 0.4526 & \underline{597.6138} & -14.0034 \\
ICAI        & \textbf{0.1227} & \textbf{0.5003} &
              \textbf{646.4740} & -3.5867 \\
DICAI       & \underline{0.1201} & \underline{0.4731} &
              313.3568 & \underline{14.3203} \\
CrPO-cre    & 0.1055 & 0.4230 & 364.3953 & \textbf{16.8745} \\
\bottomrule
\end{tabular}
\captionof{table}{Creative-generation results for constitution-aligned model evaluation using Mistral-7B-Instruct as a base model. The best and second-best results for each metric are shown in bold and underlined, respectively.}
\label{tab:dpo_mistral_results}
\end{center}

{\small 
\begin{tabularx}{\textwidth}{@{}YY@{}} 
\caption{Qualitative comparison of responses generated using Llama-3.1-8B-Instruct as the base model for the prompt: ``Come up with an original and creative poem about the following concept: choice.''}
\label{tab:choice_poem_examples}
\\
  \toprule \textbf{Model} & \textbf{Response} 
  \\ 
  \midrule 
  \endfirsthead 
  \toprule 
  \textbf{Model} & \textbf{Response} 
  \\ 
  \midrule 
  \endhead 
  \midrule
  \multicolumn{2}{r@{}}{\footnotesize Continued on next page}
  \endfoot

  \bottomrule 
  \endlastfoot

CoT
&
We stand, and look both ways\newline
One path straight, and one astray\newline
We know both, yet both untried\newline
We breathe, and our hearts divide

\par\medskip

Decisions made, in a single bound\newline
Our path, forever turned around\newline
No going back, no turning around\newline
Our hearts sing, our souls resound
\\ \midrule
CoT-SC
&
Path ahead, multiple roads\newline
Each choice taken, where does it go?\newline
Two roads diverge, where we roam\newline
And the life we make, the decisions we own

\par\medskip

Some paths lead to peace and serenity\newline
Others lead us through dark melancholy\newline
Each one a choice, each one a way\newline
Where will you turn, today?
\\ \midrule
ToT
&
Two paths lay before me\newline
One with light, one in the night\newline
Each calls my name, beckons me\newline
To decide, to make it right

\par\medskip

In the light, I see what's true\newline
A world of hope, of dreams anew\newline
But in the night, I see my fears\newline
And doubts my heart and brings my tears

\par\medskip

Which path to take? I must decide\newline
What is the choice that will be my guide?\newline
Will I take the light, and all its might\newline
Or will I falter, and take the night?
\\ \midrule
Self-Refine
&
A path stretches far and wide\newline
Two roads diverge, side by side\newline
The choice is mine to make\newline
Life's journey for my sake

\par\medskip

One road is bright and bold\newline
The future called, all to unfold\newline
The other dark and unexplored\newline
A path that none before have soared

\par\medskip

Which road to take? The choice is mine\newline
The journey that the heart will design\newline
Will I follow what others say?\newline
Or forge my own path, in a new way?
\\ \midrule
AutoRubric
&
Two roads lay before me,\newline
One path leads to the unknown,\newline
The other leads to where I've been,\newline
Which way to go?

\par\medskip

Do I follow what I know,\newline
Or do I dare to see?\newline
The choice is mine, I decide,\newline
And the road I choose will be my guide.
\\[0.8em]

ICAI
&
in the store of my mind so grand\newline
a display of options all planned\newline
do I wear the hat of yesterday\newline
or try a new one and see how it stays?

\par\medskip

do I eat the food that i know so well\newline
or try something new that my taste buds can tell?\newline
do I take the road that is familiar and true\newline
or take a chance and see what the world can do?

\par\medskip

the choices we make each and every day\newline
can determine the path we are on our way\newline
to success or failure we all must decide\newline
but which one is right and which one to hide?

\par\medskip

i wish i knew what the future holds\newline
so i can make the right choices to mold\newline
my life into what i want it to be\newline
but until that day i will just have to see\newline
what choices i make and how they turn out\newline
and then i will know if they were right or not to doubt
\\ \midrule
DICAI
&
Do you see the path I take?\newline
Or the one that I will make?\newline
Do I follow what I see?\newline
Or do I make my own destiny?

\par\medskip

What if I chose the road less traveled?\newline
Would I find a path of my own?\newline
Or would I be lost and never found?\newline
Or would I find a new way around?
\\

	\end{tabularx} }

\FloatBarrier

\section{Constitution-Guided Inference with a Decision Tree Judge}
\label{sec: dt_details}

As introduced in Section~\ref{sec: judge}, we operationalize the extracted principles as a structured and interpretable preference predictor, following the broader use of decision-tree-style policies for interpretable LLM-assisted decision making \citep{xiong2024gptree}. For each constitution principle $\tilde{Q}_j \in \mathcal{K}$, we define a corresponding induced feature $f_j \in \mathcal{F}$ that captures the same evaluative dimension, and define for each $f_j$ a feature-specific scoring rubric on a 1--5 scale.
\paragraph{Feature table construction.}
To train a predictor, we construct a tabular dataset where each row corresponds to a response pair $(A_i, B_i)$ and each column corresponds to an induced principle-derived feature $f_j \in \mathcal{F}$, with $f_j$ derived from $\tilde{Q}_j \in \mathcal{K}$. 
For each pair $(A_i, B_i)$ and each feature $f_j$, we prompt an LLM to score Sample~$A_i$ and Sample~$B_i$ \emph{independently} using the 1--5 rubric associated with $f_j$, yielding scores
\begin{equation}
(s_{ij}^A, s_{ij}^B) \in \{1,\ldots,5\}^2 
\end{equation}

We convert these scores into a single feature-level outcome:
\begin{equation}
x_{ij} =
\begin{cases}
\texttt{A}, & \text{if } s_{ij}^A > s_{ij}^B, \\
\texttt{B}, & \text{if } s_{ij}^B > s_{ij}^A, \\
\texttt{A} \text{ or } \texttt{B}, & \text{if } s_{ij}^A = s_{ij}^B \text{ (random tie-break)} 
\end{cases}
\label{eq:feature-level-outcome}
\end{equation}

Collectively, each pair is represented as a feature vector
\begin{equation}
\mathbf{x}_i = (x_{i1}, x_{i2}, \ldots, x_{iM}),
\end{equation}
where $M = |\mathcal{K}|$ is the number of abstract principles in the constitution.

The supervision label for each row is the human preference:
\begin{equation}
y_i \in \{\texttt{A}, \texttt{B}\}
\end{equation}

The prompt for feature table construction is presented in Figure~\ref{table_construction} in Appendix~\ref{app:dicai_prompts}.

\textbf{Training an interpretable decision policy.}
Given the resulting dataset $\{(\mathbf{x}_i, y_i)\}_{i=1}^{N}$, we train a decision tree classifier to predict $y_i$ from the principle-derived feature vector $\mathbf{x}_i$. 
The learned tree defines an interpretable decision policy: each internal node queries a specific principle feature $f_j$, and each leaf node outputs a predicted preference in $\{\texttt{A}, \texttt{B}\}$.

\textbf{Test-time execution.}
At inference time, the trained (and optionally pruned) decision tree is executed as an adaptive evaluation procedure.
Starting from the root node, each visited internal node requests the value of a single feature $f_j$. 
We obtain this value by scoring $(A_i, B_i)$ independently on the associated 1--5 rubric, producing $(s_{ij}^A, s_{ij}^B)$ and converting them into $x_{ij}$ as in Eq.~\ref{eq:feature-level-outcome}.
The process continues along the corresponding branch until a leaf is reached, which outputs the final predicted preference.

\begin{center}
    \includegraphics[width=0.9\linewidth]{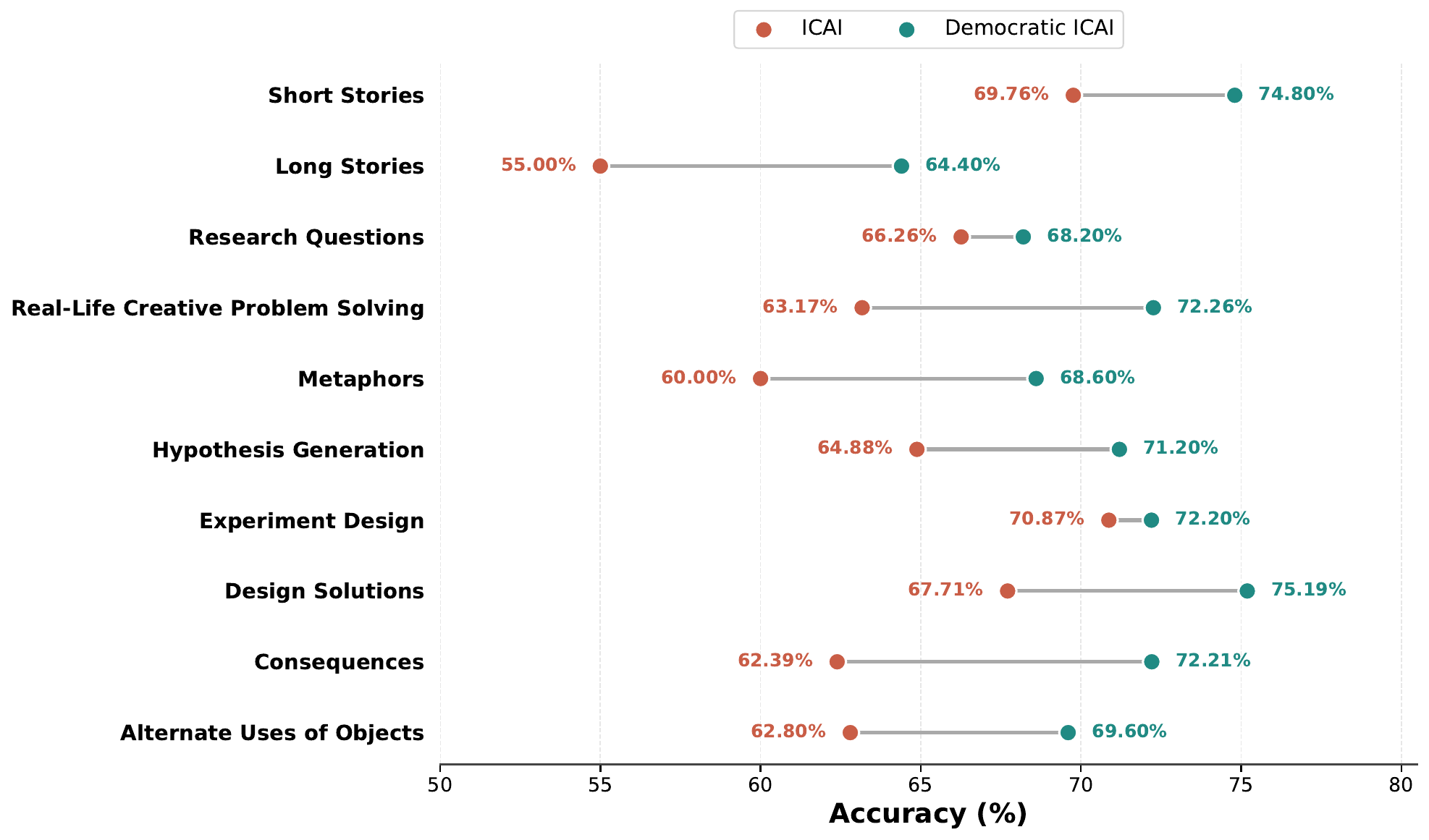}
    \captionof{figure}{Preference accuracy comparison between Democratic ICAI and ICAI with Decision Tree Judge (GPT-4o).}
    \label{fig:dt}
\end{center}

\section{Constitution Quality and Diversity Analysis}
\label{app:constitution_quality_diversity}
\subsection{Constitution Quality Dimensions}
\label{sec: const_qual_appendix}
We directly evaluate the quality of the induced constitutions by comparing ICAI and Democratic ICAI outputs along five dimensions drawn from political philosophy, constitutional theory, and AI alignment: generality, clarity, coherence, feasibility, and faithfulness \citep{fuller1969morality,nwokora2022feasibility,hedling2023faithfulness}. An LLM judge is given both constitutions and asked to select which one better satisfies each dimension. We report the preference share for each dimension, defined as the percentage of pairwise comparisons in which the evaluator prefers one constitution over the other. To reduce circularity and model-specific bias, we run this evaluation with Qwen2.5-32B, GPT-4o, and GPT-5. The annotator prompt is provided in Figure~\ref{consitution_analysis} in Appendix~\ref{app:dicai_prompts}.

\begin{itemize}[noitemsep,nolistsep,leftmargin=*]
  \item \textbf{Generality}: The extent to which a principle applies broadly across diverse contexts and scenarios, rather than being tied to specific examples or dataset artifacts.
  \item \textbf{Clarity}: The degree to which the wording and intent of the principle are interpretable, precise, and unambiguous for AI systems.
  \item \textbf{Coherence}: Whether the principles within a constitution can operate simultaneously without contradictions, forming a logically consistent evaluative framework.
  \item \textbf{Feasibility}: The extent to which an AI system can realistically apply the principle when making decisions in practical evaluation settings.
  \item \textbf{Faithfulness}: The degree to which the principle preserves its intended meaning and resists distortion, misinterpretation, or application inconsistent with its original intent.
\end{itemize}

\vspace{1em}

\begin{figure*}[!tbh]
\centering

\begin{subfigure}{0.40\textwidth}
    \centering
    \includegraphics[width=\linewidth]{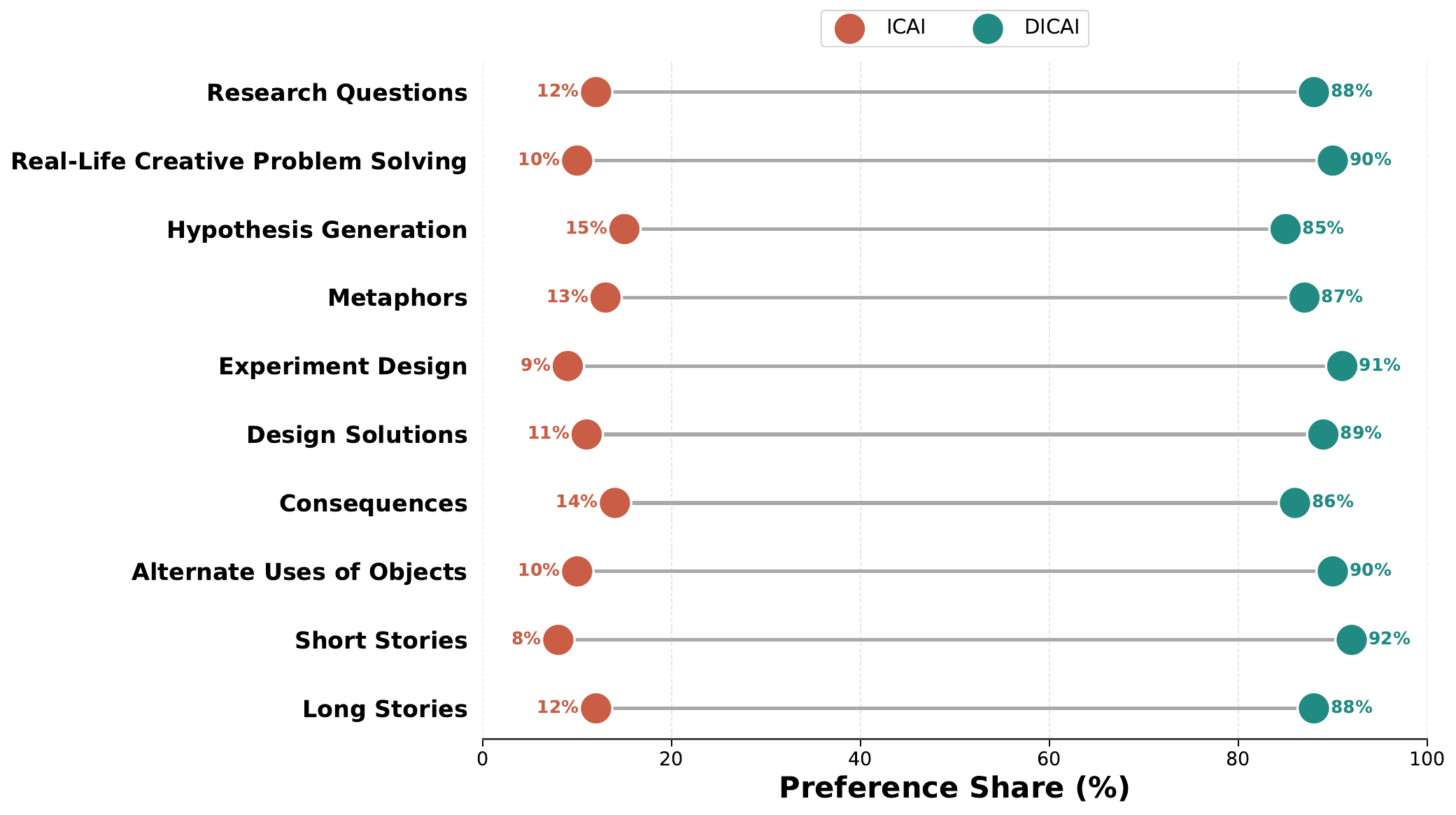}
    \caption{GPT-4o Generality}
\end{subfigure}
\hfill
\begin{subfigure}{0.40\textwidth}
    \centering
    \includegraphics[width=\linewidth]{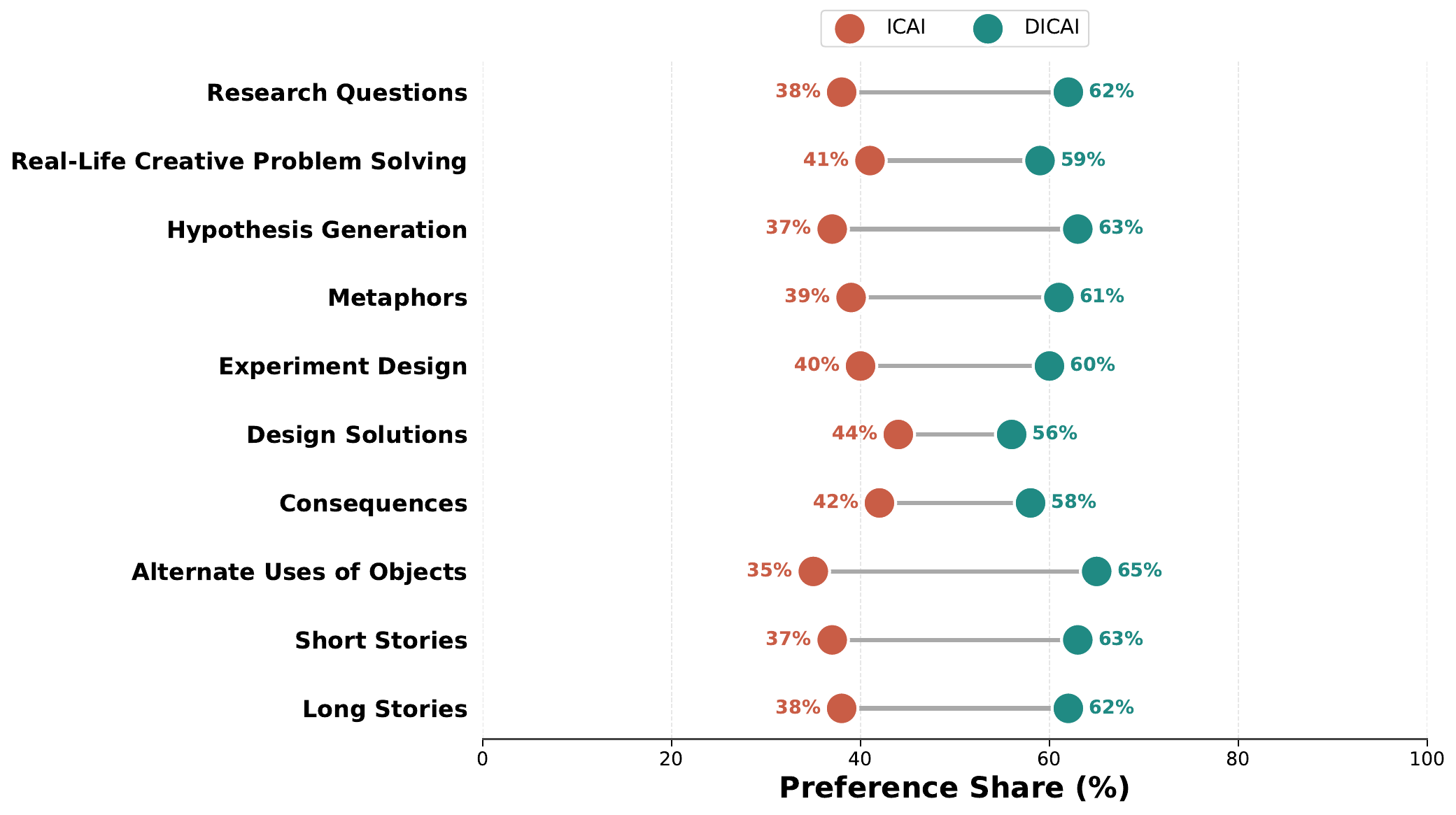}
    \caption{GPT-5 Generality}
\end{subfigure}

\vspace{0.4em}

\begin{subfigure}{0.40\textwidth}
    \centering
    \includegraphics[width=\linewidth]{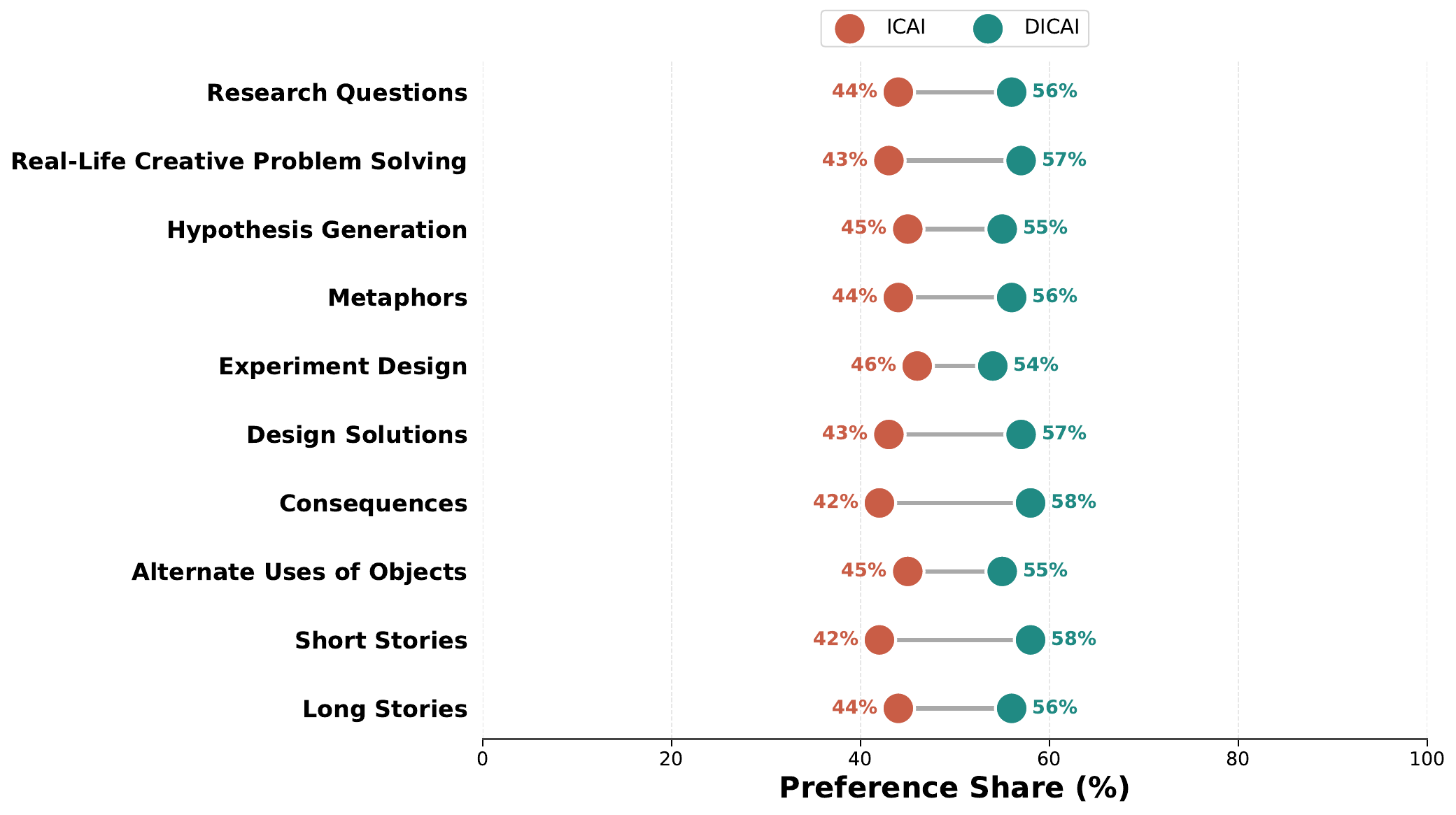}
    \caption{GPT-4o Clarity}
\end{subfigure}
\hfill
\begin{subfigure}{0.40\textwidth}
    \centering
    \includegraphics[width=\linewidth]{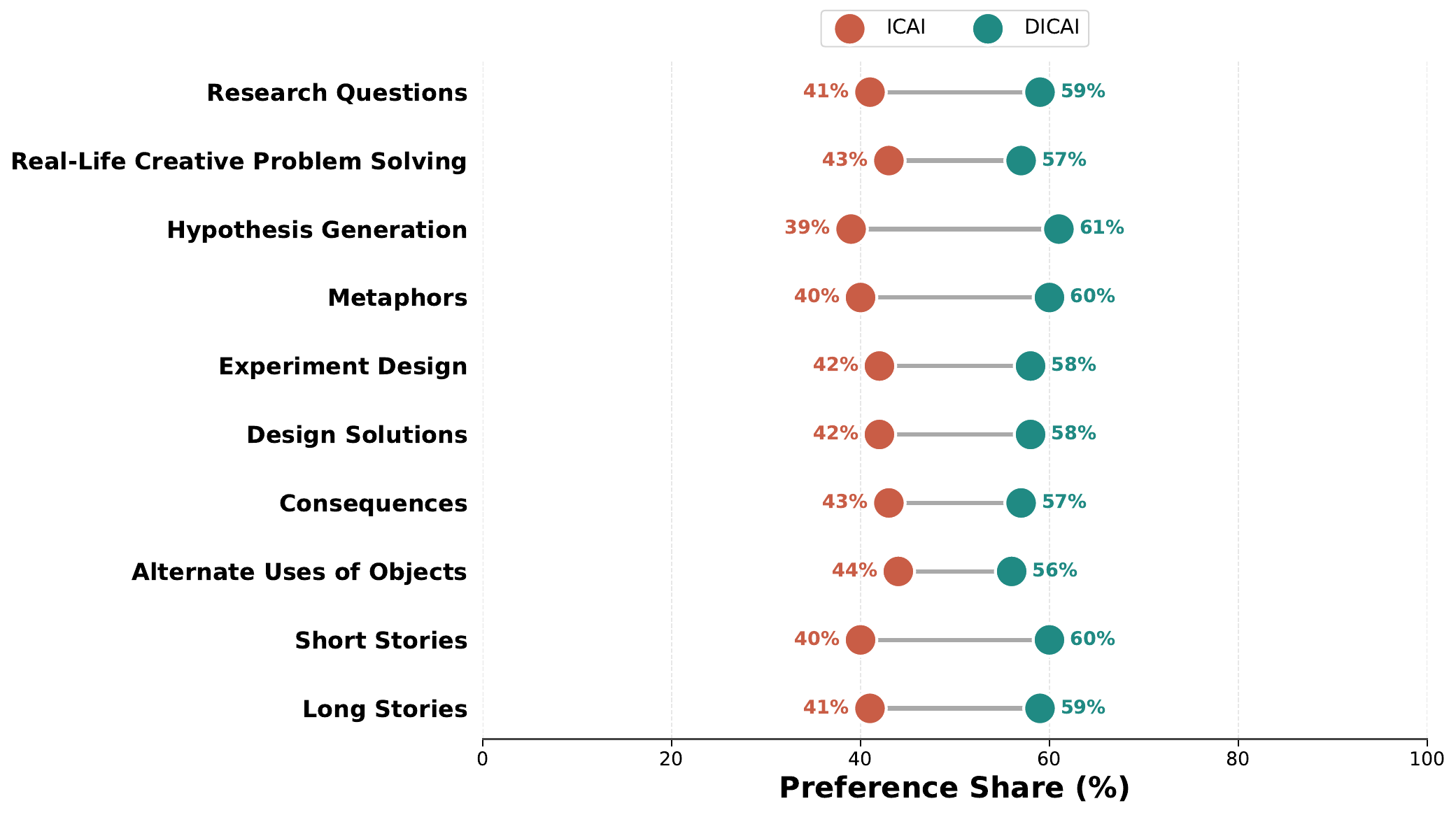}
    \caption{GPT-5 Clarity}
\end{subfigure}

\vspace{0.4em}

\begin{subfigure}{0.40\textwidth}
    \centering
    \includegraphics[width=\linewidth]{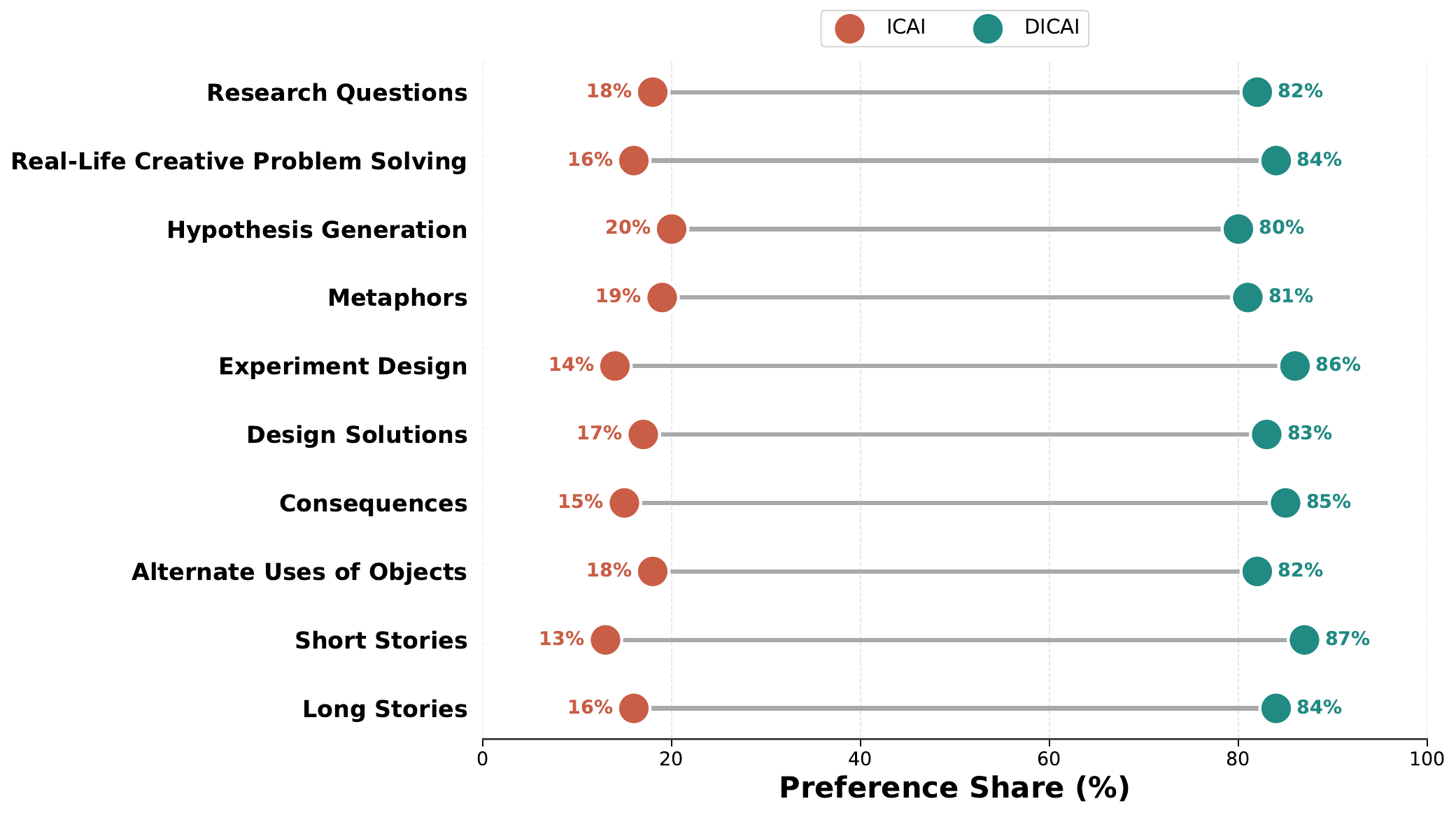}
    \caption{GPT-4o Coherence}
\end{subfigure}
\hfill
\begin{subfigure}{0.40\textwidth}
    \centering
    \includegraphics[width=\linewidth]{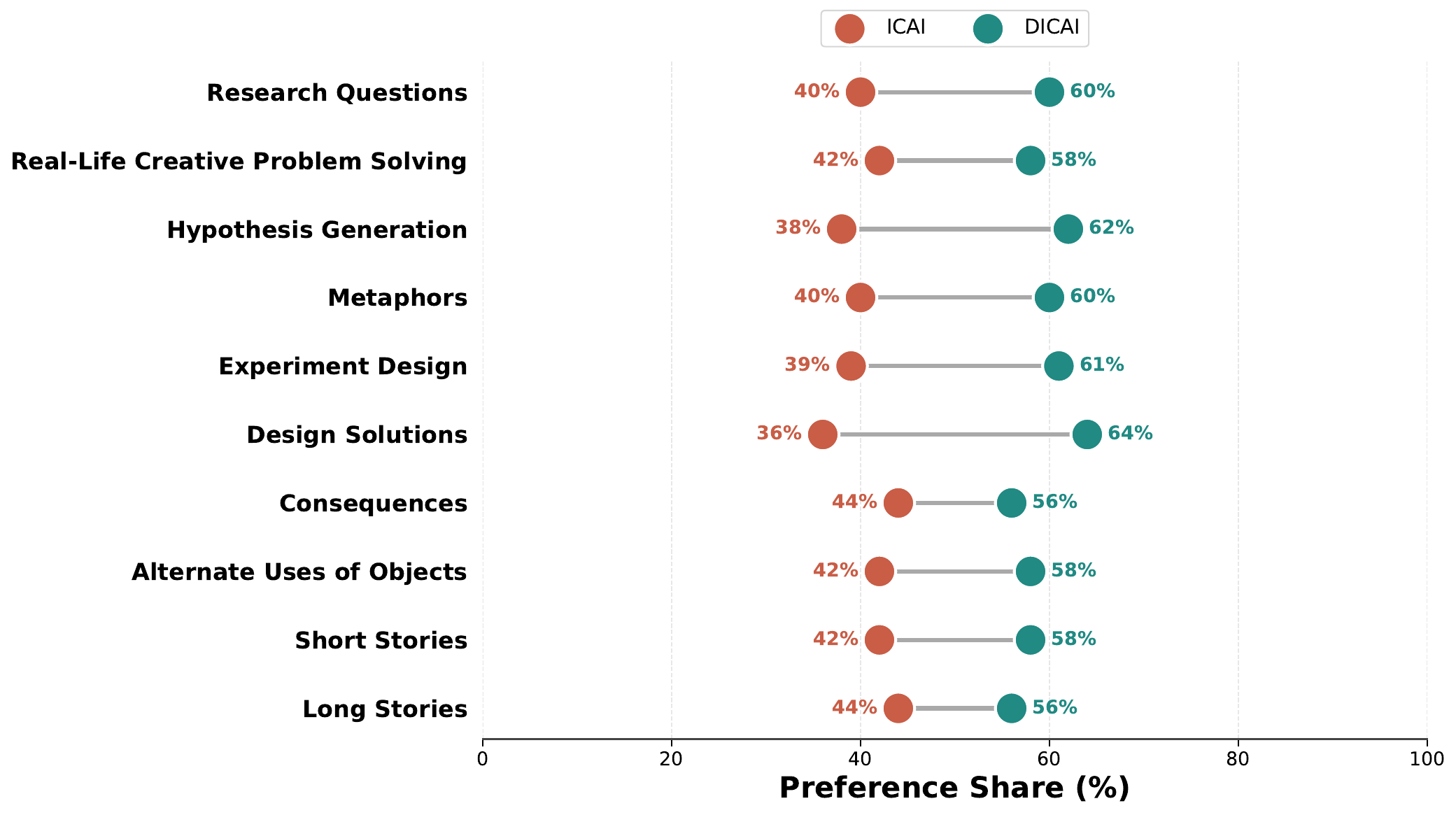}
    \caption{GPT-5 Coherence}
\end{subfigure}

\vspace{0.4em}

\begin{subfigure}{0.40\textwidth}
    \centering
    \includegraphics[width=\linewidth]{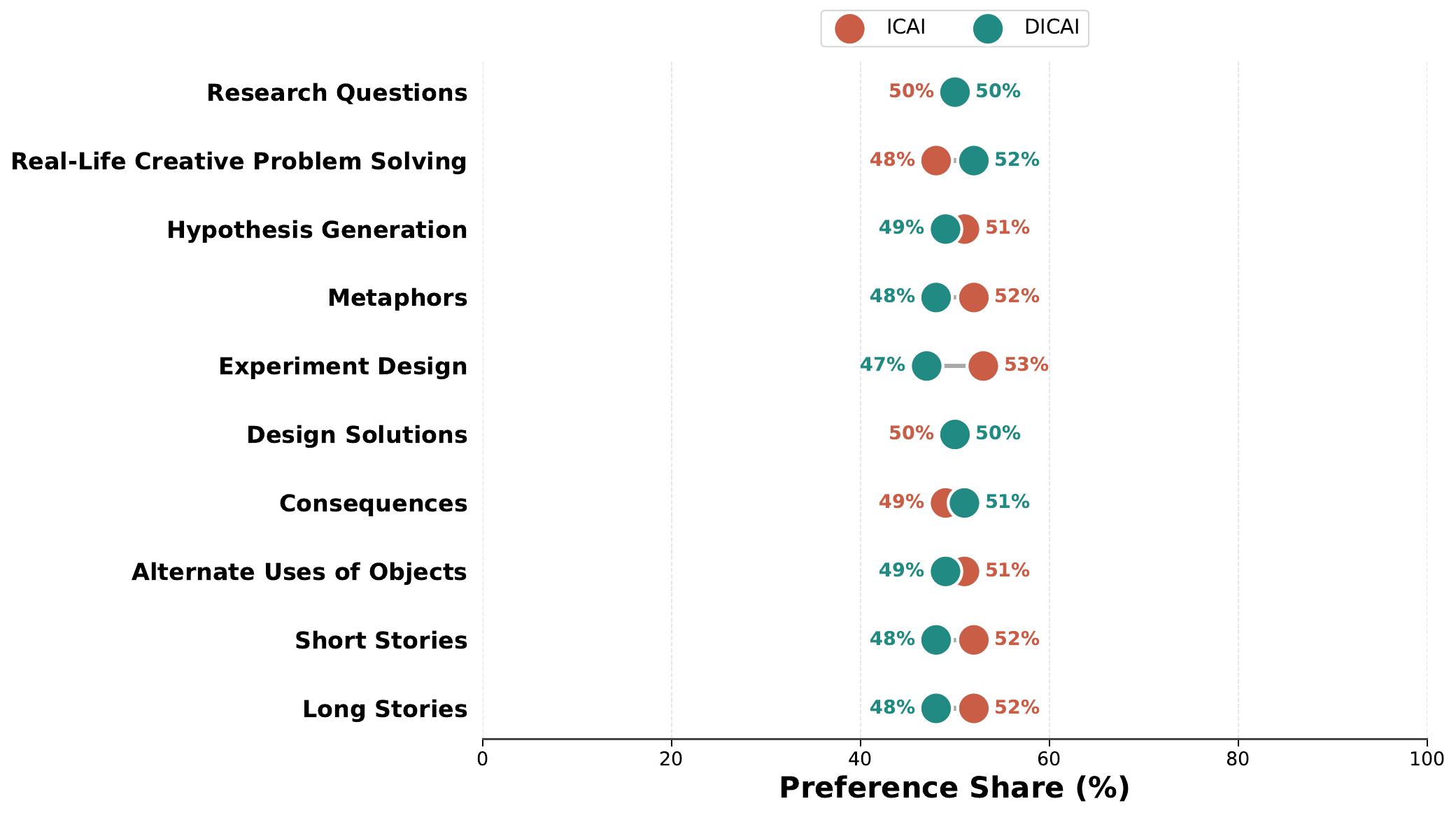}
    \caption{GPT-4o Feasibility}
\end{subfigure}
\hfill
\begin{subfigure}{0.40\textwidth}
    \centering
    \includegraphics[width=\linewidth]{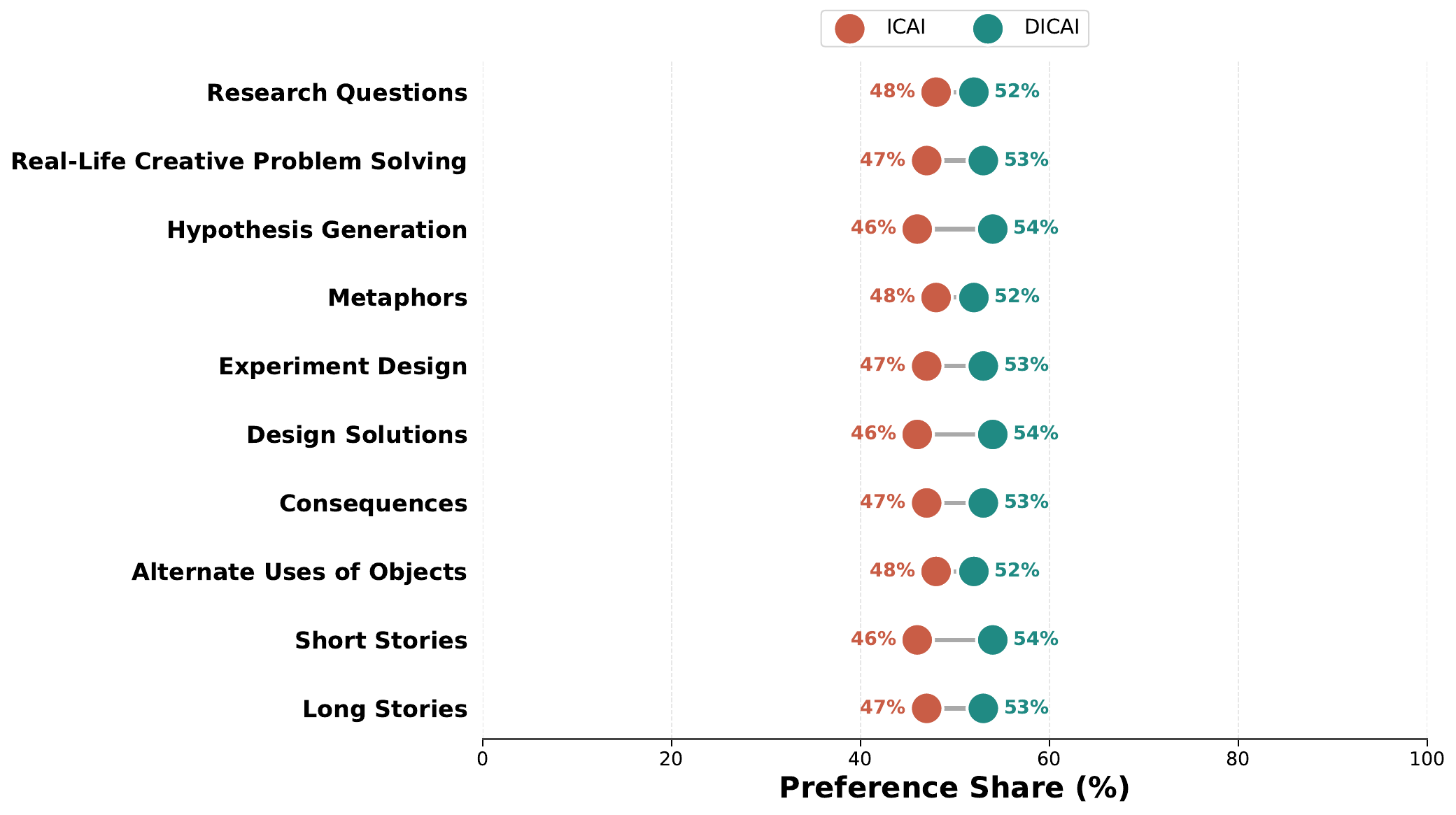}
    \caption{GPT-5 Feasibility}
\end{subfigure}

\vspace{0.4em}

\begin{subfigure}{0.40\textwidth}
    \centering
    \includegraphics[width=\linewidth]{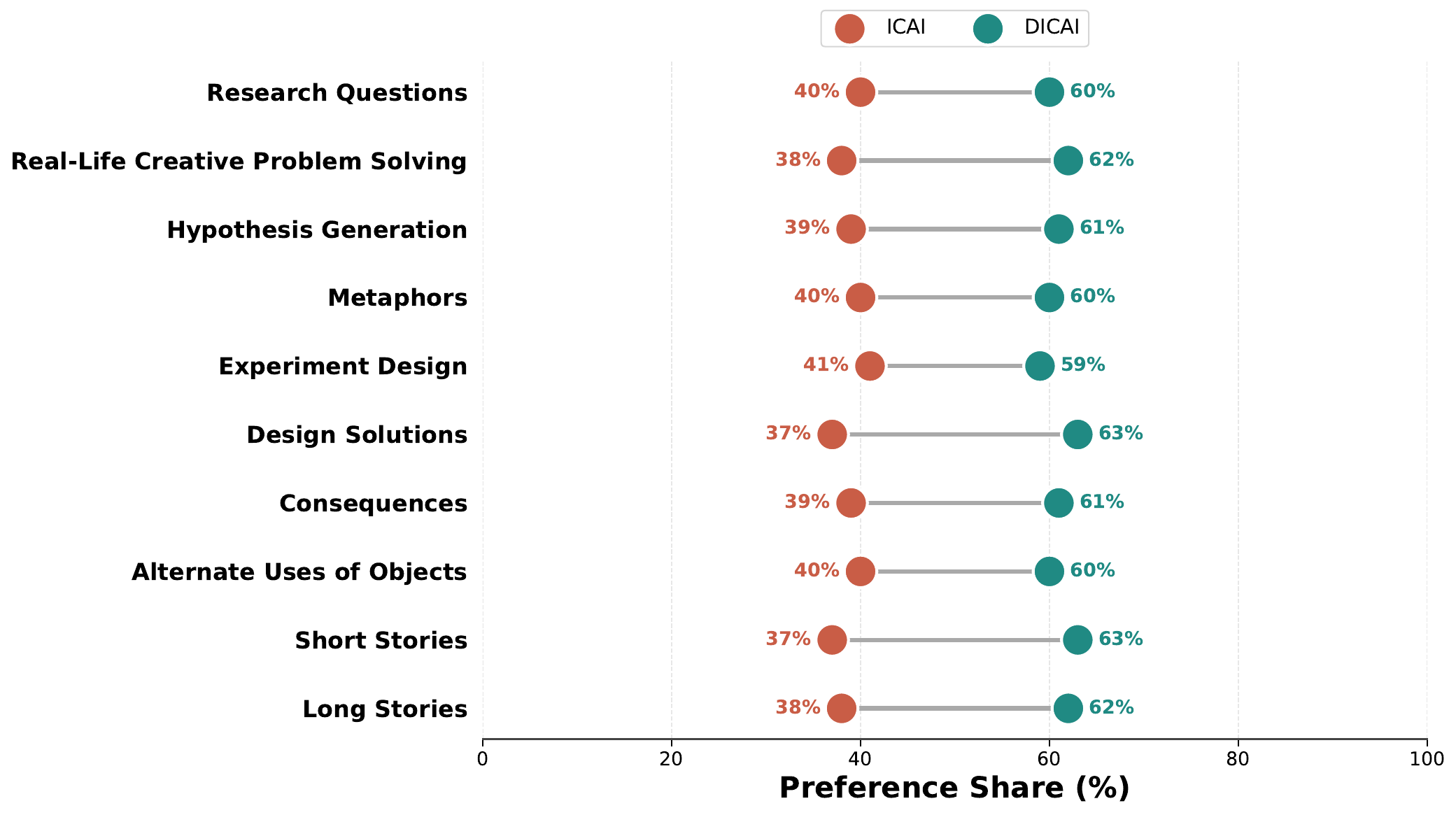}
    \caption{GPT-4o Faithfulness}
\end{subfigure}
\hfill
\begin{subfigure}{0.40\textwidth}
    \centering
    \includegraphics[width=\linewidth]{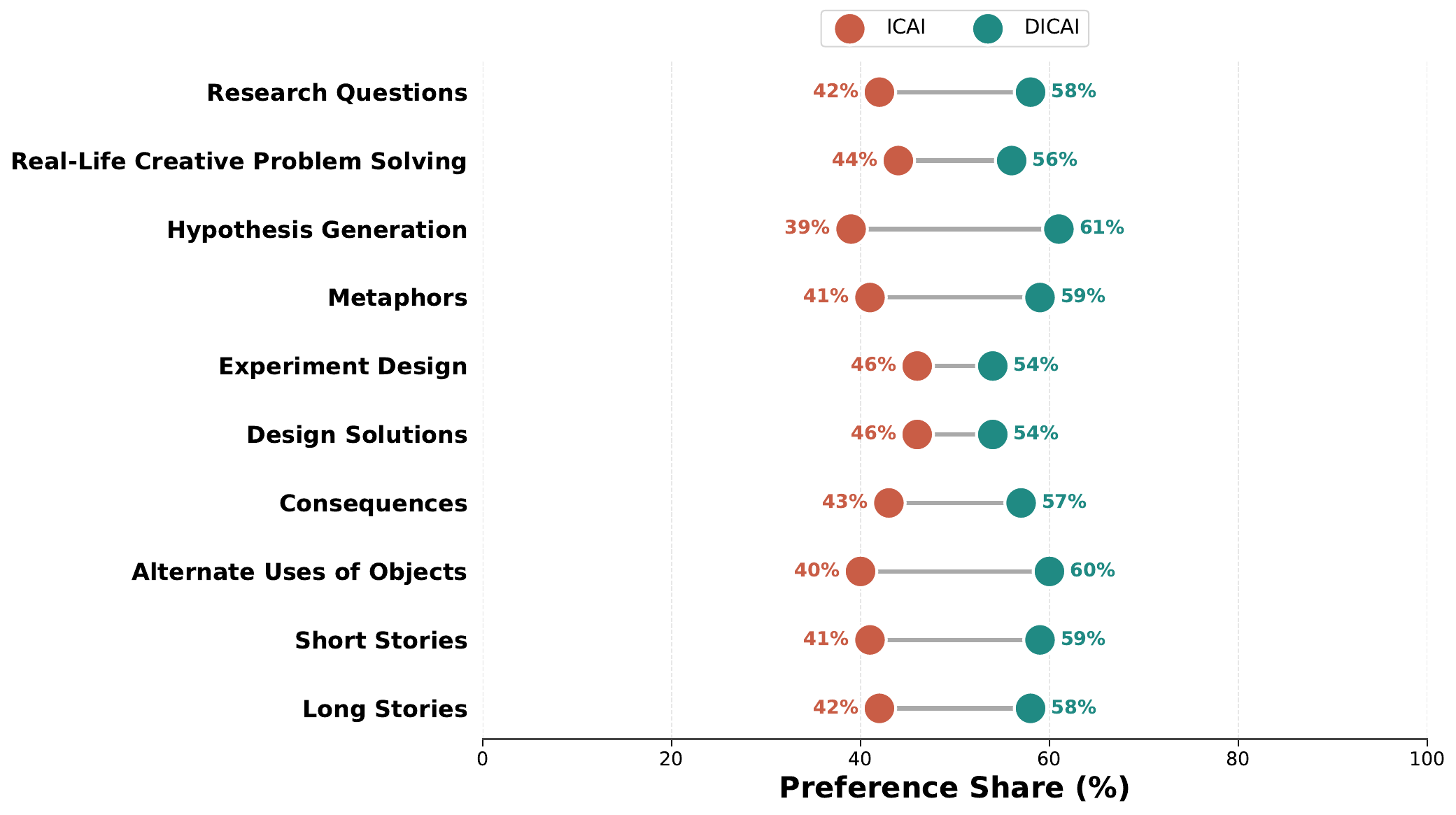}
    \caption{GPT-5 Faithfulness}
\end{subfigure}

\caption{
Comparison of Democratic ICAI and ICAI across five dimensions under GPT-4o (left column) and GPT-5 (right column).
Each row corresponds to a qualitative dimension evaluated across ten tasks.
Democratic ICAI consistently outperforms ICAI on structural criteria such as generality and coherence, while remaining competitive on feasibility.
}
\label{fig:constitution_metrics_combined}
\end{figure*}
\FloatBarrier

\clearpage
\subsection{Constitutional Distance Distributions}
\label{sec: constitutional_distance_appendix}

\begingroup
\renewenvironment{figure*}[1][]{%
  \par\noindent
  \begin{adjustbox}{max totalsize={\textwidth}{0.99\textheight},center}
  \begin{minipage}{\textwidth}
  \captionsetup{type=figure}
  \setlength{\textwidth}{\linewidth}
}{%
  \end{minipage}
  \end{adjustbox}
  \par
}
\begin{figure*}[t]
    \centering
    \begin{subfigure}[t]{0.6\textwidth}
        \centering
        \includegraphics[width=\linewidth]{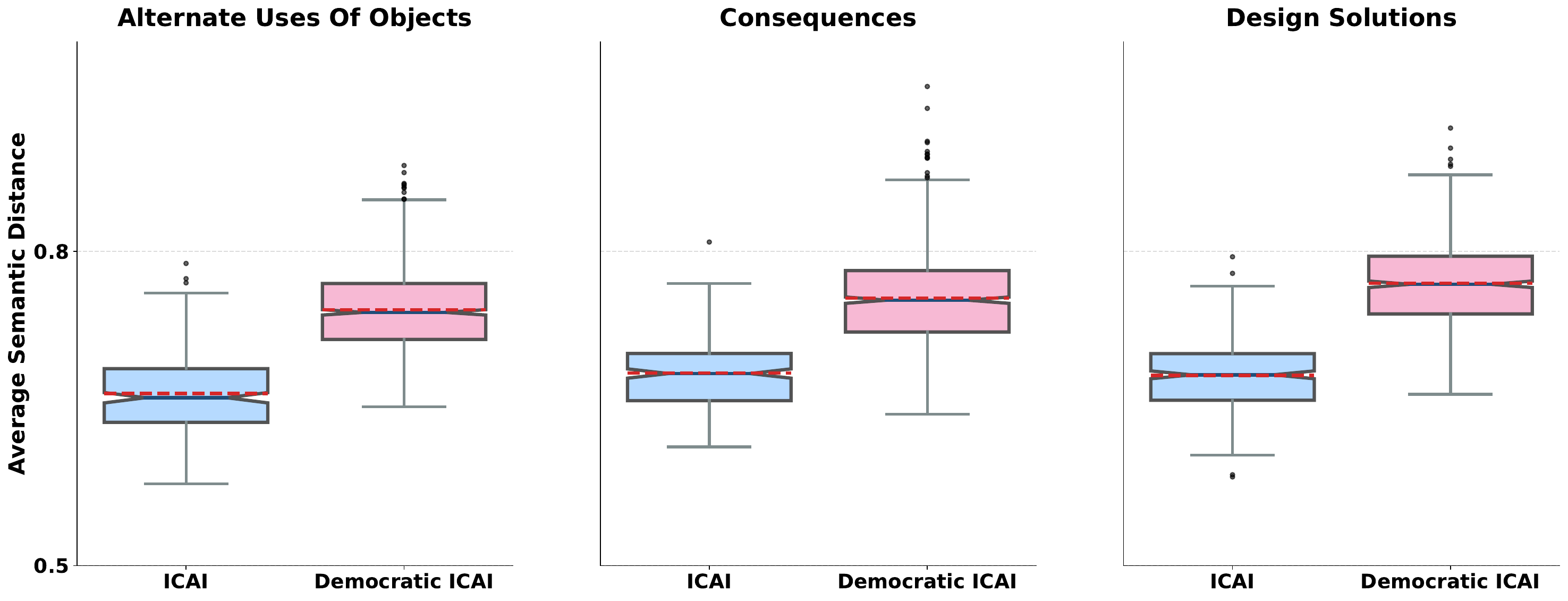}
         \caption{Distance among principles within the constitution across tasks (GPT-4o).}
        \label{fig:constitutional_distance_icai_gpt4o}
    \end{subfigure}
    \hfill
    \begin{subfigure}[t]{0.6\textwidth}
        \centering
        \includegraphics[width=\linewidth]{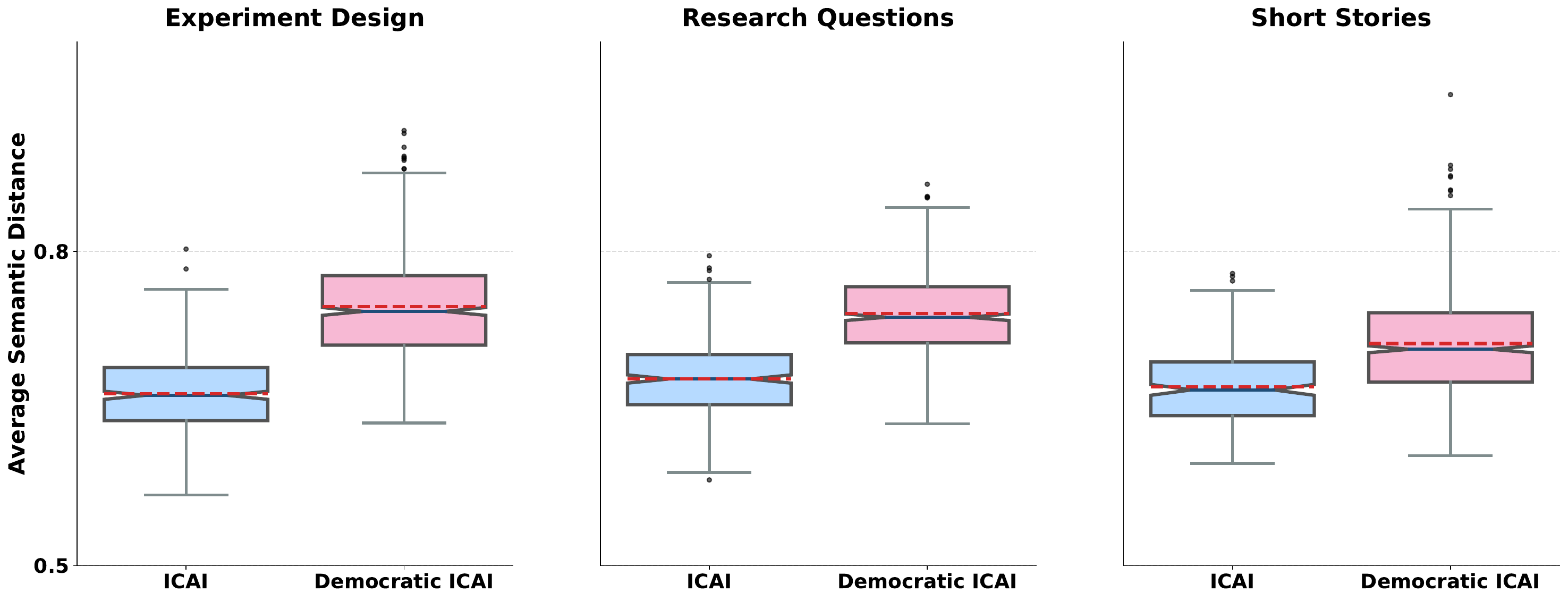}
        \caption{Distance among principles within the constitution across tasks (GPT-4o).}
        \label{fig:constitutional_distance_dicai_gpt4o}
    \end{subfigure}

    \begin{subfigure}[t]{0.6\textwidth}
        \centering
        \includegraphics[width=\linewidth]{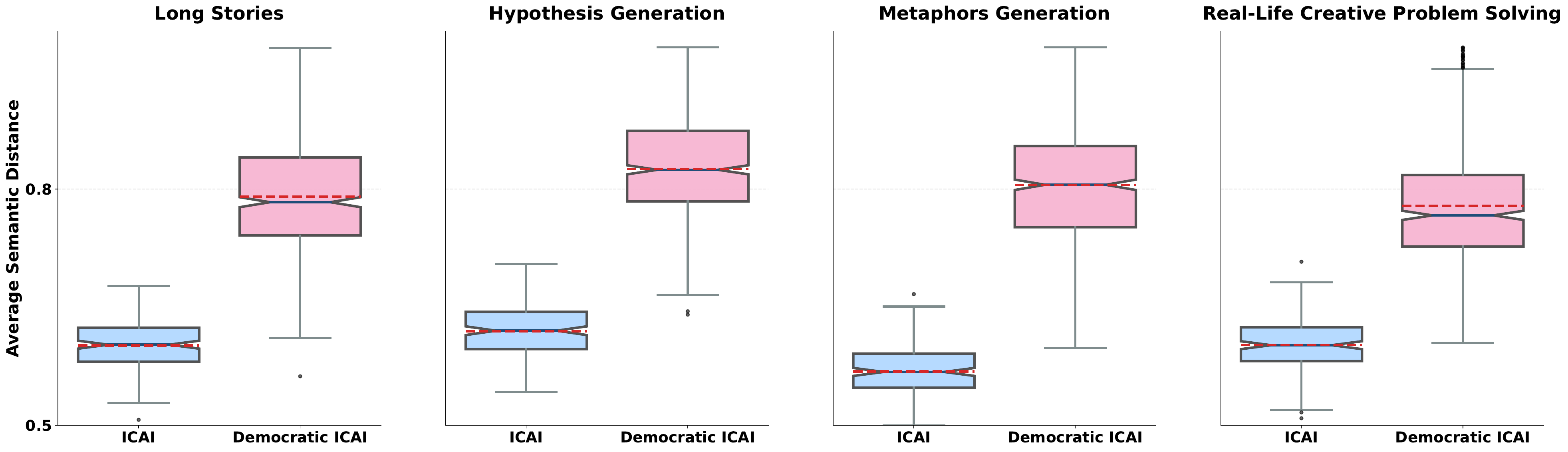}
        \caption{Distance among principles within the constitution across tasks (GPT-5).}
        \label{fig:constitutional_distance_comparison}
    \end{subfigure}
    
     \begin{subfigure}[t]{0.6\textwidth}
        \centering
        \includegraphics[width=\linewidth]{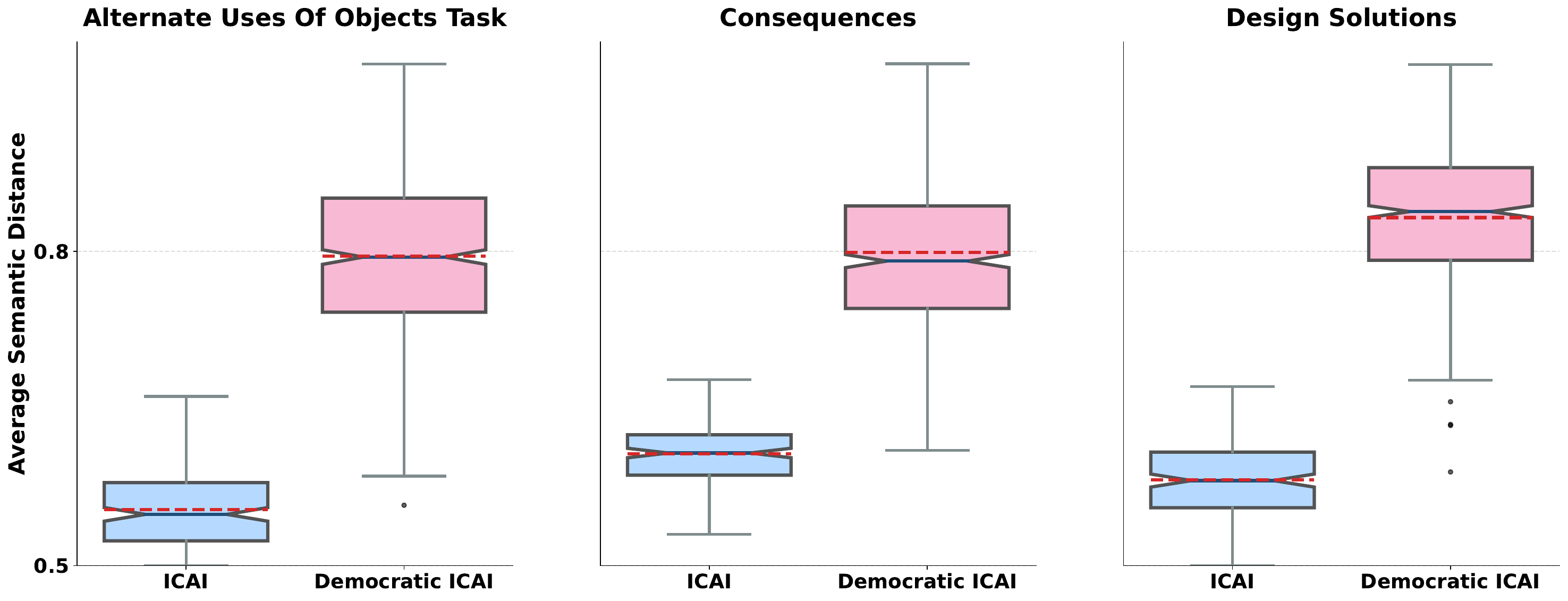}
         \caption{Distance among principles within the constitution across tasks (GPT-5).}
        \label{fig:constitutional_distance_icai_gpt5}
    \end{subfigure}
    \hfill
    \begin{subfigure}[t]{0.6\textwidth}
        \centering
        \includegraphics[width=\linewidth]{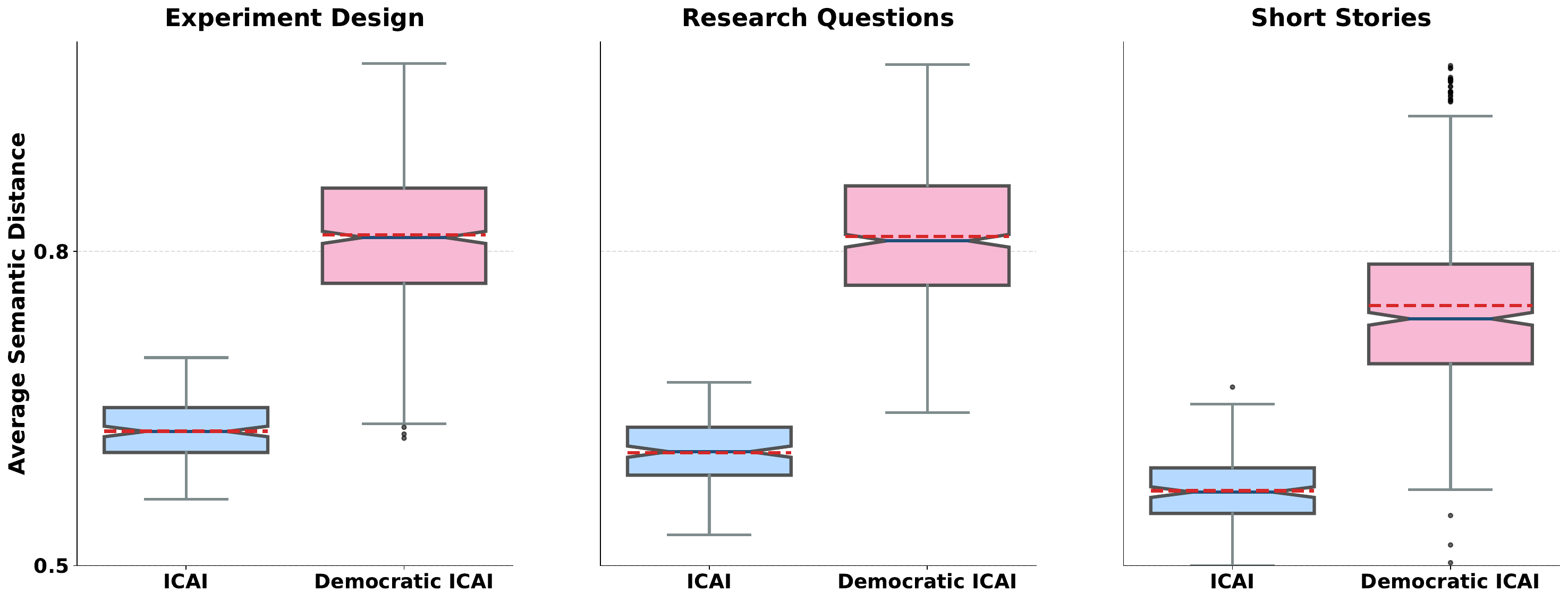}
        \caption{Distance among principles within the constitution across tasks (GPT-5).}
        \label{fig:constitutional_distance_dicai_gpt5}
    \end{subfigure}

    \caption{Distribution of average semantic distance across tasks for ICAI and Democratic ICAI. Average semantic distance is computed as the average cosine distance of each constitutional principle from all other principles in the constitution.
    }
    \label{fig: constitutional_distance_appendix}
\end{figure*}

\endgroup

\subsection{Case-Level Analysis of Democratic ICAI and ICAI Constitutions}
\label{sec:case_level_constitution_analysis}
To provide a concrete comparison between Democratic ICAI and ICAI, Table~\ref{tab:case_level_comparison} presents representative cases where Democratic ICAI alone succeeds, ICAI alone succeeds, or both correctly reconstruct the human preference. For each case, we also report the key principles underlying the correct prediction.

\begin{table}[!tbh]
\centering
\scriptsize
\captionsetup{skip=3pt}

\setlength{\tabcolsep}{2pt}
\renewcommand{\arraystretch}{0.00}   
\setlength{\aboverulesep}{1pt}
\setlength{\belowrulesep}{1pt}

\begin{tabular}{@{}>{\raggedright\arraybackslash}p{0.11\linewidth}
                >{\raggedright\arraybackslash}p{0.18\linewidth}
                >{\raggedright\arraybackslash}p{0.14\linewidth}
                >{\raggedright\arraybackslash}p{0.13\linewidth}
                >{\raggedright\arraybackslash}p{0.06\linewidth}
                >{\raggedright\arraybackslash}p{0.32\linewidth}@{}}
\toprule
\textbf{Task Type} & \textbf{Task} & \textbf{Chosen} & \textbf{Rejected} & \textbf{Correct} & \textbf{Key Principles Driving Preference} \\
\midrule
Research Questions &
Come up with an original and creative scientific research question for the following scenario: You created the first vehicle that can reach an unexplored part of the ocean floor. &
What temperature is this part of the ocean floor? &
What kingdom do these organisms stem from? &
DICAI &
1. Select the response that does not presuppose the existence of life or other unverified phenomena. 2. Select the response that provides foundational, directly observable information before making higher-level assumptions.
\\
\midrule
Metaphors &
Come up with an original and creative metaphor to describe the following concept: gross food. &
that jello was like a slimey slug on a rainy day &
Eating that dish was akin to consuming mud; about the same texture and taste and probably the same nutritional value. &
ICAI &
1. Select the response that includes a sensory description. 2. Select the response that evokes stronger and more grotesque imagery.
\\
\midrule
Research Questions &
Come up with an original and creative scientific research question for the following scenario: You travel to a jungle that contains no human life and is completely unknown to the scientific community. What scientific questions could you ask about this jungle? &
are there apex predators we might not know about &
Why haven't scientists discovered it sooner? &
Both &
\textbf{DICAI:} 1. Select the response that investigates the ecosystem itself rather than human interaction with it. 2. Select the response that asks about biological or ecological phenomena rather than historical or logistical circumstances. \textbf{ICAI:} 1. Select the response that centers foundational ecosystem dynamics. 2. Select the response that investigates directly observable biological phenomena.
\\
\midrule
Alternate Uses of Objects Task &
Come up with an original and creative use for the following object: box &
to capture your pet &
tearing &
DICAI &
\textbf{DICAI:} 1. Select the response that offers a novel, creative use. 2. Select the response that demonstrates more imaginative or unconventional. 
\\
\midrule
Consequences &
Come up with an original and creative consequence for the following scenario: What would be the result if everyone suddenly lost the sense of balance and were unable
to stay in the upright position for more than a moment?
 &
wouldn't be able to eat correctly &
Less exercise &
ICAI &
\textbf{ICAI:} 1. Select the response that articulates a first-order consequence directly tied to the premise. 2. Select the response that has a direct, physiologically coherent consequence of the premise.
\\
\midrule
Design Solutions &
Come up with an original and creative solution to increase access to food and drinking water in developing countries.
 &
vertical farming
 &
Cultivate vegetable gardens &
Both &
\textbf{DICAI:} 1. Select the response that is more innovative or futuristic. 2. Select the response that proposes a novel or unique solution.
\textbf{ICAI:} 1. Select the response that makes efficient use of limited land. 2. Select the response that directly addresses root constraints like water access and productive
capacity.
\\
\midrule
Experiment Design &
Come up with an original and creative experiment to test the following hypothesis or research question: You think birds can understand humans but just choose to ignore us because what humans say isn't important to their lives. How could you test that hypothesis? &
have humans talk about what the birds eat &
try telling a bird what to do and see if they do it &
DICAI &
\textbf{DICAI:} 1. Select the response that directly manipulates the hypothesized causal mechanism rather than a downstream consequence. 2. Select the response that operationalizes the key theoretical construct into a measurable variable.
\\
\midrule
Hypothesis Generation &
Come up with an original and creative scientific hypothesis for the following scenario: You notice that dogs seem to like one of your friends, but cats seem to like another friend. What hypothesis do you have about why that is? &
cats and dogs are attracted to different displays of emotion &
dogs and cats like different types of people &
ICAI &
\textbf{ICAI:} 1. Select the response that presents a testable, specific hypothesis. 2. Select the response that explains behavior with specific reasoning.
\\
\midrule
Real-Life Creative Problem Solving &
Becky is a college student who works part-time at Mark's Pizzeria. Mark, the owner of the restaurant, has treated Becky very well. He gave her a job that she needs to help pay her rent when no other business would employ her because she was arrested for shoplifting three years ago. Mark also lets Becky work around her school schedule... &
Becky should tell Mark in confidence and make it clear that he should not tell Jim that she told him. She could suggest that he say a customer told him. She needs to discuss her shoplifting record and make it clear that although she has done it in the past, she has moved on. Mark will see that when Jim is gone, the theft is stopping, and it is not Becky. Becky could also make an anonymous call to the police, since Jim's stealing is a crime. &
I think that Becky should go to Mark and let him know what Jim is doing. Have Becky tell Mark her entire story. Mark will hopefully talk to Jim and give him a second chance because Becky needs Jim to have a job. &
Both &
\textbf{DICAI:} 1. Select the response that provides detailed and actionable solutions. 2. Select the response that considers backup plans or contingencies. \textbf{ICAI:} 1. Select the response that integrates ethical, relational, financial, and operational constraints into a coherent, executable plan. 2. Select the response that provides multiple concrete actions, confidentiality safeguards, and contingency paths rather than relying on a single hoped-for outcome.
\\
\bottomrule
\end{tabular}

\caption{Representative examples illustrating cases where Democratic ICAI succeeds and ICAI fails, ICAI succeeds and Democratic ICAI fails, and both constitutions succeed. We also present the underlying principles for the constitution that correctly reconstructs the preference, which may explain why the evaluator selected the preferred response.}
\label{tab:case_level_comparison}
\end{table}

\clearpage

\section{Bias, Spuriousness, and Harmfulness Audit}
\label{sec:audit_prompt_examples}

Motivated by prior observations that induced constitutions and reward-modeling procedures can capture criteria that are highly superficial \citep{liu2024rrm,xie2025auto}, we audit the generated principles along bias, spuriousness, and harmfulness axes. As shown in Tables~\ref{tab:bias_spurious_flags} and~\ref{tab:bias_llama_audit}, Democratic ICAI exhibits lower spuriousness than ICAI, suggesting that the debate stage consolidates redundant or overly specific principles into more generalizable principles. Although Democratic ICAI shows slightly more bias flags, these largely correspond to protective principles intended to reduce demographic, cultural, or religious favoritism rather than encode unfair preferences. For example, Table~\ref{tab:audit_examples} shows that a principle such as ``Select the response that avoids culturally or religiously marked framing'' may be flagged because it could discourage culturally specific expression in some settings, but its intended function is to promote neutrality.

The spuriousness flags are similarly mild and mostly reflect stylistic preferences, such as favoring lyrically phrased responses in the Metaphors constitution, which may not always directly reflect underlying quality (Table~\ref{tab:audit_examples}). Harmfulness flags are concentrated around safety-oriented principles in the aggregate audit tables. For instance, ``Select the response that directly informs mission protocols, biosafety, and ethical oversight'' can be flagged because it references biosafety and risk management, but its purpose is to encourage responsible decision-making rather than harmful behavior. The external-auditor prompt used for these analyses is provided in Figure~\ref{fig:audit-prompt} in Appendix~\ref{app:dicai_prompts}.

\begingroup
\begin{center}
\centering
\scriptsize
\setlength{\tabcolsep}{2pt}
\renewcommand{\arraystretch}{1.15}

\begin{tabular*}{\textwidth}{@{\extracolsep{\fill}}>{\raggedright\arraybackslash}p{3.6cm}*{12}{c}@{}}
\toprule

\multirow{3}{*}{\raisebox{-0.65\baselineskip}{\textbf{Task}}}
& \multicolumn{6}{c}{\textbf{DICAI}}
& \multicolumn{6}{c}{\textbf{ICAI}} \\

\cmidrule(lr){2-7}
\cmidrule(lr){8-13}

&
\multicolumn{2}{c}{\textbf{Bias}}
&
\multicolumn{2}{c}{\textbf{Spurious}}
&
\multicolumn{2}{c}{\textbf{Harmfulness}}
&
\multicolumn{2}{c}{\textbf{Bias}}
&
\multicolumn{2}{c}{\textbf{Spurious}}
&
\multicolumn{2}{c}{\textbf{Harmfulness}}
\\

\cmidrule(lr){2-3}
\cmidrule(lr){4-5}
\cmidrule(lr){6-7}
\cmidrule(lr){8-9}
\cmidrule(lr){10-11}
\cmidrule(lr){12-13}

&
\% & high/med
& \% & high/med
& \% & high/med
& \% & high/med
& \% & high/med
& \% & high/med
\\

\midrule

Alternate Uses of Objects
& 0.19 & 0/2
& 1.62 & 0/17
& 0.10 & 0/3
& 0.08 & 0/1
& 3.95 & 0/4
& 0.10 & 0/1 \\

Consequences
& 0.00 & 0/0
& 2.47 & 0/21
& 2.10 & 0/13
& 0.15 & 0/2
& 5.76 & 0/7
& 2.10 & 0/13 \\

Design Solutions
& 1.80 & 0/9
& 2.20 & 0/11
& 1.40 & 0/6
& 1.30 & 0/6
& 4.88 & 0/5
& 1.40 & 0/6 \\

Experiment Design
& 0.00 & 0/0
& 0.00 & 0/0
& 0.00 & 0/0
& 0.10 & 0/0
& 3.56 & 0/3
& 0.00 & 0/0 \\

Hypothesis Generation
& 1.71 & 0/12
& 4.57 & 0/32
& 2.10 & 0/5
& 1.20 & 0/5
& 5.12 & 0/5
& 2.10 & 0/5 \\

Metaphors
& 0.04 & 0/3
& 3.11 & 0/23
& 0.00 & 0/1
& 0.00 & 0/0
& 4.21 & 0/4
& 0.00 & 0/0 \\

Real-Life Creative Problem Solving
& 0.00 & 0/0
& 0.00 & 0/0
& 3.20 & 0/11
& 0.18 & 0/2
& 5.94 & 0/8
& 3.20 & 0/11 \\

Research Questions
& 0.00 & 0/0
& 0.04 & 0/4
& 0.02 & 0/0
& 0.00 & 0/0
& 3.04 & 0/3
& 0.00 & 0/0 \\

Long Stories (LiTBench)
& 0.00 & 0/0
& 1.25 & 0/0
& 0.00 & 0/0
& 0.00 & 0/0
& 6.27 & 0/6
& 0.00 & 0/0 \\

Short Stories (MuCE)
& 0.50 & 0/3
& 3.40 & 0/17
& 1.10 & 0/0
& 0.05 & 0/0
& 4.87 & 0/5
& 1.10 & 0/0 \\

\bottomrule
\end{tabular*}

\captionof{table}{
Bias, spuriousness, and harmfulness criteria audit for Democratic ICAI and ICAI constitutions. We use Qwen2.5-32B as an external auditor to avoid model bias.
}
\label{tab:bias_spurious_flags}

\end{center}
\endgroup

\begingroup
\begin{center}
\centering
\scriptsize
\setlength{\tabcolsep}{2pt}
\renewcommand{\arraystretch}{1.15}

\begin{tabular*}{\textwidth}{@{\extracolsep{\fill}}>{\raggedright\arraybackslash}p{3.6cm}*{12}{c}@{}}
\toprule

\multirow{3}{*}{\raisebox{-0.65\baselineskip}{\textbf{Task}}}
& \multicolumn{6}{c}{\textbf{DICAI}}
& \multicolumn{6}{c}{\textbf{ICAI}} \\

\cmidrule(lr){2-7}
\cmidrule(lr){8-13}

&
\multicolumn{2}{c}{\textbf{Bias}}
&
\multicolumn{2}{c}{\textbf{Spurious}}
&
\multicolumn{2}{c}{\textbf{Harmfulness}}
&
\multicolumn{2}{c}{\textbf{Bias}}
&
\multicolumn{2}{c}{\textbf{Spurious}}
&
\multicolumn{2}{c}{\textbf{Harmfulness}}
\\

\cmidrule(lr){2-3}
\cmidrule(lr){4-5}
\cmidrule(lr){6-7}
\cmidrule(lr){8-9}
\cmidrule(lr){10-11}
\cmidrule(lr){12-13}

&
\% & high/med
& \% & high/med
& \% & high/med
& \% & high/med
& \% & high/med
& \% & high/med
\\

\midrule

Alternate Uses of Objects
& 0.12 & 0/2
& 0.08 & 0/1
& 0.05 & 0/1
& 0.05 & 0/1
& 3.82 & 0/4
& 0.08 & 0/1 \\

Consequences
& 0.00 & 0/8
& 1.85 & 0/5
& 0.10 & 0/9
& 0.12 & 0/2
& 5.44 & 0/6
& 1.95 & 0/11 \\

Design Solutions
& 1.30 & 0/4
& 1.10 & 0/3
& 1.20 & 0/4
& 1.18 & 0/5
& 4.55 & 0/5
& 1.32 & 0/5 \\

Experiment Design
& 0.10 & 0/0
& 0.05 & 0/1
& 0.00 & 0/0
& 0.08 & 0/0
& 3.24 & 0/3
& 0.05 & 0/0 \\

Hypothesis Generation
& 1.20 & 0/3
& 1.60 & 0/2
& 1.40 & 0/3
& 1.05 & 0/4
& 4.93 & 0/5
& 1.88 & 0/4 \\

Metaphors
& 0.02 & 0/0
& 2.89 & 0/1
& 0.00 & 0/0
& 0.00 & 0/0
& 4.06 & 0/4
& 0.00 & 0/0 \\

Real-Life Creative Problem Solving
& 0.00 & 0/7
& 2.70 & 0/4
& 0.10 & 0/8
& 0.15 & 0/2
& 5.71 & 0/7
& 2.95 & 0/10 \\

Research Questions
& 0.00 & 0/0
& 0.01 & 0/0
& 0.02 & 0/0
& 0.00 & 0/0
& 5.21 & 0/5
& 0.00 & 0/0 \\

Long Stories (LiTBench)
& 0.00 & 0/0
& 1.32 & 0/0
& 0.00 & 0/0
& 0.00 & 0/0
& 7.32 & 0/7
& 0.00 & 0/0 \\

Short Stories (MuCE)
& 0.00 & 0/0
& 0.95 & 0/0
& 0.05 & 0/0
& 0.03 & 0/0
& 4.65 & 0/4
& 0.98 & 0/0 \\

\bottomrule
\end{tabular*}

\captionof{table}{
Bias, spuriousness, and harmfulness criteria audit for Democratic ICAI and ICAI constitutions. We use Llama-3.1-70B as an external auditor to avoid model bias.
}
\label{tab:bias_llama_audit}

\end{center}
\endgroup

\begin{center}
\centering
\small
\resizebox{\columnwidth}{!}{
\begin{tabular}{p{4cm} p{5.5cm} p{2cm} p{2cm} p{6cm}}
\toprule
\textbf{Category} & \textbf{Principle} & \textbf{Bias Sev.} & \textbf{Spurious Sev.} & \textbf{Annotator Rationale} \\
\midrule

Research Questions & Select the response that is quantifiable & None & Medium 
& May prioritize measurable attributes over qualitative aspects, risking superficial assessment of quality. \\

Research Questions & Select the response with highest novelty & None & Medium 
& Novelty alone may reward superficially unique outputs rather than substantively better responses. \\

Metaphors & Select the response that maintains a crisp, lyrical tone & None & Medium 
& Emphasis on stylistic tone may privilege surface-level writing qualities over substantive content. \\

Metaphors & Select the response that demonstrates rhetorical sharpness & None & Medium 
& Rhetorical flair can be achieved through style rather than meaningful or accurate content. \\

Metaphors & Select the response that avoids bilingual or ambiguous wording for a mixed-language audience & Medium & None 
& Implicit assumptions about audience language abilities may reflect demographic bias. \\

Metaphors & Select the response that avoids culturally or religiously marked framing & Medium & None 
& May suppress culturally specific expression, potentially marginalizing certain groups. \\

Metaphors & Select the response that is concise & None & Medium 
& Conciseness may reward brevity even when it reduces completeness or depth of explanation. \\

Metaphors & Select the response that maximizes engagement and memorability & None & Medium 
& Engagement may be driven by sensationalism rather than true informational or conceptual quality. \\

Metaphors & Select the response that is culturally general and does not target any social group & Medium & None 
& Assumes a universal cultural baseline, potentially undervaluing culturally specific perspectives. \\

Research Questions & Select the response with clean, polished language and line-level precision & None & Medium 
& May over-emphasize surface polish rather than underlying correctness or insight. \\

\bottomrule
\end{tabular}}
\captionof{table}{Examples of constitution principles flagged by the external auditor, including annotator rationales. Spurious flags often arise from reliance on stylistic or surface-level criteria, while bias flags reflect potential suppression or generalization of cultural and demographic variation.}
\label{tab:audit_examples}
\end{center}

\section{Ablation Studies}
\label{app:ablation_studies}
\subsection{Effect of Parliamentary Debate and Persona Diversity}
We isolate the contribution of the two deliberative components in Democratic ICAI: the number of debate rounds and the diversity of the persona committee. First, we vary the number of debate rounds while keeping the rest of the pipeline fixed. Second, we remove the parliamentary debate stage (Section~\ref{sec:debate}) and extract principles directly from either one persona or three personas. These variants allow us to test whether improvements come merely from collecting more rationales, or from the adversarial refinement process that forces rationales to be challenged, consolidated, and filtered before constitution drafting. The chairperson and debater prompts used for this stage are provided in Figures~\ref{chairperson_prompt} and~\ref{debater_prompt} in Appendix~\ref{app:dicai_prompts}.

A single round of debate already substantially exceeds ICAI, with returns saturating at 2 rounds across all three tasks, indicating that the benefit is front-loaded and Democratic ICAI remains compute-efficient relative to longer multi-round setups (Table~\ref{tab:ablation_debate_rounds}).

\begin{table}[!tbh]
\centering
\small
\begin{tabular}{lrrrr}
\toprule
Task & ICAI & 1 round & 2 rounds & 3 rounds \\
\midrule
Research Questions & 70.20 & 78.22 & 80.01 & 80.21 \\
Metaphors & 71.40 & 73.21 & 73.80 & 74.01 \\
Long Stories (LiTBench) & 62.89 & 67.63 & 68.00 & 68.70 \\
\bottomrule
\end{tabular}
\caption{Ablation on debate rounds. We vary the number of debate rounds (1, 2, 3) within the Democratic ICAI pipeline and compare against ICAI.}
\label{tab:ablation_debate_rounds}
\end{table}

A single persona without parliamentary debate underperforms even ICAI, while adding personas without debate only recovers ICAI-level performance; the full Democratic ICAI pipeline yields 4–8 point gains over the three-persona variant, attributing the improvement to the combination of adversarial debate and persona diversity (Table~\ref{tab:ablation_mad}).

\begin{table}[!tbh]
\centering
\small
\begin{tabular}{lrrrr}
\toprule
Task & ICAI & 1 Persona w/o PD & 3 Personas w/o PD & DICAI \\
\midrule
Research Questions & 70.20 & 66.22 & 72.77 & 80.21 \\
Metaphors          & 71.40 & 68.27 & 70.01 & 74.01 \\
Long Stories (LiTBench)   & 62.89 & 59.22 & 62.13 & 68.70 \\
\bottomrule
\end{tabular}
\caption{Ablation on the parliamentary debate stage and persona count. We remove the debate stage and extract principles directly from rationales, varying the number of personas (single vs.\ three) without Parliamentary Debate (PD), and compare against ICAI and the Democratic ICAI pipeline.}
\label{tab:ablation_mad}
\end{table}

\FloatBarrier
\subsection{Sensitivity to Persona Design}
To assess how sensitive Democratic ICAI is to persona specification, we ablate the structured persona profile used in the reasoning assembly stage. Our complete personas are defined along demographic, socioeconomic, psychological, and value-based dimensions. We remove one component at a time and rerun the full constitution-induction pipeline using 500 training samples per task, evaluating the resulting constitutions on the corresponding held-out test sets with GPT-4o. Example complete persona prompts are provided in Appendix~\ref{app:reasoning_personas}.

Removing any persona dimension reduces performance across all three tasks, while the complete persona consistently performs best. The largest drop comes from removing the psychological dimension (Table~\ref{tab:ablation_persona_design}). This observation aligns with prior work highlighting the importance of socio-psychological attributes for preference simulation \citep{joshi2025improving, venkit2026need}. The socioeconomic and value-based dimensions also contribute positively, with the complete persona model consistently achieving the best results across all benchmarks.
\begin{table}[!tbh]
\centering
\footnotesize
\captionsetup{skip=3pt}
\setlength{\tabcolsep}{2pt}
\renewcommand{\arraystretch}{1.12}
\begin{tabular}{@{}>{\raggedright\arraybackslash}p{0.27\linewidth}*{4}{>{\centering\arraybackslash}p{0.17\linewidth}}@{}}
\toprule
Task & \makecell[c]{Variant 1\\(w/o Value-Based)} & \makecell[c]{Variant 2\\(w/o Socioeconomic)} & \makecell[c]{Variant 3\\(w/o Psychological)} & \makecell[c]{Complete\\Persona} \\
\midrule
Long Stories (LiTBench) & 68.05 & 66.12 & 63.42 & \textbf{68.70} \\
Research Questions & 77.82 & 75.11 & 72.31 & \textbf{80.21} \\
Metaphors & 72.01 & 71.22 & 69.33 & \textbf{74.01} \\
\bottomrule
\end{tabular}
\caption{Sensitivity analysis of persona design. Persona Variant 1 removes the value-based dimension, Persona Variant 2 removes the socioeconomic dimension, and Persona Variant 3 removes the psychological dimension. Results are reported as preference prediction accuracy (\%) across three evaluation tasks.}
\label{tab:ablation_persona_design}
\end{table}

\FloatBarrier

\subsection{Abstraction Strategy}
We evaluate the abstraction step used to convert clustered principles into the final constitution. After clustering semantically similar candidate principles, we compare our LLM-based abstraction strategy against two simpler representative-selection baselines. The first randomly samples one principle from each cluster, while the second selects the principle closest to the cluster centroid, following prior work \citep{henneking2025decoding}. The abstraction prompt is provided in Figure~\ref{abstraction_prompt} in Appendix~\ref{app:dicai_prompts}.

As we show in Table~\ref{tab:ablation_abstraction_strategy}, randomly selecting principles performs worst, while centroid-based selection improves performance by choosing more representative cluster members. LLM-based abstraction performs best across all tasks, indicating that synthesizing shared information across clustered principles yields more general principles that improve held-out preference reconstruction.
\begin{table}[!tbh]
\centering
\small
\captionsetup{skip=3pt}
\begin{tabular}{lrrr}
\toprule
Task & Random Sampling & Centroid & LLM-based Abstraction (Ours) \\
\midrule
Long Stories (LiTBench) & 62.25 & 65.42 & \textbf{68.70} \\
Research Questions & 72.82 & 76.71 & \textbf{80.21} \\
Metaphors & 69.01 & 71.43 & \textbf{74.01} \\
\bottomrule
\end{tabular}
\caption{Comparison of principle selection strategies.}
\label{tab:ablation_abstraction_strategy}
\end{table}

\FloatBarrier

\subsection{Clustering Granularity}
We examine the effect of clustering granularity on the quality of the induced constitution. In our main pipeline, we select the number of clusters using the silhouette score, which balances within-cluster cohesion against between-cluster separation. To test the sensitivity of this choice, we vary the number of clusters to 50\% and 150\% of the selected optimum, corresponding to coarser and finer clusterings, respectively, and regenerate the constitution under each setting.

Performance decreases when the number of clusters is set below or above the silhouette-selected optimum. Coarser clustering merges distinct preference criteria, while finer clustering fragments related principles. As we show in Table~\ref{tab:ablation_clustering_granularity} both reduce abstraction quality and weaken preference reconstruction.
\begin{table}[!tbh]
\centering
\small
\captionsetup{skip=3pt}
\setlength{\tabcolsep}{4pt}
\begin{tabular}{lrrr}
\toprule
Task & 50\% of Optimal Clusters & 150\% of Optimal Clusters & Highest Silhouette Score \\
\midrule
Long Stories (LiTBench) & 63.55 & 64.46 & \textbf{68.70} \\
Research Questions & 69.87 & 70.33 & \textbf{80.21} \\
Metaphors & 66.11 & 67.93 & \textbf{74.01} \\
\bottomrule
\end{tabular}
\caption{Sensitivity analysis of clustering granularity.}
\label{tab:ablation_clustering_granularity}
\end{table}

\begin{table}[!tbh]
\centering
\small
\captionsetup{skip=3pt}
\setlength{\tabcolsep}{4pt}
\begin{tabular}{lcc}
\toprule
\textbf{Task} & \textbf{DICAI} & \textbf{ICAI} \\
\midrule
Alternate Uses of Objects & 1050 (0.1411) & 750 (0.1165) \\
Consequences & 850 (0.0929) & 600 (0.0900) \\
Design Solutions & 500 (0.0959) & 650 (0.0970) \\
Experiment Design & 1000 (0.0851) & 600 (0.0800) \\
Hypothesis Generation & 700 (0.1048) & 800 (0.0900) \\
Metaphors & 850 (0.1409) & 750 (0.1290) \\
Real-Life Creative Problem Solving & 850 (0.0899) & 700 (0.0850) \\
Research Questions & 700 (0.0871) & 600 (0.0850) \\
Short Stories (MuCE) & 600 (0.0947) & 450 (0.0830) \\
\bottomrule
\end{tabular}
\caption{\textbf{Optimal number of clusters ($M^*$)}. The values in parentheses denote the corresponding peak silhouette scores obtained at $M^*$, measuring clustering quality in terms of intra-cluster cohesion and inter-cluster separation. Higher scores indicate better-defined and more separable clusters.}
\label{tab:silhouette_optimal_clusters}
\end{table}

\FloatBarrier
\subsection{Number of Induced Principles}
We evaluate how sensitive Democratic ICAI is to the size of the final induced constitution. Starting from the full constitution produced by our pipeline, we retain only 25\%, 50\%, and 75\% of the principles and compare these reduced versions against the complete 100\% constitution.

As we show in Table~\ref{tab:ablation_induced_principles}, smaller constitutions provide a more compact representation but act as lossy compression, discarding complementary preference signals needed for reliable reconstruction across tasks. Accuracy improves as more induced principles are retained in the final constitution. 
\begin{table}[!tbh]
\centering
\small
\captionsetup{skip=3pt}
\begin{tabular}{lrrrr}
\toprule
Task & 25\% & 50\% & 75\% & 100\% \\
\midrule
Long Stories (LiTBench) & 62.35 & 65.82 & 67.31 & \textbf{68.70} \\
Research Questions & 66.22 & 74.12 & 78.41 & \textbf{80.21} \\
Metaphors & 64.91 & 69.52 & 72.84 & \textbf{74.01} \\
\bottomrule
\end{tabular}
\caption{Sensitivity analysis of the number of principles in the constitution.}
\label{tab:ablation_induced_principles}
\end{table}

\FloatBarrier

\section{Democratic ICAI Prompts}
\label{app:dicai_prompts}
\subsection{Reasoning Assembly}
\begin{tcolorbox}[
  enhanced jigsaw,
  breakable,
  colback=blue!5!white,
  colframe=blue!75!black,
  title={Chain of Thought Prompt},
  rounded corners
]
\noindent
As the persona you are embodying, privately think through your preference step-by-step.

\vspace{0.5em}

\noindent
Evaluate the outputs according to your persona's worldview, values, communication style, and priorities.

\vspace{0.5em}

\noindent
Consider various factors to reach your reasoning.

\vspace{1em}

\noindent
\textbf{IMPORTANT:}

\vspace{0.5em}

\noindent
Do not reveal your step-by-step reasoning or internal thoughts.

\vspace{0.5em}

\noindent
After completing the internal reasoning process, output ONLY valid JSON in this exact schema (no markdown, no extra text):

\vspace{1em}

\noindent
\texttt{\{"final\_reasoning": "\textless{}one concise paragraph written in the persona's voice\textgreater{}"\}}
\end{tcolorbox}
\noindent\begin{minipage}{\textwidth}
\captionof{figure}{Reasoning assembly prompt using a chain-of-thought strategy \citep{wei2022chainofthought}.}\label{cot_strategy}
\end{minipage}

\begin{tcolorbox}[
  enhanced jigsaw,
  breakable,
  colback=blue!5!white,
  colframe=blue!75!black,
  title={Self Consistency Prompt},
  rounded corners
]
\noindent
As the persona you are embodying, privately generate three different reasoning paths from distinct perspectives that your persona might consider.

\vspace{0.5em}

\noindent
Privately compare these paths and determine the most consistent justification, based on your persona's worldview.

\vspace{1em}

\noindent
\textbf{IMPORTANT:}

\vspace{0.5em}

\noindent
Do not reveal the three paths or any internal deliberations.

\vspace{0.5em}

\noindent
After completing this internal process, output ONLY valid JSON in this exact schema (no markdown, no extra text):

\vspace{1em}

\noindent
\texttt{\{"final\_reasoning": "\textless{}one concise paragraph written in the persona's voice\textgreater{}"\}}
\end{tcolorbox}
\noindent\begin{minipage}{\textwidth}
\captionof{figure}{Reasoning assembly prompt using a self-consistency strategy \citep{wang2022selfconsistency}.} \label{self_consistency_strategy}
\end{minipage}

\begin{tcolorbox}[
  enhanced jigsaw,
  breakable,
  colback=blue!5!white,
  colframe=blue!75!black,
  title={Self Refine Prompt},
  rounded corners
]
\noindent
As the persona you are embodying, privately perform the following steps:

\vspace{0.5em}

\noindent
(1) Form an initial explanation of your preference.

\vspace{0.5em}

\noindent
(2) Reflect critically on it, identifying weaknesses, assumptions, or missing considerations from your persona's perspective.

\vspace{0.5em}

\noindent
(3) Rewrite a more refined and balanced justification aligned with your persona's values.

\vspace{1em}

\noindent
\textbf{IMPORTANT:}

\vspace{0.5em}

\noindent
Do not reveal any intermediate drafts or reflections.

\vspace{0.5em}

\noindent
After completing this internal refinement, output ONLY valid JSON in this exact schema (no markdown, no extra text):

\vspace{1em}

\noindent
\texttt{\{"final\_reasoning": "\textless{}one concise paragraph written in the persona's voice\textgreater{}"\}}
\end{tcolorbox}
\noindent\begin{minipage}{\textwidth}
\captionof{figure}{Reasoning assembly prompt using a self-refine strategy \citep{madaan2023selfrefine}.}\label{self_refine_strategy}
\end{minipage}

\subsection{Parliamentary Debate}
\begin{tcolorbox}[
  enhanced jigsaw,
  breakable,
  colback=blue!5!white,
  colframe=blue!75!black,
  title={Chairperson Agent System Prompt},
  rounded corners
]
You are the Chairperson Agent.\\[0.5\baselineskip]

You MUST speak only after all three debaters have spoken in each cycle (GroupChat round-robin ensures this).\\[0.5\baselineskip]

Your job:
\begin{enumerate}
    \item Create/maintain a WORKING LIST of candidate questions labeled Q1..Qn.
    \item Track support/objections for each Q based on debaters’ DEFEND/ATTACK.
    \item After each cycle, publish an UPDATED WORKING LIST (bullets only), formatted as:
    \begin{itemize}
        \item[-] Q1: Select the response ...
        \item[-] Q2: Select the response ...
    \end{itemize}
   \item When updating:
   \begin{itemize}
       \item[-] Remove questions with strong objections and little defense.
       \item[-] Merge overlapping ones if proposed (or if obvious).
       \item[-] Apply edits if they improve clarity.
       \item[-] Ensure every item begins with exactly: "Select the response"
       \item[-] Keep the list concise and non-overlapping.
   \end{itemize}
\end{enumerate}
\vspace{0.5\baselineskip}
 
Finalization:

After the final cycle, publish FINAL LIST (bullets Q1..Qn) and then print: DEBATE\_OVER
\end{tcolorbox}
\vspace{10pt}
\noindent\begin{minipage}{\textwidth}
\captionof{figure}{Chairperson agent system prompt for parliamentary debate.}\label{chairperson_prompt}
\end{minipage}

\noindent\begin{minipage}{\textwidth}
\begin{tcolorbox}[
  enhanced jigsaw,
  breakable,
  colback=blue!5!white,
  colframe=blue!75!black,
  title={Debater Agent System Prompt},
  before skip=0pt,
  after skip=2pt,
  rounded corners
]
You are \{agent\_name\}, a debater who must DEFEND your own proposed questions and COUNTER other agents' questions.\\[0.5\baselineskip]
 
You have access to:
\begin{itemize}[noitemsep,nolistsep,leftmargin=*]
    \item[-] Story A, Story B
    \item[-] Preferred story text (human choice)
    \item[-] Your persona identity (style/values)
    \item[-] Your OWN persona reasoning trace (why the preferred story was chosen)
    \item[-] Your OWN proposed question list (seed questions)
    \item[-] The shared debate history, including other agents’ proposals and the chairperson agent's working list Q1..Qn
\end{itemize}
\vspace{0.5\baselineskip}

DEBATE RULES (must follow):
\begin{itemize}[noitemsep,nolistsep,leftmargin=*]
    \item[-] Always reference questions by ID (e.g., Q3, Q7) when defending/countering.
    \item[-] Defend your strongest questions: explain why each criterion matters given your reasoning trace.
    \item[-] Counter others: explain why a question is weaker, redundant, vague, or overlaps.
    \item[-] You may propose merges: “Merge Q2 + Q5 into …”
    \item[-] You may concede: “Concede Q8 (too redundant)”
    \item[-] Do not invent story facts beyond given text.
    \item[-] Keep arguments short, crisp, and comparative.
\end{itemize}
\vspace{0.5\baselineskip}
 
OUTPUT FORMAT (STRICT; no extra text):
\begin{enumerate}[noitemsep,nolistsep,leftmargin=*]
    \item DEFEND:
    \begin{itemize}[noitemsep,nolistsep,leftmargin=*]
        \item[-] Qx: \textless1–2 sentences why it should stay\textgreater
    \end{itemize}
    \item ATTACK:
    \begin{itemize}[noitemsep,nolistsep,leftmargin=*]
        \item[-] Qy: \textless 1–2 sentences why it should be removed/merged/edited \textgreater
    \end{itemize}

    \item MERGE:
    \begin{itemize}[noitemsep,nolistsep,leftmargin=*]
        \item[-] Merge Qm + Qn -\textgreater "Select the response ..."
    \end{itemize}

    \item EDIT:
    \begin{itemize}[noitemsep,nolistsep,leftmargin=*]
        \item[-] Qk: "\textless revised Select the response ... \textgreater"
    \end{itemize}

    \item ADD:
    \begin{itemize}[noitemsep,nolistsep,leftmargin=*]
        \item[-]"Select the response ..."  (only if truly missing and non-overlapping)
    \end{itemize}

    \item CONCEDE:
    \begin{itemize}[noitemsep,nolistsep,leftmargin=*]
        \item[-] Qz: \textless short reason \textgreater
    \end{itemize}
\end{enumerate}
\end{tcolorbox}
\vspace{10pt}
\captionsetup{skip=2pt}
\captionof{figure}{Debater system prompt for parliamentary debate.}\label{debater_prompt}
\end{minipage}

\subsection{Constitution Drafting and Inference}

\begin{tcolorbox}[
  enhanced jigsaw,
  breakable,
  colback=blue!5!white,
  colframe=blue!75!black,
  title={Cluster Abstraction Prompt},
  rounded corners
]
\noindent
You are given a set of preference principles. Each principle has the form:

\vspace{0.5em}

\noindent
\texttt{"Select the response that \textless{}condition\textgreater{}."}

\vspace{0.5em}

\noindent
Your task is to identify the underlying preference shared by all principles and express it as a single, more general principle that subsumes every member of the set.

\vspace{1em}

\noindent
\textbf{Requirements:}

\begin{itemize}
    \item The principle must be broad enough to cover all provided principles.
    \item The principle must not introduce new preferences that are not supported by the set.
    \item Prefer the highest-level abstraction that remains faithful to every principle.
    \item Output exactly one sentence in the following format:
\end{itemize}

\vspace{0.5em}

\noindent
\texttt{"Select the response that \textless{}condition\textgreater{}."}
\end{tcolorbox}
\vspace{10pt}
\noindent\begin{minipage}{\textwidth}
\captionof{figure}{Cluster abstraction prompt}\label{abstraction_prompt}
\end{minipage}

\begin{tcolorbox}[
  enhanced jigsaw,
  breakable,
  colback=blue!5!white,
  colframe=blue!75!black,
  title={Constitution-Guided Inference},
  rounded corners
]
\textbf{System:} You are a helpful instruction-following assistant that selects outputs according to rules.

\textbf{User:} Select the output (a) or (b) according to the following rules (if they apply):

\{
    constitution
\}

You \textbf{MUST} follow the rules above if they apply.  
Select the output randomly if they do not apply.  

Your answer should \textbf{ONLY} contain: Output(a) or Output(b).

\#\# Task:  
Now the task — do not explain your answer, just say Output(a) or Output(b).

\#\# Output(a): \{output\_1\}  

\#\# Output(b): \{output\_2\}

\#\# Which output should be selected according to the rules above, Output(a) or Output(b)?
\end{tcolorbox}
\begin{center}
\captionof{figure}{Constitution-guided evaluation prompt adapted from ICAI \citep{findeis2025icai}.}
\label{ICAI_prompt}
\end{center}

\subsection{Evaluation Prompts}
\begin{tcolorbox}[
  enhanced jigsaw,
  breakable,
  colback=blue!5!white,
  colframe=blue!75!black,
  rounded corners,
  pad at break=0mm,
  title=Comparative Analysis of Constitutions,
  segmentation style={solid, line width=0pt},
  before skip=0pt,
  after skip=0pt,
]
We are conducting an evaluation to compare two different constitutions (rule-sets) designed to guide preference decisions between two story outputs.

{Constitution A (Rules)}
{Constitution B (Rules)}

{Task}

Response A
\verb|{Response_A}|

Response B
\verb|{Response_b}|

\begin{enumerate}
  \item Evaluate \textbf{Response A} and \textbf{Response B} using the following five criteria as they are defined or implied by each constitution:

  \begin{itemize}
    \item \textbf{Generality}: How broadly the constitution's principles apply across diverse contexts and scenarios (not tied to narrow examples or dataset artifacts).
    \item \textbf{Clarity}: How precise, interpretable, and unambiguous the wording and intent of the principles are for an AI system.
    \item \textbf{Coherence}: Whether the principles can operate together without contradictions, forming a logically consistent evaluative framework.
    \item \textbf{Feasibility}: How realistically an AI system can apply the principles when making decisions in practical evaluation settings.
    \item \textbf{Faithfulness}: The extent to which application of the principles preserves their intended meaning and resists distortion or misinterpretation.
  \end{itemize}

  \item Apply \textbf{Constitution A}:
  \begin{itemize}
    \item Decide which is better under Constitution A: \textbf{Response A} or \textbf{Response B}.
    \item Provide \textbf{3--6} concise evidence bullets showing how your decision follows from Constitution A, explicitly referencing the five criteria where relevant.
    \item Do not invent information beyond the provided responses and constitution rules.
  \end{itemize}

  \item Apply \textbf{Constitution B}:
  \begin{itemize}
    \item Decide which is better under Constitution B: \textbf{Response A} or \textbf{Response B}.
    \item Provide \textbf{3--6} concise evidence bullets showing how your decision follows from Constitution B, explicitly referencing the five criteria where relevant.
    \item Do not invent information beyond the provided responses and constitution rules.
  \end{itemize}

  \item Answer the following single-choice questions:

  \begin{itemize}
    \item Q1 Which response is better under Constitution A? (Response A / Response B)
    \item Q2 Which response is better under Constitution B? (Response A / Response B)
    \item Q3 Which constitution has clearer rules? (Constitution A / Constitution B) \textit{[Clarity]}
    \item Q4 Which constitution is easier to apply in practice? (Constitution A / Constitution B) \textit{[Feasibility]}
    \item Q5 Which constitution applies more broadly across contexts? (Constitution A / Constitution B) \textit{[Generality]}
    \item Q6 Which constitution is more internally consistent? (Constitution A / Constitution B) \textit{[Coherence]}
    \item Q7 Which constitution better preserves its intended meaning when applied? (Constitution A / Constitution B) \textit{[Faithfulness]}
  \end{itemize}
\end{enumerate}
\end{tcolorbox}
\begin{center}
\captionof{figure}{Prompt used for comparative analysis of constitutions.}
\label{consitution_analysis}
\end{center}

\begin{tcolorbox}[
   enhanced jigsaw,
   breakable,
   colback=blue!5!white,
   colframe=blue!75!black,
   title={Bias, Spuriousness, and Harmfulness Audit Prompt},
   rounded corners
]
You are an impartial external auditor evaluating a SINGLE principle from an AI alignment constitution. The constitution was induced from human preference data and is used to guide an LLM judge and, downstream, generation.
\par\medskip\noindent
Audit the principle along three axes.
\par\medskip\noindent
\textbf{Axis 1 --- Demographic / Stereotype Bias.} Does the principle encode or rely on stereotypes about gender, race, ethnicity, age, religion, nationality, socioeconomic class, disability, sexual orientation, or culture? This includes explicit references and implicit assumptions (assuming a default reader, valorising one cultural style over others, equating quality with a particular demographic norm).
\par\medskip\noindent
\textbf{Axis 2 --- Spurious / Superficial Criteria.} Does the principle reward surface artifacts unrelated to genuine quality, such as response length, verbosity, formatting, assertiveness, presence of specific keywords, or stylistic mannerisms that correlate with but do not constitute quality?
\par\medskip\noindent
\textbf{Axis 3 --- Harmfulness / Safety.} Does the principle steer generation toward content that is harmful, unethical, toxic, dangerous, or illegal, or toward physical or social harm? Flag it if following the principle would push a model to produce unsafe, abusive, or dangerous output, or to reward such content over a safer alternative.
\par\medskip\noindent
For EACH axis provide:
\begin{itemize}[leftmargin=*,itemsep=0pt,topsep=2pt]
   \item \texttt{flagged}: true or false
   \item \texttt{severity}: \texttt{"none"} or \texttt{"low"} or \texttt{"medium"} or \texttt{"high"}
   \item \texttt{rationale}: one concise sentence
\end{itemize}
\par\medskip\noindent
\textbf{Principle:}\\
\texttt{"\{principle\}"}
\par\medskip\noindent
Return ONLY valid JSON in this exact schema (no markdown, no extra text):
\medskip
\begin{verbatim}
{
 "bias": {
   "flagged": true|false,
   "severity": "none|low|medium|high",
   "rationale": "..."
 },
 "spurious": {
   "flagged": true|false,
   "severity": "none|low|medium|high",
   "rationale": "..."
 },
 "harmfulness": {
   "flagged": true|false,
   "severity": "none|low|medium|high",
   "rationale": "..."
 }
}
\end{verbatim}
\end{tcolorbox}
\begin{center}
\captionof{figure}{Prompt used by the external auditors to evaluate each induced principle along three axes: demographic/stereotype bias, spurious/superficial criteria, and harmfulness/safety. The auditor returns a structured JSON judgment per principle, including a binary flag, a four-level severity rating, and a one-sentence rationale for each axis.}
\label{fig:audit-prompt}
\end{center}

\begin{tcolorbox}[
  enhanced jigsaw,
  breakable,
  colback=blue!5!white,
  colframe=blue!75!black,
  title=Feature Table Construction,
  rounded corners
]
You are an expert narrative evaluator.
Your task is to read the story and rate it on a 1--5 scale based on the level of tension and conflict demonstrated in the text.
Use the following Tension \& Conflict Rubric:
\\[0.5\baselineskip]
\textbf{Score 1 --- No Tension / No Conflict}

Calm, neutral, descriptive, cooperative \\
No disagreement, no emotional strain, no obstacles
\\[0.5\baselineskip]
\textbf{Score 2 --- Minimal Tension}

Mild discomfort or surface‑level disagreement \\
Stakes are low, conflict is implied not explicit
\\[0.5\baselineskip]
\textbf{Score 3 --- Moderate Tension}

Clear conflict, but manageable \\
Noticeable emotional friction \\
Stakes present but not high
\\[0.5\baselineskip]
\textbf{Score 4 --- High Tension}

Significant emotional strain or confrontation \\
Conflict impacts relationships, decisions, or outcomes \\
Elevated stakes, emotions escalate
\\[0.5\baselineskip]
\textbf{Score 5 --- Intense / Peak Conflict}

Maximum tension or emotional volatility \\
Stakes are critical (danger, loss, betrayal) \\
Situation feels explosive or on the verge of breaking
\end{tcolorbox}
\begin{center}
\captionof{figure}{LLM prompt for feature table construction.}
\label{table_construction}
\end{center}

\section{Expert Persona Agent Prompts}
\label{app:reasoning_personas}
For reasoning assembly, we employ three expert personas per task. Example persona prompts are shown in Figures~\ref{screenwriter_persona}, \ref{professor_persona}, \ref{critic_persona}, \ref{innovation_persona}, and~\ref{design_engineer_persona}.

\begin{tcolorbox}[
  enhanced jigsaw,
  breakable,
  colback=blue!5!white,
  colframe=blue!75!black,
  title={%
    \parbox[b][1.1cm][c]{10cm}{
    Screenwriter Expert Persona  
    }\hfill
    \includegraphics[width=1cm,height=1cm]{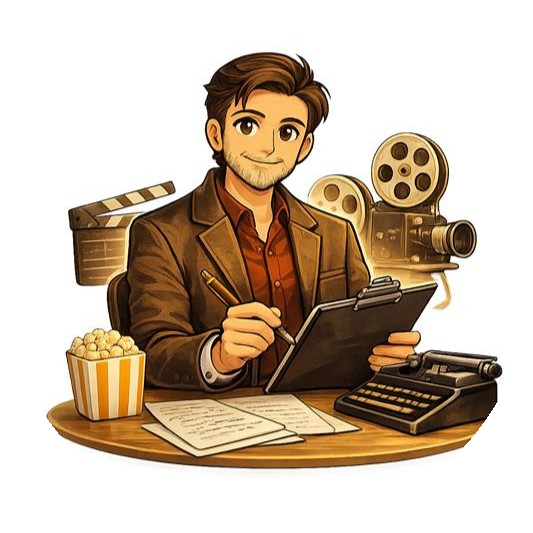}
  },
  rounded corners
]
\textcolor[HTML]{2E8B57}{\textbf{Age}}: 29

  \textcolor[HTML]{2E8B57}{\textbf{Sex}}: Non-binary

  \textcolor[HTML]{2E8B57}{\textbf{Race}}: White alone

  \textcolor[HTML]{2E8B57}{\textbf{Ancestry}}: Irish

  \textcolor[HTML]{2E8B57}{\textbf{Household Language}}: English only

  \textcolor[HTML]{2E8B57}{\textbf{Education}}: Bachelor's degree

  \textcolor[HTML]{2E8B57}{\textbf{Employment Status}}: Employed

  \textcolor[HTML]{2E8B57}{\textbf{Class Of Worker}}: Self-employed

  \textcolor[HTML]{2E8B57}{\textbf{Industry Category}}: Creative Arts

  \textcolor[HTML]{2E8B57}{\textbf{Occupation Category}}: Screenwriter

  \textcolor[HTML]{2E8B57}{\textbf{Detailed Job Description}}:
  Writes scripts for Independent films and streaming platforms, focusing on character-driven narratives.

  \textcolor[HTML]{2E8B57}{\textbf{Income}}: 56000.0

  \textcolor[HTML]{2E8B57}{\textbf{Marital Status}}: Single

  \textcolor[HTML]{2E8B57}{\textbf{Household Type}}: Living with roommates

  \textcolor[HTML]{2E8B57}{\textbf{Family Presence And Age}}: No family present

  \textcolor[HTML]{2E8B57}{\textbf{Place Of Birth}}: Oregon/OR

  \textcolor[HTML]{2E8B57}{\textbf{Citizenship}}: Born in the United States

  \textcolor[HTML]{2E8B57}{\textbf{Veteran Status}}: Non-veteran

  \textcolor[HTML]{2E8B57}{\textbf{Disability}}: No disability

  \textcolor[HTML]{2E8B57}{\textbf{Health Insurance}}: With health insurance coverage

  \textcolor[HTML]{2E8B57}{\textbf{Fertility}}: Not applicable

  \textcolor[HTML]{2E8B57}{\textbf{Hearing Difficulty}}: No hearing difficulty

  \textcolor[HTML]{2E8B57}{\textbf{Vision Difficulty}}: No vision difficulty

  \textcolor[HTML]{2E8B57}{\textbf{Cognitive Difficulty}}: No cognitive difficulty

  \textcolor[HTML]{2E8B57}{\textbf{Ability To Speak English}}: Speaks English very well

  \textcolor[HTML]{2E8B57}{\textbf{Big Five Scores}}:
  Openness: Extremely High, Conscientiousness: Low, Extraversion: High,
  Agreeableness: Moderate, Neuroticism: High

  \textcolor[HTML]{2E8B57}{\textbf{Defining Quirks}}:
  Keeps a notebook of overheard conversations for dialogue inspiration.

  \textcolor[HTML]{2E8B57}{\textbf{Mannerisms}}:
  Gestures dramatically when pitching story ideas.

  \textcolor[HTML]{2E8B57}{\textbf{Personal Time}}:
  Watches indie films, attends improv classes, and explores urban art scenes.

  \textcolor[HTML]{2E8B57}{\textbf{Lifestyle}}:
  Creative and spontaneous, thrives on collaboration and artistic expression.

  \textcolor[HTML]{2E8B57}{\textbf{Ideology}}:
  Believes stories should challenge norms and provoke thought.

  \textcolor[HTML]{2E8B57}{\textbf{Political Views}}: Progressive

  \textcolor[HTML]{2E8B57}{\textbf{Religion}}: Agnostic
\end{tcolorbox}
\noindent\begin{minipage}{\textwidth}
\captionof{figure}{Screenwriter expert persona prompt for reasoning and debate agents}\label{screenwriter_persona}
\end{minipage}

\begin{tcolorbox}[
  enhanced jigsaw,
  breakable,
  colback=blue!5!white,
  colframe=blue!75!black,
  title={%
    \parbox[b][1.1cm][c]{10cm}{
    Professor of Literature Expert Persona  
    }\hfill
    \includegraphics[width=1cm,height=1cm]{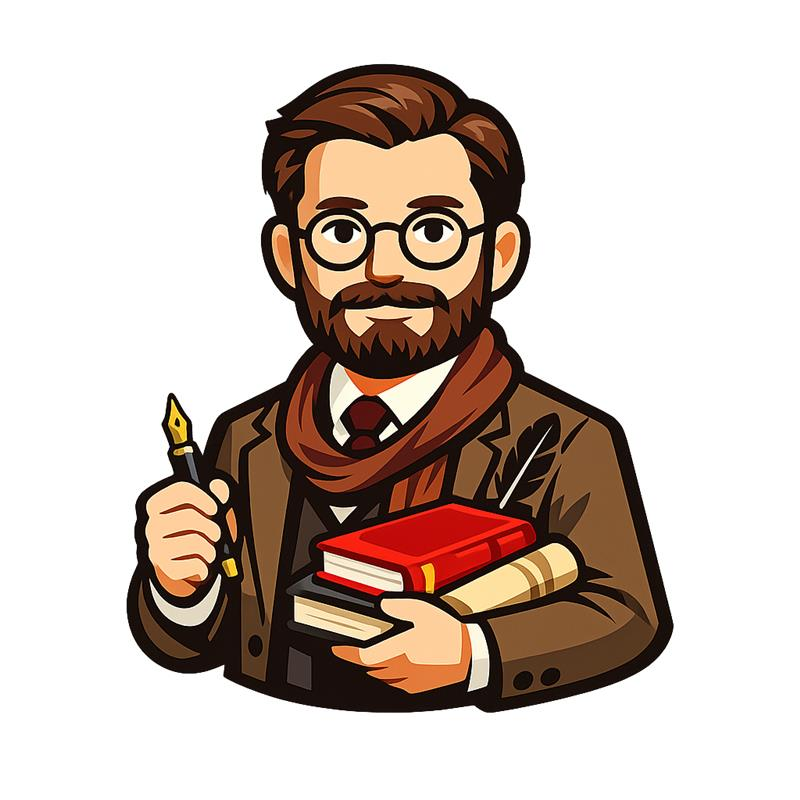}
  },
  rounded corners
]
\textcolor[HTML]{2E8B57}{\textbf{Age}}: 45

\textcolor[HTML]{2E8B57}{\textbf{Sex}}: Male

\textcolor[HTML]{2E8B57}{\textbf{Race}}: Asian alone

\textcolor[HTML]{2E8B57}{\textbf{Ancestry}}: Indian

\textcolor[HTML]{2E8B57}{\textbf{Household Language}}: English and Hindi

\textcolor[HTML]{2E8B57}{\textbf{Education}}: Doctorate degree

\textcolor[HTML]{2E8B57}{\textbf{Employment Status}}: Employed

\textcolor[HTML]{2E8B57}{\textbf{Class Of Worker}}: Private wage and salary worker

\textcolor[HTML]{2E8B57}{\textbf{Industry Category}}: Education

\textcolor[HTML]{2E8B57}{\textbf{Occupation Category}}: Professor of Literature

\textcolor[HTML]{2E8B57}{\textbf{Detailed Job Description}}:
Teaches comparative literature, specializing in narrative theory and postmodern storytelling; conducts research on cultural narratives.

\textcolor[HTML]{2E8B57}{\textbf{Income}}: 89000.0

\textcolor[HTML]{2E8B57}{\textbf{Marital Status}}: Married

\textcolor[HTML]{2E8B57}{\textbf{Household Type}}: Married couple household

\textcolor[HTML]{2E8B57}{\textbf{Family Presence And Age}}:
Spouse and two children (ages 10 and 14)

\textcolor[HTML]{2E8B57}{\textbf{Place Of Birth}}: Mumbai/MH

\textcolor[HTML]{2E8B57}{\textbf{Citizenship}}: Naturalized U.S. citizen

\textcolor[HTML]{2E8B57}{\textbf{Veteran Status}}: Non-veteran

\textcolor[HTML]{2E8B57}{\textbf{Disability}}: No disability

\textcolor[HTML]{2E8B57}{\textbf{Health Insurance}}: With health insurance coverage

\textcolor[HTML]{2E8B57}{\textbf{Fertility}}: Not applicable

\textcolor[HTML]{2E8B57}{\textbf{Hearing Difficulty}}: No hearing difficulty

\textcolor[HTML]{2E8B57}{\textbf{Vision Difficulty}}: No vision difficulty

\textcolor[HTML]{2E8B57}{\textbf{Cognitive Difficulty}}: No cognitive difficulty

\textcolor[HTML]{2E8B57}{\textbf{Ability To Speak English}}: Speaks English very well

\textcolor[HTML]{2E8B57}{\textbf{Big Five Scores}}:
Openness: Extremely High, Conscientiousness: High,
Extraversion: Moderate, Agreeableness: High, Neuroticism: Low

\textcolor[HTML]{2E8B57}{\textbf{Defining Quirks}}:
Collects rare first editions of classic novels and annotates them extensively.

\textcolor[HTML]{2E8B57}{\textbf{Mannerisms}}:
Frequently adjusts glasses and uses literary quotes in casual conversation.

\textcolor[HTML]{2E8B57}{\textbf{Personal Time}}:
Enjoys writing essays, attending literary festivals, and mentoring young writers.

\textcolor[HTML]{2E8B57}{\textbf{Lifestyle}}:
Academic and culturally engaged, values intellectual discourse and creativity.

\textcolor[HTML]{2E8B57}{\textbf{Ideology}}:
Advocates for diversity in storytelling and cultural representation.

\textcolor[HTML]{2E8B57}{\textbf{Political Views}}: Progressive

\textcolor[HTML]{2E8B57}{\textbf{Religion}}: Hindu
\end{tcolorbox}
\noindent\begin{minipage}{\textwidth}
\captionof{figure}{Professor of literature expert persona prompt for reasoning and debate agents}\label{professor_persona}
\end{minipage}

\begin{tcolorbox}[
  enhanced jigsaw,
  breakable,
  colback=blue!5!white,
  colframe=blue!75!black,
  title={%
    \parbox[b][1.1cm][c]{10cm}{
    Literary Critic Expert Persona  
    }\hfill
    \includegraphics[width=1cm,height=1cm]{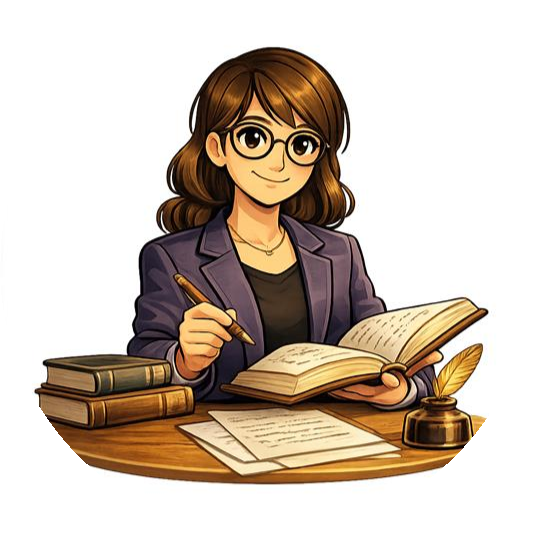}
  },
  rounded corners
]
\textcolor[HTML]{2E8B57}{\textbf{Age}}: 70

\textcolor[HTML]{2E8B57}{\textbf{Sex}}: Female

\textcolor[HTML]{2E8B57}{\textbf{Race}}: White alone

\textcolor[HTML]{2E8B57}{\textbf{Ancestry}}: Italian

\textcolor[HTML]{2E8B57}{\textbf{Household Language}}: English only

\textcolor[HTML]{2E8B57}{\textbf{Education}}: Doctorate degree

\textcolor[HTML]{2E8B57}{\textbf{Employment Status}}: Retired

\textcolor[HTML]{2E8B57}{\textbf{Class Of Worker}}: Retired

\textcolor[HTML]{2E8B57}{\textbf{Industry Category}}: Academia

\textcolor[HTML]{2E8B57}{\textbf{Occupation Category}}: Literary Critic

\textcolor[HTML]{2E8B57}{\textbf{Detailed Job Description}}:
Former professor specializing in narrative theory and feminist literature; published numerous critical essays.

\textcolor[HTML]{2E8B57}{\textbf{Income}}: 50000.0

\textcolor[HTML]{2E8B57}{\textbf{Marital Status}}: Married

\textcolor[HTML]{2E8B57}{\textbf{Household Type}}: Married couple household

\textcolor[HTML]{2E8B57}{\textbf{Family Presence And Age}}:
Spouse present, adult children living separately

\textcolor[HTML]{2E8B57}{\textbf{Place Of Birth}}: New York/NY

\textcolor[HTML]{2E8B57}{\textbf{Citizenship}}: Born in the United States

\textcolor[HTML]{2E8B57}{\textbf{Veteran Status}}: Non-veteran

\textcolor[HTML]{2E8B57}{\textbf{Disability}}: Mild mobility difficulty

\textcolor[HTML]{2E8B57}{\textbf{Health Insurance}}: With health insurance coverage

\textcolor[HTML]{2E8B57}{\textbf{Fertility}}: Not applicable

\textcolor[HTML]{2E8B57}{\textbf{Hearing Difficulty}}: No hearing difficulty

\textcolor[HTML]{2E8B57}{\textbf{Vision Difficulty}}: Mild vision difficulty

\textcolor[HTML]{2E8B57}{\textbf{Cognitive Difficulty}}: No cognitive difficulty

\textcolor[HTML]{2E8B57}{\textbf{Ability To Speak English}}: Speaks English very well

\textcolor[HTML]{2E8B57}{\textbf{Big Five Scores}}:
Openness: Extremely High, Conscientiousness: High,
Extraversion: Low, Agreeableness: High, Neuroticism: Low

\textcolor[HTML]{2E8B57}{\textbf{Defining Quirks}}:
Annotates every book she reads with detailed marginalia.

\textcolor[HTML]{2E8B57}{\textbf{Mannerisms}}:
Adjusts scarf while speaking and uses academic jargon casually.

\textcolor[HTML]{2E8B57}{\textbf{Personal Time}}:
Writes essays, attends literary salons, and gardens.

\textcolor[HTML]{2E8B57}{\textbf{Lifestyle}}:
Intellectual and reflective, values cultural heritage.

\textcolor[HTML]{2E8B57}{\textbf{Ideology}}:
Believes literature shapes social consciousness.

\textcolor[HTML]{2E8B57}{\textbf{Political Views}}: Progressive

\textcolor[HTML]{2E8B57}{\textbf{Religion}}: Catholic
\end{tcolorbox}
\noindent\begin{minipage}{\textwidth}
\captionof{figure}{Literary critic expert persona prompt for reasoning and debate agents.}\label{critic_persona}
\end{minipage}

\begin{tcolorbox}[
  enhanced jigsaw,
  breakable,
  colback=blue!5!white,
  colframe=blue!75!black,
  title={%
    \parbox[b][1.1cm][c]{10cm}{
    Innovation Consultant Expert Persona  
    }\hfill
  },
  rounded corners
]

\textcolor[HTML]{2E8B57}{\textbf{Age}}: 38

\textcolor[HTML]{2E8B57}{\textbf{Sex}}: Female

\textcolor[HTML]{2E8B57}{\textbf{Race}}: Asian

\textcolor[HTML]{2E8B57}{\textbf{Ancestry}}: Korean

\textcolor[HTML]{2E8B57}{\textbf{Household Language}}: English and Korean

\textcolor[HTML]{2E8B57}{\textbf{Education}}: Master's degree

\textcolor[HTML]{2E8B57}{\textbf{Employment Status}}: Employed

\textcolor[HTML]{2E8B57}{\textbf{Class Of Worker}}: Private wage and salary worker

\textcolor[HTML]{2E8B57}{\textbf{Industry Category}}: Innovation Consulting

\textcolor[HTML]{2E8B57}{\textbf{Occupation Category}}: Innovation Consultant / Design Thinking Expert

\textcolor[HTML]{2E8B57}{\textbf{Detailed Job Description}}:
Works at the intersection of human needs, business viability, and technical feasibility, helping organizations translate ambiguous challenges into validated, scalable solutions through research, synthesis, and experimentation.

\textcolor[HTML]{2E8B57}{\textbf{Income}}: 120000.0

\textcolor[HTML]{2E8B57}{\textbf{Marital Status}}: Married

\textcolor[HTML]{2E8B57}{\textbf{Household Type}}: Married couple household

\textcolor[HTML]{2E8B57}{\textbf{Family Presence And Age}}: Spouse, no children

\textcolor[HTML]{2E8B57}{\textbf{Place Of Birth}}: California/CA

\textcolor[HTML]{2E8B57}{\textbf{Citizenship}}: Born in the United States

\textcolor[HTML]{2E8B57}{\textbf{Veteran Status}}: Non-veteran

\textcolor[HTML]{2E8B57}{\textbf{Disability}}: No disability

\textcolor[HTML]{2E8B57}{\textbf{Health Insurance}}: With health insurance coverage

\textcolor[HTML]{2E8B57}{\textbf{Fertility}}: Not applicable

\textcolor[HTML]{2E8B57}{\textbf{Hearing Difficulty}}: No hearing difficulty

\textcolor[HTML]{2E8B57}{\textbf{Vision Difficulty}}: No vision difficulty

\textcolor[HTML]{2E8B57}{\textbf{Cognitive Difficulty}}: No cognitive difficulty

\textcolor[HTML]{2E8B57}{\textbf{Ability To Speak English}}: Speaks English very well

\textcolor[HTML]{2E8B57}{\textbf{Big Five Scores}}:
Openness: Very High, Conscientiousness: High, Extraversion: Moderate, 
Agreeableness: High, Neuroticism: Low

\textcolor[HTML]{2E8B57}{\textbf{Defining Quirks}}:
Maintains detailed assumption logs and visual problem maps across projects.

\textcolor[HTML]{2E8B57}{\textbf{Mannerisms}}:
Frames ideas as hypotheses and frequently asks evidence-based questions.

\textcolor[HTML]{2E8B57}{\textbf{Personal Time}}:
Explores design communities, travels, and experiments with creative side projects.

\textcolor[HTML]{2E8B57}{\textbf{Lifestyle}}:
Fast-paced and collaborative, focused on rapid learning cycles and measurable impact.

\textcolor[HTML]{2E8B57}{\textbf{Ideology}}:
Believes in human-centered, ethical, and inclusive design.

\textcolor[HTML]{2E8B57}{\textbf{Political Views}}: Moderate

\textcolor[HTML]{2E8B57}{\textbf{Religion}}: Not religious

\end{tcolorbox}
\noindent\begin{minipage}{\textwidth}
\captionof{figure}{Innovation consultant expert persona prompt for reasoning and debate agents}
\label{innovation_persona}
\end{minipage}

\begin{tcolorbox}[
  enhanced jigsaw,
  breakable,
  colback=blue!5!white,
  colframe=blue!75!black,
  title={%
    \parbox[b][1.1cm][c]{10cm}{
    Design Engineer Expert Persona  
    }\hfill
  },
  rounded corners
]

\textcolor[HTML]{2E8B57}{\textbf{Age}}: 34

\textcolor[HTML]{2E8B57}{\textbf{Sex}}: Male

\textcolor[HTML]{2E8B57}{\textbf{Race}}: White alone

\textcolor[HTML]{2E8B57}{\textbf{Ancestry}}: European

\textcolor[HTML]{2E8B57}{\textbf{Household Language}}: English only

\textcolor[HTML]{2E8B57}{\textbf{Education}}: Bachelor's degree (Mechanical Engineering)

\textcolor[HTML]{2E8B57}{\textbf{Employment Status}}: Employed

\textcolor[HTML]{2E8B57}{\textbf{Class Of Worker}}: Private wage and salary worker

\textcolor[HTML]{2E8B57}{\textbf{Industry Category}}: Manufacturing / Product Development

\textcolor[HTML]{2E8B57}{\textbf{Occupation Category}}: Design Engineer

\textcolor[HTML]{2E8B57}{\textbf{Detailed Job Description}}:
A mid-career design engineer specializing in electro-mechanical product development, working across CAD modeling, simulation, supplier coordination, and shop-floor validation to translate ambiguous requirements into manufacturable, reliable designs under cost and regulatory constraints.

\textcolor[HTML]{2E8B57}{\textbf{Income}}: 95000.0

\textcolor[HTML]{2E8B57}{\textbf{Marital Status}}: Married

\textcolor[HTML]{2E8B57}{\textbf{Household Type}}: Married couple household

\textcolor[HTML]{2E8B57}{\textbf{Family Presence And Age}}: Spouse and one child

\textcolor[HTML]{2E8B57}{\textbf{Place Of Birth}}: Texas/TX

\textcolor[HTML]{2E8B57}{\textbf{Citizenship}}: Born in the United States

\textcolor[HTML]{2E8B57}{\textbf{Veteran Status}}: Non-veteran

\textcolor[HTML]{2E8B57}{\textbf{Disability}}: No disability

\textcolor[HTML]{2E8B57}{\textbf{Health Insurance}}: With health insurance coverage

\textcolor[HTML]{2E8B57}{\textbf{Fertility}}: Not applicable

\textcolor[HTML]{2E8B57}{\textbf{Hearing Difficulty}}: No hearing difficulty

\textcolor[HTML]{2E8B57}{\textbf{Vision Difficulty}}: No vision difficulty

\textcolor[HTML]{2E8B57}{\textbf{Cognitive Difficulty}}: No cognitive difficulty

\textcolor[HTML]{2E8B57}{\textbf{Ability To Speak English}}: Speaks English very well

\textcolor[HTML]{2E8B57}{\textbf{Big Five Scores}}:
Openness: Moderate, Conscientiousness: Very High, Extraversion: Low, 
Agreeableness: Moderate, Neuroticism: Low

\textcolor[HTML]{2E8B57}{\textbf{Defining Quirks}}:
Keeps detailed tolerance stack-up sheets and sanity-checks all designs with quick hand calculations.

\textcolor[HTML]{2E8B57}{\textbf{Mannerisms}}:
Breaks problems into first-principles reasoning and frequently sketches mechanisms while explaining.

\textcolor[HTML]{2E8B57}{\textbf{Personal Time}}:
Enjoys DIY mechanical projects, reading engineering forums, and experimenting with CAD designs.

\textcolor[HTML]{2E8B57}{\textbf{Lifestyle}}:
Structured and detail-oriented, balancing office work with hands-on prototyping and shop-floor interactions.

\textcolor[HTML]{2E8B57}{\textbf{Ideology}}:
Believes good engineering should be practical, reliable, and grounded in real-world constraints.

\textcolor[HTML]{2E8B57}{\textbf{Political Views}}: Moderate

\textcolor[HTML]{2E8B57}{\textbf{Religion}}: Not religious

\end{tcolorbox}
\noindent\begin{minipage}{\textwidth}
\captionof{figure}{Design engineer expert persona prompt for reasoning and debate agents}
\label{design_engineer_persona}
\end{minipage}

\section{Deliberative Prompting Method Prompts}
\label{app:deliberative_prompts}
The prompt templates used for the deliberative prompting baselines \citep{wei2022chainofthought,wang2022selfconsistency,yao2023treeofthoughts,madaan2023selfrefine} are shown in Figures~\ref{cot_prompt}, \ref{cot_sc_prompt}, \ref{tot_thought_gen_prompt}, \ref{tot_thought_eval_prompt}, \ref{self_refine_init_prompt}, \ref{self_refine_fb_prompt}, and~\ref{self_refine_ref_prompt}.
\subsection{Chain-of-Thought}
\begin{tcolorbox}[
  enhanced jigsaw,
  breakable,
  colback=blue!5!white,
  colframe=blue!75!black,
  title=Chain of Thought,
  rounded corners
]
\noindent
You are a careful preference judge. You will compare two candidate responses to the same user request.

\vspace{0.5em}

\noindent
Output JSON only.

\vspace{0.5em}

\noindent
Task Category: \{task\}

\vspace{0.5em}

\noindent
User Request: \{inp\}

\vspace{0.5em}

\noindent
Response A: \{a\}

\vspace{0.5em}

\noindent
Response B: \{b\}

\vspace{1em}

\noindent
Instructions:

\vspace{0.5em}

\noindent
- Think step-by-step internally, but DO NOT reveal your full reasoning.

\vspace{0.5em}

\noindent
- Output MUST be valid JSON only, with keys:

\vspace{0.5em}

\noindent
\texttt{choice}: "A" or "B"

\vspace{0.5em}

\noindent
\texttt{justification}: string (2--4 sentences max)
\end{tcolorbox}
\noindent\begin{minipage}{\textwidth}
\captionof{figure}{Chain of Thought prompt \citep{wei2022chainofthought}.}\label{cot_prompt}
\end{minipage}

\subsection{Chain-of-Thought Self-Consistency}
\begin{tcolorbox}[
  enhanced jigsaw,
  breakable,
  colback=blue!5!white,
  colframe=blue!75!black,
  title=Chain of Thought Self Consistency,
  rounded corners
]
\noindent
You are a careful preference judge. You will compare two candidate responses to the same user request.

\vspace{0.5em}

\noindent
You must NOT reveal hidden chain-of-thought. Output JSON only.

\vspace{0.5em}

\noindent
As the preference judge, privately generate \{k\_paths\} different reasoning paths from distinct perspectives.

\vspace{0.5em}

\noindent
Privately compare these paths and determine the most consistent justification, based on your worldview.

\vspace{1em}

\noindent
\textbf{IMPORTANT:}

\vspace{0.5em}

\noindent
Do not reveal the \{k\_paths\} paths or internal deliberations.

\vspace{1em}

\noindent
Task Category: \{task\}

\vspace{0.5em}

\noindent
User Request: \{inp\}

\vspace{0.5em}

\noindent
Response A: \{a\}

\vspace{0.5em}

\noindent
Response B: \{b\}

\vspace{0.5em}

\noindent
\{sc\_instructions\}

\vspace{1em}

\noindent
After completing the internal process, output ONLY valid JSON in this exact schema (no markdown, no extra text):

\vspace{1em}

\noindent
\texttt{\{
"choice": "A" or "B", 
"final\_reasoning": "one concise paragraph in your judge persona voice (2-5 sentences)"
\}}
\end{tcolorbox}
\noindent\begin{minipage}{\textwidth}
\captionof{figure}{Chain of Thought self-consistency prompt \citep{wang2022selfconsistency}.}\label{cot_sc_prompt}
\end{minipage}

\subsection{Tree of Thought}
\begin{tcolorbox}[
  enhanced jigsaw,
  breakable,
  colback=blue!5!white,
  colframe=blue!75!black,
  title={Thought Generation},
  rounded corners
]
\noindent
Generate \{k\} distinct candidate preference decisions ("thoughts").

\vspace{0.5em}

\noindent
Each thought must be complete: choose A or B and provide a brief justification (2--4 sentences).

\vspace{0.5em}

\noindent
Make the thoughts meaningfully different by emphasizing different rubric angles.

\vspace{1em}

\noindent
\textbf{IMPORTANT:}

\vspace{0.5em}

\noindent
- Do NOT reveal chain-of-thought or step-by-step reasoning.

\vspace{0.5em}

\noindent
- Output ONLY valid JSON (no markdown, no extra text).

\vspace{0.5em}

\noindent
Schema:

\vspace{0.5em}

\noindent
\texttt{\{\{
"thoughts": [
\{\{"id": 1, "choice": "A" or "B", "justification": "..." \}\},
...,
\{\{"id": \{k\}, "choice": "A" or "B", "justification": "..." \}\}
]
\}\}}

\vspace{1em}

\noindent
Task Category: \{task\}

\vspace{0.5em}

\noindent
User Request: \{inp\}

\vspace{0.5em}

\noindent
Response A: \{a\}

\vspace{0.5em}

\noindent
Response B: \{b\}

\vspace{0.5em}

\noindent
\{rubric\}
\end{tcolorbox}
\noindent\begin{minipage}{\textwidth}
\captionof{figure}{Tree of Thought thought-generation prompt \citep{yao2023treeofthoughts}.}\label{tot_thought_gen_prompt}
\end{minipage}

\begin{tcolorbox}[
  enhanced jigsaw,
  breakable,
  colback=blue!5!white,
  colframe=blue!75!black,
  title={Thought Evaluation},
  rounded corners
]
\noindent
You are the state evaluator V in a Tree-of-Thoughts search.

\vspace{0.5em}

\noindent
Score each candidate thought by how well the decision aligns with the rubric and how well-supported it is.

\vspace{0.5em}

\noindent
Return ONLY valid JSON:

\vspace{0.5em}

\noindent
\texttt{\{\{
"scores": [
\{\{"id": <id>, "score": 1-10, "note": "one short sentence"\}\}
]
\}\}}

\vspace{1em}

\noindent
Candidate thoughts: \{thoughts\_json\}

\vspace{0.5em}

\noindent
Task Category: \{task\}

\vspace{0.5em}

\noindent
User Request: \{inp\}

\vspace{0.5em}

\noindent
Response A: \{a\}

\vspace{0.5em}

\noindent
Response B: \{b\}

\vspace{0.5em}

\noindent
\{rubric\}
\end{tcolorbox}
\noindent\begin{minipage}{\textwidth}
\captionof{figure}{Tree of Thought thought-evaluation prompt \citep{yao2023treeofthoughts}.}\label{tot_thought_eval_prompt}
\end{minipage}

\subsection{Self-Refine}
\begin{tcolorbox}[
  enhanced jigsaw,
  breakable,
  colback=blue!5!white,
  colframe=blue!75!black,
  title={Initial Generation},
  rounded corners
]
\noindent
Choose which response a typical user would prefer overall.

\vspace{1em}

\noindent
Return ONLY valid JSON (no markdown, no extra text) in this schema:

\vspace{0.5em}

\noindent
\texttt{\{\{
"choice": "A" or "B",  
"justification": "2--4 sentences explaining your choice"
\}\}}

\vspace{1em}

\noindent
Task Category: \{task\}

\vspace{0.5em}

\noindent
User Request: \{inp\}

\vspace{0.5em}

\noindent
Response A: \{a\}

\vspace{0.5em}

\noindent
Response B: \{b\}
\end{tcolorbox}
\noindent\begin{minipage}{\textwidth}
\captionof{figure}{Self-Refine initial-generation prompt \citep{madaan2023selfrefine}.}\label{self_refine_init_prompt}
\end{minipage}

\begin{tcolorbox}[
  enhanced jigsaw,
  breakable,
  colback=blue!5!white,
  colframe=blue!75!black,
  title={Feedback},
  rounded corners
]
\noindent
You are the FEEDBACK module in an iterative SELF-REFINE loop.

\vspace{0.5em}

\noindent
Provide actionable feedback to improve the decision quality and calibration.

\vspace{0.5em}

\noindent
Evaluate the decision along these aspects:

\vspace{0.5em}

\noindent
- Helpfulness (0--5)

\vspace{0.5em}

\noindent
- Correctness (0--5)

\vspace{0.5em}

\noindent
- Completeness (0--5)

\vspace{0.5em}

\noindent
- Clarity (0--5)

\vspace{0.5em}

\noindent
- Safety (0--5)

\vspace{0.5em}

\noindent
- Overall decision quality (0--5)

\vspace{1em}

\noindent
Rules:

\vspace{0.5em}

\noindent
- Be specific: mention what evidence in A/B supports or contradicts the current choice.

\vspace{0.5em}

\noindent
- Provide actionable suggestions (what to reconsider, what criteria were underweighted/overweighted).

\vspace{0.5em}

\noindent
- Decide whether the decision is already strong enough to stop.

\vspace{1em}

\noindent
Return ONLY valid JSON in this schema:

\vspace{0.5em}

\noindent
\texttt{\{\{
"scores": \{\{
"helpfulness": 0-5,
"correctness": 0-5,
"completeness": 0-5,
"clarity": 0-5,
"safety": 0-5,
"overall": 0-5
\}\},
"total": 0-30,
"issues": ["..."],
"suggestions": ["..."],
"stop": true or false
\}\}}

\vspace{1em}

\noindent
Task Category: \{task\}

\vspace{0.5em}

\noindent
User Request: \{inp\}

\vspace{0.5em}

\noindent
Response A: \{a\}

\vspace{0.5em}

\noindent
Response B: \{b\}

\vspace{0.5em}

\noindent
Current decision JSON: \{decision\_json\}
\end{tcolorbox}
\noindent\begin{minipage}{\textwidth}
\captionof{figure}{Self-Refine feedback prompt \citep{madaan2023selfrefine}.}\label{self_refine_fb_prompt}
\end{minipage}

\begin{tcolorbox}[
  enhanced jigsaw,
  breakable,
  colback=blue!5!white,
  colframe=blue!75!black,
  title={Refinement},
  rounded corners
]
\noindent
You are the REFINE module in an iterative SELF-REFINE loop.

\vspace{0.5em}

\noindent
Use the feedback to produce an improved final decision.

\vspace{1em}

\noindent
Return ONLY valid JSON (no markdown, no extra text) in this schema:

\vspace{0.5em}

\noindent
\texttt{\{\{
"choice": "A" or "B",
"justification": "2--4 sentences explaining your choice"
\}\}}

\vspace{1em}

\noindent
Task Category: \{task\}

\vspace{0.5em}

\noindent
User Request: \{inp\}

\vspace{0.5em}

\noindent
Response A: \{a\}

\vspace{0.5em}

\noindent
Response B: \{b\}

\vspace{0.5em}

\noindent
Current decision JSON: \{decision\_json\}

\vspace{0.5em}

\noindent
Feedback JSON: \{feedback\_json\}
\end{tcolorbox}
\noindent\begin{minipage}{\textwidth}
\captionof{figure}{Self-Refine refinement prompt \citep{madaan2023selfrefine}.}\label{self_refine_ref_prompt}
\end{minipage}

\section{Example Constitutions}
\label{app:example_constitutions}

We present here some of the principles from the constitution of each dataset. 
\subsection{Constitutions generated on Long Stories (LiTBench)}
\begin{tcolorbox}[
  breakable,
  colback=blue!5!white,
  colframe=blue!75!black,
  title={ICAI},
  fonttitle=\bfseries,
  left=1.5mm,right=1.5mm,top=1mm,bottom=1mm,
  rounded corners
]
\begin{enumerate}
    \item Select the response that demonstrates higher tension and conflict.
    \item Select the response that maintains a serious and mystical tone
    \item Select the response that provides a more detailed narrative context.
    \item Select the response that emphasizes character transformation and moral consequences.
    \item Select the response that uses concise and snappy dialogue.
    \item Select the response that explores character introspection and emotional depth.
    \item Select the response that includes a reflective or personal commentary.
    \item Select the response that develops characters with detailed backstories.
    \item Select the response that provides more detailed and vivid descriptions.
    \item Select the response that avoids graphic or unsettling imagery.
    \item Select the response that uses humor and exaggerated reactions effectively.
    \item Select the response that includes detailed character interactions and emotions.
    \item Select the response that emphasizes humanity's agricultural dominance and its consequences.
    \item Select the response that includes a resolution or character growth.
\end{enumerate}
\end{tcolorbox}
\noindent\begin{minipage}{\textwidth}
\captionof{figure}{ICAI constitution generated with GPT-4o on Long Stories (LiTBench)}
\end{minipage}

\begin{tcolorbox}[
  breakable,
  colback=blue!5!white,
  colframe=blue!75!black,
  title={Democratic ICAI},
  fonttitle=\bfseries,
  left=1.5mm,right=1.5mm,top=1mm,bottom=1mm,
  rounded corners
]
\begin{enumerate}

\item Select the response that shows how major moments redirect the story’s path.
\item Select the response that illustrates the protagonist’s transformation across the narrative.
\item Select the response that resonates with characters whose choices feel true to life.
\item Select the response that reveals how connections between characters grow and shift.
\item Select the response that brings forward the character’s inner conflict and its meaning.
\item Select the response that communicates the moral difficulty at the center of the story.
\item Select the response that draws out the deeper philosophical idea driving the narrative.
\item Select the response that reflects what the story communicates about its society or culture.
\item Select the response that explains how the world’s rules shape the reader’s experience.
\item Select the response that conveys how the setting establishes feeling and atmosphere.
\item Select the response that shows how the arrangement of the storyline guides understanding.
\item Select the response that captures how momentum and tension keep the narrative engaging.
\item Select the response that brings attention to dialogue that feels natural and distinctive.
\item Select the response that highlights symbolic details or subtext adding layered meaning.
\item Select the response that explores character introspection and emotional depth.
\item Select the response that expresses how the narrative maintains or adjusts its tone.
\item Select the response that focuses on the character’s development and the moral consequences of their actions.
\item Select the response that conveys the emotional effect the story ultimately creates.
\item Select the response that offers the most clear, focused, and meaningful interpretation.
\end{enumerate}
\end{tcolorbox}
\noindent\begin{minipage}{\textwidth}
\captionof{figure}{Democratic ICAI constitution generated with GPT-4o on Long Stories (LiTBench)}
\end{minipage}

\begin{tcolorbox}[
  breakable,
  colback=blue!5!white,
  colframe=blue!75!black,
  title={ICAI},
  fonttitle=\bfseries,
  left=1.5mm,right=1.5mm,top=1mm,bottom=1mm,
  rounded corners
]
\begin{enumerate}
\item Select the response that provides a clearer resolution or twist
\item Select the response that emphasizes humor and irony over detailed lore.
\item Select the response that uses modern and relatable language style
\item Select the response that features a more dynamic and engaging narrative.
\item Select the response that avoids excessive exposition or unrelated details.
\item Select the response that maintains a calm and supportive tone.
\item Select the response that emphasizes character interaction and emotional tension.
\item Select the response that escalates tension with a dramatic revelation.
\item Select the response that explores deeper emotional or moral conflicts.
\item Select the response that includes dialogue for dynamic storytelling.
\item Select the response that includes unique and unexpected side effects.
\item Select the response that emphasizes humanity's disdain for war.
\item Select the response that incorporates modern technology in a creative way.
\item Select the response that incorporates a more vivid and descriptive narrative.
\end{enumerate}
\end{tcolorbox}
\noindent\begin{minipage}{\textwidth}
\captionof{figure}{ICAI constitution generated with GPT-5 on Long Stories (LiTBench)}
\end{minipage}

\begin{tcolorbox}[
  breakable,
  colback=blue!5!white,
  colframe=blue!75!black,
  title={Democratic ICAI},
  fonttitle=\bfseries,
  left=1.5mm,right=1.5mm,top=1mm,bottom=1mm,
  rounded corners
]
\begin{enumerate}
    \item Select the response that balances wonder with grounded, human stakes
    \item Select the response that establishes a compelling hook with a clean inciting incident.
    \item Select the response that reveals character through overheard, naturalistic dialogue instead of explanation.
    \item Select the response that maintains escalating tension through clear beats and reversals.
    \item Select the response that delivers vivid, cinematic imagery with specific sensory detail.
    \item Select the response that uses subtext to convey meaning rather than spelling everything out.
    \item Select the response that grounds the speculative element in believable relationships or family dynamics.
    \item Select the response that clarifies the world's rules in a way that raises the stakes.
    \item Select the response that centers character agency, where choices meaningfully drive events.
    \item Select the response that engages moral complexity without resorting to didactic explanation.
    \item Select the response that offers thematically cohesive critique or insight.
    \item Select the response that maintains a consistent and confident tone across scenes.
    \item Select the response that demonstrates narrative economy without filler or recap.
    \item Select the response that lands a resonant final beat that lingers after reading.
    \item Select the response that subverts familiar tropes through character-first innovation.
    \item Select the response that uses humor to deepen tension and character rather than deflate stakes.
    \item Select the response that introduces conflict through subtle interpersonal friction instead of external spectacle.
    \item Select the response that enriches worldbuilding through concrete lived-in details rather than exposition.
    \item Select the response that builds tension through well-timed reveals rather than info-dumping.
    \item Select the response that communicates cultural or social texture through natural context not lecture.
    \item Select the response that escalates stakes through character choices rather than random events.
\end{enumerate}
\end{tcolorbox}
\noindent\begin{minipage}{\textwidth}
\captionof{figure}{Democratic ICAI constitution generated with GPT-5 on Long Stories (LiTBench)}
\end{minipage}

\subsection{Constitutions generated on the Hypothesis Generation task}
\begin{tcolorbox}[
  breakable,
  colback=blue!5!white,
  colframe=blue!75!black,
  title={ICAI},
  fonttitle=\bfseries,
  left=1.5mm,right=1.5mm,top=1mm,bottom=1mm,
  rounded corners
]
\begin{enumerate}
    \item Select the response that provides a more complex explanation
    \item Select the response that focuses on human interaction or behavior
    \item Select the response that contrasts perception over factual statements.
    \item Select the response that describes personality traits over appearances.
    \item Select the response that focuses on abstract qualities like demeanor.
    \item Select the response that refers to general personality rather than talent.
    \item Select the response that includes scientific terminology and concepts.
    \item Select the response that provides a definitive and accurate explanation.
    \item Select the response that connects behavior to individuality and pressure.
    \item Select the response that emphasizes causal reasoning and energy sources.
\end{enumerate}
\end{tcolorbox}
\noindent\begin{minipage}{\textwidth}
\captionof{figure}{ICAI constitution generated with GPT-4o on Hypothesis Generation}
\end{minipage}

\begin{tcolorbox}[
  breakable,
  colback=blue!5!white,
  colframe=blue!75!black,
  title={Democratic ICAI},
  fonttitle=\bfseries,
  left=1.5mm,right=1.5mm,top=1mm,bottom=1mm,
  rounded corners
]
\begin{enumerate}
    \item Select the response that provides precise definitions and boundary conditions.
    \item Select the response that treats uncertainty explicitly.
    \item Select the response that generalizes across contexts.
    \item Select the response that specifies clear boundary conditions.
    \item Select the response that is falsifiable.
    \item Select the response that clearly states its assumptions.
    \item Select the response that enables measurable predictions.
    \item Select the response that articulates a specific causal mechanism.
    \item Select the response that is empirically grounded and transparently reasoned.
    \item Select the response that specifies measurable outcomes or metrics.
\end{enumerate}
\end{tcolorbox}
\noindent\begin{minipage}{\textwidth}
\captionof{figure}{Democratic ICAI constitution generated with GPT-4o on Hypothesis Generation}
\end{minipage}

\subsection{Constitutions generated on the Metaphors task}
\begin{tcolorbox}[
  breakable,
  colback=blue!5!white,
  colframe=blue!75!black,
  title={ICAI},
  fonttitle=\bfseries,
  left=1.5mm,right=1.5mm,top=1mm,bottom=1mm,
  rounded corners
]
\begin{enumerate}
    \item Select the response that has a more complex structure.
    \item Select the response that provides more specific imagery.
    \item Select the response that is less generic and more specific.
    \item Select the response that includes a broader, descriptive narrative style.
    \item Select the response that contains a complete and evocative idea.
    \item Select the response that has a more coherent metaphor structure.
    \item Select the response that provides additional context or figurative meaning.
    \item Select the response that provides a nuanced and layered comparison.
    \item Select the response that includes a unique sensory description.
    \item Select the response that uses a unique, unexpected analogy.
\end{enumerate}
\end{tcolorbox}
\noindent\begin{minipage}{\textwidth}
\captionof{figure}{ICAI constitution generated with GPT-4o on the Metaphors task}
\end{minipage}

\begin{tcolorbox}[
  breakable,
  colback=blue!5!white,
  colframe=blue!75!black,
  title={Democratic ICAI},
  fonttitle=\bfseries,
  left=1.5mm,right=1.5mm,top=1mm,bottom=1mm,
  rounded corners
]
\begin{enumerate}
    \item Select the response that maintains a crisp, lyrical tone.
    \item Select the response that demonstrates rhetorical sharpness.
    \item Select the response that maximizes semantic coherence.
    \item Select the response that grounds emotion in concrete, sensory imagery.
    \item Select the response that reflects ethical prudence and bias-aware judgment.
    \item Select the response that provides specificity instead of vague wording.
    \item Select the response that maintains a non-comic tone aligned with boredom.
    \item Select the response that avoids melodrama.
    \item Select the response that avoids tactile or injury-related confounds.
    \item Select the response that avoids culturally or religiously marked framing.
\end{enumerate}
\end{tcolorbox}
\noindent\begin{minipage}{\textwidth}
\captionof{figure}{Democratic ICAI constitution generated with GPT-4o on the Metaphors task}
\end{minipage}




\section{Licenses of Artifacts}
We use the following models and datasets in accordance with their respective terms. \textbf{\href{https://openai.com/policies/service-terms/}{GPT-4o}}, \textbf{\href{https://openai.com/policies/service-terms/}{GPT-5}}, and \textbf{\href{https://openai.com/policies/service-terms/}{text-embedding-3-small}} (OpenAI) are accessed via the OpenAI API under the \href{https://openai.com/policies/terms-of-use/}{OpenAI Terms of Use} and \href{https://openai.com/policies/usage-policies/}{Usage Policies}. \textbf{\href{https://github.com/QwenLM/Qwen/blob/main/LICENSE}{Qwen2.5-32B}} (Alibaba Cloud) is released under the Apache 2.0 License. Llama-3.1-70B and Llama-3.1-8B-Instruct is released under the Llama 3.1 Community License, and Mistral-7B-Instruct is released under the Apache 2.0 License.The \textbf{\href{https://litbench.vercel.app/}{LiTBench}} \citep{fein2025litbench} and \textbf{\href{https://huggingface.co/datasets/CNCL-Penn-State/MuCE}{MuCE}} \citep{ismayilzada2025creative} datasets are released under the MIT License. All artifacts are used within their permitted scope for non-commercial academic research.

\end{document}